\documentclass[11pt]{article}
\usepackage[letterpaper,margin=1in]{geometry}
\AtBeginDocument{\setlength{\bibsep}{2pt plus 1pt}}

\usepackage[round]{natbib}

\usepackage{amsmath,amssymb,amsthm,mathtools}
\usepackage{algorithm}
\usepackage{algpseudocode}
\usepackage{booktabs}
\usepackage{float}
\usepackage[hidelinks]{hyperref}
\hypersetup{pdftitle={Cost-Aware Stopping for Freeze-Thaw Bayesian Optimization: When Does One-Step Lookahead Suffice?},pdfauthor={Long Vu, Jongseong Brad Choi, Binoy Ravindran, Zhoulai Fu}}
\usepackage{url}
\usepackage{graphicx}
\ifdefined\pdfinfoomitdate\pdfinfoomitdate=1 \pdftrailerid{}\fi

\makeatletter
\newcommand{\spanfigures}{%
  \renewenvironment{figure}{\@dblfloat{figure}}{\end@dblfloat}}
\makeatother

\newtheorem{assumption}{Assumption}
\newtheorem{proposition}{Proposition}
\newtheorem{theorem}{Theorem}
\newtheorem{corollary}{Corollary}

\newtheorem{lemma}{Lemma}
\newtheorem*{definition*}{Definition}

\DeclareMathOperator*{\argmax}{arg\,max}
\newcommand{\E}{\mathbb{E}}
\newcommand{\Prob}{\mathbb{P}}
\newcommand{\cX}{\mathcal{X}}
\newcommand{\cC}{\mathcal{C}}
\newcommand{\cI}{\mathcal{I}}
\newcommand{\EI}{\mathrm{EI}}

\newcommand{\fitwidth}[1]{%
  \resizebox{\ifdim\width>\linewidth\linewidth\else\width\fi}{!}{#1}}

\renewcommand{\topfraction}{0.9}

\begin{document}

\title{Cost-Aware Stopping for Freeze--Thaw Bayesian Optimization: When Does One-Step Lookahead Suffice?}
\author{Long Vu\textsuperscript{1,2} \and Jongseong Brad Choi\textsuperscript{1,3} \and Binoy Ravindran\textsuperscript{2} \and Zhoulai Fu\textsuperscript{1,2,3}\\[0.4em]
{\normalsize \textsuperscript{1}State University of New York Korea \quad \textsuperscript{2}Virginia Tech \quad \textsuperscript{3}Stony Brook University}\\
{\normalsize \texttt{long-vu@outlook.com}}}
\date{}
\maketitle

\begin{abstract}
Freeze--thaw Bayesian optimization (BO) evaluates candidates in resumable increments but usually stops at a budget or
convergence test. We maximize a budget-free utility, the best observed score minus a cost per increment, with a one-step lookahead:
advance the candidate with the largest expected improvement (EI) under a fixed surrogate, and stop once no EI exceeds the
cost. This rule, \emph{EI stopping}, never falls below stopping immediately in expected utility, yet it can stop too early. We construct a certificate, a
lower bound on the value of two-step lookahead, that proves such a stop premature, and repeating this correction never lowers
expected utility under the surrogate. Under the surrogate, EI stops are premature on average at low costs and lose little at
high costs. On three hyperparameter optimization benchmarks, a pre-registered test finds EI stopping noninferior to the
correction in most high-cost settings, while the correction gains at low costs on two of the three; more information sharing
in the surrogate raises EI stopping's utility. On the surrogates of two published methods, EI stopping trades lower scores
for fewer training increments. EI stopping is a simple cost-aware reference whose utility depends on the surrogate.
\end{abstract}

\section{Introduction}
\label{sec:intro}

Freeze--thaw Bayesian optimization (BO) allocates evaluation effort among candidates in resumable increments, allowing any candidate's evaluation to be paused and resumed~\citep{swersky2014,ifbo2024,cfbo2025}. In hyperparameter optimization (HPO), candidates are hyperparameter configurations and increments are epochs or checkpoints. The original Freeze--Thaw BO method~\citep{swersky2014}
and later learning-curve models use this to allocate training
adaptively~\citep{wistuba2022dyhpo,kadra2023dpl,ifbo2024}, and curve extrapolation also ends
unpromising runs early~\citep{domhan2015,dai2019,adriaensen2023}. These searches usually end at a
supplied budget~\citep{swersky2014,wistuba2022dyhpo,ifbo2024}, by a regret-based convergence
test~\citep{makarova2022termination,ishibashi2023stopping,wilson2024prb} or by a utility-regret threshold~\citep{cfbo2025}. We treat
allocation and stopping in fixed-pool freeze--thaw BO as one decision problem and maximize the \emph{utility}
$\tilde y_\tau-\lambda\tau$, the best score observed when the search stops after $\tau$ evaluation
increments minus a cost $\lambda$ per increment (Section~\ref{sec:problem}). The cost alone governs termination, so no
budget or horizon is needed, and maximizing utility minimizes the cost-adjusted simple regret of \citet{xie2025coststop}
with a uniform cost (Appendix~\ref{app:formulation-omissions}).

We study a simple rule that targets this objective, EI stopping: train the configuration with the largest
expected improvement (EI)~\citep{jones1998ego} while $\max_i\EI_i>\lambda$ and stop otherwise. EI-threshold stopping
has been analyzed for its regret~\citep{nguyen2017ei,zhou2024correctedei}, and for complete
evaluations \citet{xie2025coststop} derive the threshold from the Pandora's box index~\citep{weitzman1979}. The rule looks one step ahead: the
Bellman equation splits the value of an advance into its net gain $\EI_i-\lambda$ and a
nonnegative option value (Equation~\eqref{eq:delta-recursion}), and EI stops are Bellman-optimal
under forward closure (Theorem~\ref{thm:myopic-optimal}), which learning curves need not satisfy:
an unprofitable increment can unlock a later fidelity, or reveal which other configuration
deserves training (Appendix~\ref{app:counterexample}); the
second requires a surrogate that shares information across configurations (Proposition~\ref{prop:gp-pool}).

We ask whether EI stopping is limited by the depth of its lookahead or by the surrogate. We compare it, exactly under the
surrogate and on recorded curves, with \emph{H2 stopping}, its certified two-step correction, which keeps greedy EI
allocation and at an EI stop certifies a candidate advance by its two-step continuation value. Our contributions are as follows.
\begin{itemize}
\item \textbf{Theory.} We characterize premature EI stops through the Bellman option value and show that a positive
lower bound on the value of a feasible two-step lookahead, the \emph{gate}, proves an EI stop premature under the
surrogate. The bound isolates a switching premium that arises only when the surrogate shares information across
configurations, and repeating the gate at every EI stop never lowers expected utility, for any finite cap on further
training (Section~\ref{sec:theory}). EI stopping itself trains for at most $1/\lambda$ further increments in expectation
and never falls below stopping immediately (Lemma~\ref{lem:ei-beats-stop}), the freeze--thaw BO counterpart of the
guarantee of~\citet{xie2025coststop}.
\item \textbf{Certified computation.} For Gaussian surrogates with affine fantasy updates,
slope-screened endpoint enclosures bound the selected configuration's H2 advantage with directed rounding~\citep{fousse2007mpfr} (Section~\ref{sec:analytical-enclosure}).
\item \textbf{Controlled empirical study.} We build three Gaussian learning-curve surrogates, GP-a, GP-at and
GP-atJ, that share asymptotes across configurations and differ only in how they share transients: not at all,
through the hyperparameter kernel, or also through a task-wide term; GP-atJ is the primary surrogate
(Section~\ref{sec:shared-models}). We cross them
with EI stopping and H2 stopping on LCBench~\citep{zimmer2021autopytorch}, PD1~\citep{wang2024hyperbo} and the
held-out ODBench~\citep{cfbo2025} (Section~\ref{sec:experiments}). Four findings emerge.
\textbf{(F1)} Under the surrogate, EI stops are premature on average at low costs and lose little at high costs.
\textbf{(F2)} On recorded curves, certified departures gain most at low costs, except on LCBench, and little at high costs, where EI stopping is mostly noninferior.
\textbf{(F3)} The pattern holds across all three Gaussian surrogates, and more information sharing significantly raises EI stopping's average utility.
\textbf{(F4)} On the surrogates of ifBO and DyHPO, EI stopping trades lower scores for fewer training increments.
\end{itemize}

\section{Background and Related Work}
\label{sec:background}

\paragraph{Learning-Curve Models and Sharing.}
Freeze--Thaw BO couples asymptotes through a GP over hyperparameters, models transients
independently, and has no rule for ending the search~\citep{swersky2014}. DyHPO, DPL, LC-PFN and ifBO/FT-PFN
are learned alternatives~\citep{wistuba2022dyhpo,kadra2023dpl,adriaensen2023,ifbo2024}, and CFBO
adds cost-sensitive acquisition, a stopping rule and a transfer-trained
extrapolator~\citep{cfbo2025}. Our controls are Gaussian processes~\citep{rasmussen2006gpml} that keep the
asymptote-plus-transient form of Freeze--Thaw BO's learning-curve model; they isolate information sharing
while keeping fantasies certifiable.

\paragraph{Cost-Aware Stopping.}
EI-threshold termination has prior treatments~\citep{nguyen2017ei,zhou2024correctedei}.
The closest work to our objective is \citet{xie2025coststop}, who pair a cost-aware stopping
rule with PBGI and LogEIPC~\citep{xie2024pbgi}, the log variant~\citep{ament2023logeipc} of EI per unit cost~\citep{snoek2012practical}; under uniform cost, LogEIPC with
their rule is EI stopping. Their evaluations are complete, a full 51-epoch run on
LCBench, and multi-fidelity evaluation is listed as an extension; we study freeze--thaw
BO, where an action is one epoch or checkpoint, and ask what downstream value the EI stopping boundary misses there. Other whole-search criteria use error scales, regret gaps or
probabilistic tolerances~\citep{makarova2022termination,ishibashi2023stopping,wilson2024prb,wang2026epsdelta}.
These tests are cost-blind: the cost of continuing does not enter them, whereas EI stopping weighs the
expected gain of the next increment against its cost.
Fixed-schedule methods~\citep{jamieson2016sh,li2018hyperband,falkner2018bohb,li2020asha} follow a preset
schedule. We compare with neither.

\paragraph{Non-Myopic Decisions and Policy Improvement.}
Forward closure makes one-step stopping optimal in classical monotone
problems and on transition-closed sets of metalevel
states~\citep{chowrobbinssiegmund1971,ferguson2008stopping,hay2012selecting}; Theorem~\ref{thm:myopic-optimal} gives the version
for our setting. Rollout and policy improvement instead
compare actions using a base policy~\citep{tesauro1996online,bertsekas1997rollout}, and non-myopic BO values future information by simulating the observations it would
receive~\citep{jiang2020nonmyopic,yue2020rollout}.
Conservative and safe policy improvement certify gains from
sampled data~\citep{kakade2002approximately,laroche2019spibb}; ours holds under the surrogate. Martingale duality and information relaxations give dual upper bounds on the optimal
value~\citep{rogers2002,haughkogan2004,brownsmithsun2010}; our certificate gives lower bounds only and thus measures what a
correction recovers, and F1 (Section~\ref{sec:controlled-results}) computes the exact gap on reduced instances. Appendix~\ref{app:related-switching}
relates allocation in freeze--thaw BO to optimal switching and Pandora-type
search~\citep{carmonaludkovski2008,weitzman1979}.

\section{Problem Formulation}
\label{sec:formulation}

\subsection{Fixed-Pool Freeze--Thaw BO and Utility}
\label{sec:problem}

\paragraph{State.} Let $\cX=\{x_1,\ldots,x_N\}$ be a fixed pool of arms (candidates) and let one action advance one candidate's evaluation by one fidelity unit. We call the action an \emph{advance} and the unit an \emph{increment}. In our HPO experiments, arms are hyperparameter configurations and increments are benchmark-defined training checkpoints. After $b$ increments, the incumbent $\tilde y_b$ is the best score so far, starting from a baseline $y_0$, and $t_{i,b}$ is arm $i$'s local fidelity. The history state is $S_b=(\cC_b,t_{1:N,b},\tilde y_b,b)$.

\paragraph{Planning Law.} The planning law $p_\theta(\cdot\mid S_b,i)$ is a fixed surrogate's predictive distribution of arm $i$'s next score, with parameters $\theta$ set in advance. The common context $\cC_b$ may contain every pool observation and configuration, which lets $p_\theta$ update predictions for all arms after any observation.

\paragraph{Actions and Transition.} The actions at $S_b$ are $\{\mathrm{stop}\}\cup\cI(S_b)$, where $\cI(S)$ contains arms with another unit available. Advancing arm $i$ from $S_b$ costs one unit and reveals exactly one new number, the score $Y_{b+1}$ of arm $i$ at its next fidelity $t_{i,b}+1$; under the planning law, $Y_{b+1}\sim p_\theta(\cdot\mid S_b,i)$, and under the curve models of Section~\ref{sec:shared-models} it is $Y_i(t_{i,b}+1)$. Everything else about the step is determined by $S_b$ and $i$; the successor $S_{b+1}=F(S_b,i,Y_{b+1})$ is a deterministic map applied to that one draw,
\[
\begin{aligned}
F(S_b,i,Y_{b+1})=\bigl(&\cC_b\cup\{(i,t_{i,b}+1,Y_{b+1})\},\\
&t_{1:N,b}+e_i,\ \max\{\tilde y_b,Y_{b+1}\},\ b+1\bigr):
\end{aligned}
\]
the observation joins the context, arm $i$'s fidelity advances by one ($e_i$ is the $i$th unit vector), the incumbent absorbs the new score, and the step count increments. Whenever an action $i$ is evaluated at $S_b$, $S_{b+1}$ denotes this successor, whether or not a policy would take $i$.

\paragraph{Policies and Conventions.} Unless stated otherwise, every action maximum is over $\cI(S_b)$, every successor maximum is over $\cI(S_{b+1})$, and $\max\varnothing=-\infty$. The uncapped model is the special case $\cI(S)=[N]$; finite-record replay masks exhausted arms. Availability is arm-local: advancing $i$ never makes another arm available. A policy $\pi$ chooses, at each state $S_b$, an action in $\{\mathrm{stop}\}\cup\cI(S_b)$, measurably in the history; $\tau_\pi$ is the first step at which it stops, a stopping time, and $\pi$ is admissible when $\E_{\theta,\pi}[\tau_\pi\mid S_0]<\infty$. Throughout, $\E_{\theta,\pi}[\,\cdot\mid S_b]$ is expectation under the planning law and policy $\pi$ from $S_b$, $\E_\theta[\,\cdot\mid S_b,i]$ the one-step expectation over $Y_{b+1}$ alone, and $\tau_{\rm EI}$, $\tau_{\rm H2}$ abbreviate $\tau_{\pi_{\rm EI}}$, $\tau_{\pi_{\rm H2}}$.

The planning law is supported on a bounded score interval, normalized to $[0,1]$. We maximize
\begin{equation}
\sup_{\pi}\;\E_{\theta,\pi}\!\left[\tilde y_{\tau_\pi}-\lambda\tau_\pi\mid S_0\right],
\qquad \lambda>0,
\label{eq:opt-problem}
\end{equation}
over admissible policies. The stop-now payoff is $h_\lambda(S_b)=\tilde y_b-\lambda b$.
Here $\lambda$ is a score penalty per increment, chosen in advance and read against the
planning law's predictive scale (Appendix~\ref{app:fixed-law-sensitivity}). The term $\lambda b$ is already incurred at state $S_b$;
comparisons charge only additional increments.
For any admissible $\pi$ and state $S_b$, the value function of $\pi$ is
$V_\pi(S_b)=\E_{\theta,\pi}[\tilde y_{\tau_\pi}-\lambda\tau_\pi\mid S_b]$, and its action-value
function $Q_\pi(S_b,i)=\E_\theta[V_\pi(F(S_b,i,Y_{b+1}))\mid S_b,i]$ is the value of forcing
$i$ once and then following $\pi$.

\subsection{Bellman Equation and Option Value}
\label{sec:bellman}

The optimal value function $V_\lambda^\star=\sup_\pi V_\pi$ and its action-value function
$Q_\lambda^\star$ satisfy the Bellman equation
\begin{align}
V_\lambda^\star(S_b)&=\max\!\left\{h_\lambda(S_b),\max_{i\in\cI(S_b)}Q_\lambda^\star(S_b,i)\right\},\label{eq:bellman-linear}\\
Q_\lambda^\star(S_b,i)&=\E_\theta\!\left[V_\lambda^\star(F(S_b,i,Y_{b+1}))\mid S_b,i\right].\nonumber
\end{align}
Define $\Delta_\lambda^\star(S_b,i)=Q_\lambda^\star(S_b,i)-h_\lambda(S_b)$ and the Bellman stopping region $\mathcal S_\lambda=\{S:h_\lambda(S)=V_\lambda^\star(S)\}$. The centered advantage satisfies (Appendix~\ref{app:delta-recursion})
\begin{equation}
\begin{aligned}
\Delta_\lambda^\star(S_b,i)
&=\underbrace{\EI_i(S_b)-\lambda}_{\text{net gain}}
+\underbrace{O_i(S_b)}_{\text{option value}},\\
O_i(S_b)
&=\E_\theta\!\left[\max\!\left\{0,\max_{j\in\cI(S_{b+1})}\Delta_\lambda^\star(S_{b+1},j)\right\}\mid S_b,i\right]\ \ge\ 0,
\end{aligned}
\label{eq:delta-recursion}
\end{equation}
where $\EI_i(S_b)=\E_\theta[(Y_{b+1}-\tilde y_b)^+\mid S_b,i]$~\citep{jones1998ego}. At an EI stop, the decision is premature exactly when some available $i$ has $O_i(S)>\lambda-\EI_i(S)\ge0$. The option value includes later same-arm advances as well as observation-dependent re-ranking and switching. Both sides of the test move with the cost: the payoff of stopping is free of $\lambda$ while every continuation pays it, so $O_i(S)$ is nonincreasing in $\lambda$ and $\lambda-\EI_i(S)$ increasing; among states that are EI stops at two costs, a stop premature at the higher cost is premature at the lower, never the reverse.

\paragraph{Standing Assumptions.} Three assumptions hold throughout.
Assumption~\ref{assm:fixed-sim}: the planning law $p_\theta$ is a fixed measurable kernel, never refitted during a run.
Assumption~\ref{assm:bounded-perf}: scores stay in bounds $[y_{\min},y_{\max}]$ fixed in advance.
Assumption~\ref{assm:coercive}: every increment costs $\lambda>0$.
Appendix~\ref{app:assumptions} gives the measurability details.

\section{Pool-Aware Continuation and Certification}
\label{sec:theory}

\subsection{What the Pool Adds at an EI Stop}
Let $g(S)=\max(\{0\}\cup\{\EI_i(S):i\in\cI(S)\})$ and
$\mathcal S^{\mathrm{myo}}_\lambda=\{S:g(S)\le\lambda\}$. The policy $\pi_{\rm EI}$, \emph{EI stopping} (EI-S), advances the lowest-index EI maximizer outside this set and stops
inside it, at an \emph{EI stop}. With $y_{\max}$ the upper bound on scores (Assumption~\ref{assm:bounded-perf}), write $d(S_b)=y_{\max}-\tilde y_b$ for the remaining score headroom.

\begin{lemma}[EI stopping terminates and is no worse than stopping immediately]\label{lem:ei-beats-stop}
Under Assumptions~\ref{assm:fixed-sim}--\ref{assm:coercive}, from every state $S_b$,
$\pi_{\rm EI}$ satisfies $\E_{\theta,\pi_{\rm EI}}[\tau_{\rm EI}-b\mid S_b]\le d(S_b)/\lambda$ and
\[
V_{\pi_{\rm EI}}(S_b)-h_\lambda(S_b)
=\E_{\theta,\pi_{\rm EI}}\Big[\textstyle\sum_{k=b}^{\tau_{\rm EI}-1}\big(g(S_k)-\lambda\big)\Big]\ge0.
\]
\end{lemma}
In words, EI stopping cannot train indefinitely, and every increment it trains has positive expected net gain.
\begin{theorem}[Exact stopping under forward closure]\label{thm:myopic-optimal}
Under Assumptions~\ref{assm:fixed-sim}--\ref{assm:coercive}, if every action from every state
in $\mathcal S^{\mathrm{myo}}_\lambda$ leads almost surely back into it (forward closure), then $\mathcal S_\lambda=\mathcal S^{\mathrm{myo}}_\lambda$ and
$O_i(S_b)=0$ there for every available $i$.
\end{theorem}
One-step lookahead is thus exactly right when no action at an EI stop can lead to a state where EI stopping
would continue. Forward closure is sufficient but not necessary, and learning curves can violate it
(Appendix~\ref{app:counterexample}); proofs are in Appendix~\ref{app:proofs}. For capped replay, write $\EI^{\rm cap}_i(S_{b+1})=\EI_i(S_{b+1})$ when arm $i$
remains available and $0$ otherwise.

At an EI stop define the centered advantage of the pool-aware two-step
lower rollout
\begin{equation}
A^{\rm pool}_{2,i}(S_b)=(\EI_i(S_b)-\lambda)+\E_\theta\!\left[(g(S_{b+1})-\lambda)^+\mid S_b,i\right].
\label{eq:depth-two-advantage}
\end{equation}
This is the exact centered value of a feasible policy: force $i$ once; at the
child $S_{b+1}$, advance a lowest-index EI maximizer iff $g(S_{b+1})>\lambda$; then stop.
The first term is the expected net root increment, and the positive-part term is
the value of an observation-dependent second action. \begin{proposition}[Two-step lower rollout]\label{prop:depth-two-rollout}
Under Assumptions~\ref{assm:fixed-sim}--\ref{assm:coercive}, for every
$S_b\in\mathcal S^{\mathrm{myo}}_\lambda$ and $i\in\cI(S_b)$,
\[
h_\lambda(S_b)+A^{\rm pool}_{2,i}(S_b)\le Q_{\pi_{\rm EI}}(S_b,i).
\]
Thus $A^{\rm pool}_{2,i}(S_b)>0$ certifies a profitable feasible departure from stop-now.
\end{proposition}

The same-arm advantage $A_{2,i}$ replaces $g(S_{b+1})$ in~\eqref{eq:depth-two-advantage} by
$\EI^{\rm cap}_i(S_{b+1})$, allowing only a second advance of the root arm, hence $A_{2,i}\le A^{\rm pool}_{2,i}$. Their difference, the switching premium
$\Gamma_i=A^{\rm pool}_{2,i}-A_{2,i}\ge0$, is positive exactly when, with positive probability,
another arm's EI at the child exceeds both $\lambda$ and the root arm's. Switching is essential to a
profitable H2 departure when $A_{2,i}\le0<A^{\rm pool}_{2,i}$. If no other arm's EI can exceed
$\lambda$ at the child, as under per-arm conditioning, the two advantages coincide at every
EI stop (Corollary~\ref{cor:same-arm-reduction}, Appendix~\ref{app:same-arm}).

\subsection{H2 Stopping}
H2 stopping, the certified rule, retains greedy EI allocation wherever EI recommends continuing. At an EI stop, our implementation ranks candidates by estimated two-step continuation value and certifies the selected candidate. This candidate can differ from the immediate-EI maximizer: its observation may make a subsequent advance on itself or another configuration profitable. The guarantee allows any measurable candidate rule $\kappa$ on nonterminal EI stops ($\cI(S)\ne\varnothing$), with
$\kappa(S)\in\cI(S)$, and a current-history-measurable rigorous bound
$\underline A^{\rm pool}_{2,\kappa(S)}(S)\le A^{\rm pool}_{2,\kappa(S)}(S)$. The
complete justification at a nonterminal EI stop is the one-sided chain
\begin{equation}
\underline A^{\rm pool}_{2,\kappa(S)}(S)\le A^{\rm pool}_{2,\kappa(S)}(S)\le Q_{\pi_{\rm EI}}(S,\kappa(S))-V_{\pi_{\rm EI}}(S),\qquad V_{\pi_{\rm EI}}(S)=h_\lambda(S).
\label{eq:selective-lower-chain}
\end{equation}
The first inequality is the certified lower enclosure, the second is policy comparison, and the equality holds because $\pi_{\rm EI}$ stops at $S$.

This equality explains the focus on the stopping boundary: EI's reference value is known exactly there. While EI continues, its value includes future advances, so a positive two-step advantage over stopping need not improve on its selected action. Certifying an allocation change would require an additional comparison with EI's continuation value, through evaluation or suitable bounds. At an EI stop, the gate establishes policy improvement directly from the two-step lower bound.

\begin{algorithm}[t]
\caption{H2 stopping, $\pi_{\rm H2}$.}
\label{alg:selective-receding}
\begin{algorithmic}[1]
\State \textbf{input:} state $S$, cost $\lambda>0$
\While{$\cI(S)\ne\varnothing$}
  \State $i\gets$ lowest-index maximizer of $\EI_i(S)$
  \State \textbf{if} $\EI_i(S)>\lambda$ \textbf{then} advance $i$; \textbf{continue}
  \State\label{line:certify} select $\kappa(S)$; certify $\underline A^{\rm pool}_{2,\kappa(S)}(S)\le A^{\rm pool}_{2,\kappa(S)}(S)$
  \State \textbf{if} $\underline A^{\rm pool}_{2,\kappa(S)}(S)>0$ \textbf{then} advance $\kappa(S)$\label{line:gate}
  \State \textbf{else return} \textsc{stop} \Comment{abstain: keep the EI stop}
\EndWhile
\State \Return \textsc{stop}
\end{algorithmic}
\end{algorithm}
Algorithm~\ref{alg:selective-receding} defines $\pi_{\rm H2}$, \emph{H2 stopping} (H2-S): it follows $\pi_{\rm EI}$ until an EI stop,
and at a nonterminal EI stop it advances $\kappa(S)$ iff the gate
$\underline A^{\rm pool}_{2,\kappa(S)}(S)>0$ opens (line~\ref{line:gate}). It executes one advance and reassesses at the successor; the two-step rollout justifies that advance without committing the policy to its second action.
The certificate is one-sided. An open gate certifies profitable continuation for the selected candidate, and a closed gate is an abstention that returns the EI stop uncertified. The guarantee requires a profitable departure, not an optimal choice among candidates.

\begin{corollary}[H2 stopping improves on EI stopping]
\label{cor:receding-policy-improvement}
Under Assumptions~\ref{assm:fixed-sim}--\ref{assm:coercive}, fix an initial
state $S$ such that some deterministic $H(S)<\infty$ bounds the number of further increments on every feasible path, and suppose the bound above is available at
every $\pi_{\rm H2}$-reachable nonterminal EI stop. Then
\[
V_{\pi_{\rm H2}}(S)\ge V_{\pi_{\rm EI}}(S)
\]
and the same holds throughout that reachable subtree. At each state where the gate opens,
$V_{\pi_{\rm H2}}(S)>V_{\pi_{\rm EI}}(S)$. Finite-record replay supplies $H(S)$.
\end{corollary}
It is policy improvement with base policy $\pi_{\rm EI}$~\citep{bertsekas1997rollout}, proved by finite-cap backward induction
(Appendix~\ref{app:receding-proof}). With
Lemma~\ref{lem:ei-beats-stop},
$h_\lambda(S)\le V_{\pi_{\rm EI}}(S)\le V_{\pi_{\rm H2}}(S)$ at every reachable state.
The corollary compares ideal policies. The implementation certifies each departure; applying the full policy comparison to floating-point EI also requires action agreement throughout the reachable subtree (Appendix~\ref{app:enclosure-procedure}).

\subsection{Analytical Enclosure}
\label{sec:analytical-enclosure}
We compute the certified bound of Algorithm~\ref{alg:selective-receding} (line~\ref{line:certify}) for the fixed Gaussian
planning laws of Section~\ref{sec:shared-models}. Only this enclosure is specific to Gaussian surrogates;
Corollary~\ref{cor:receding-policy-improvement} holds for any fixed planning law. The present enclosure does not cover the learned surrogates
of Section~\ref{sec:external-comparators}: FT-PFN's fantasized predictive has no known affine Gaussian form, and DyHPO refits its deep-kernel GP after each observation
(Appendix~\ref{app:no-certificate-learned}).
At a fixed state and selected root $i$, write $q=\tilde y(S)$ and $z\in[0,1]$
for the root observation. Each feasible child $j$ has
$X_j\mid z\sim\mathcal N(\mu_j(z),\sigma_j^2)$ restricted and renormalized to
$[0,1]$, where $\mu_j(z)=m_j+\gamma_j(z-m_i)$ with root and child predictive means $m_i,m_j$ and slope $\gamma_j$, and $\sigma_j>0$ is independent of $z$.
These are sequential operational Gaussian updates, each followed by one-step truncation.
Child $j$'s EI after the root outcome $z$ is $f_j(z)=\E[(X_j-\max\{q,z\})^+\mid z]$.
\begin{proposition}[Endpoint enclosure]\label{prop:endpoint-main}
If $\gamma_j\le0$, $f_j$ is nonincreasing on $[0,1]$. If $0\le\gamma_j\le1$, it is nondecreasing below
$q$ and nonincreasing above $q$. Hence on any cell $[z_-,z_+]$ that does not cross $q$, $f_j$ lies
between its values at $z_-$ and $z_+$.
\end{proposition}
The full statement, with directed rounding and interval-enclosed coefficients, is
Proposition~\ref{prop:slope-endpoint} (Appendix~\ref{app:slope-endpoint-proof}); a child
whose enclosed slope fits neither case is bounded by interval evaluation over the cell. The enclosure is rigorous because the gate is a sign test: it opens when a positive-part integral exceeds a deficit of at most $\lambda$, a comparison that quadrature or rounding error could decide when $\lambda$ is small. An open gate proves positive continuation value under the planning law; its realized effect on recorded curves is measured separately.

\paragraph{Screen, Enclose, Integrate.}
The certificate prunes children whose rigorous peak is at most $\lambda$, bounds the positive-part
excess on a partition of $[0,1]$ refined at $q$, and encloses the expectation
in~\eqref{eq:depth-two-advantage} with rigorous root-mass bounds.
Under all three surrogates the gate costs $O(N^3)$ operations for $N$ configurations, against $O(N^2)$ for an EI decision (Appendix~\ref{app:enclosure-procedure}).

\newcommand{\BCeilingFinding}{%
Across the family the UCL is at most $0.00035$ on LCBench, $0.00505$ on PD1 and $0.00456$ on ODBench; H2 stopping thus raises mean utility over EI stopping by at most $0.00505$ on seeds 1--50; the UCL concerns that correction and bounds the gap to no other policy.
}
\newcommand{\BNoninferiorityFinding}{%
With the tolerance fixed at one increment's cost before the run, EI stopping is noninferior in 10 of 21 cells: on LCBench 6 of 7 cells (all but $\lambda=0.0001$); on PD1 2 of 7 cells (only $\lambda\in\{0.005, 0.01\}$); on ODBench 2 of 7 cells (only $\lambda\in\{0.005, 0.01\}$). The remaining cells leave noninferiority open.
}

\newcommand{\BHorizonTable}{%
\begin{table}[ht]
\centering\small
\caption{H2 stopping (H2-S) minus EI stopping (EI-S) under GP-atJ, split by native horizon: equal-task paired $U^{\text{GP-atJ}}_{\text{H2-S}}-U^{\text{GP-atJ}}_{\text{EI-S}}$ with its two-sided simultaneous 95\% fixed-task seed-block band ($^\dagger$/$^\ddagger$: above/below zero), one 28-column family per seed set (critical values 3.27 and 3.24). Scores lie in $[0,1]$, and utility is in the same units. The long-horizon group holds the 3 PD1 tasks with more than 100 native checkpoints; the short-horizon group holds the other 12. ``Tasks $+$'' counts tasks whose 50-seed mean favors H2-S, on seeds 1--50 and 51--100. ``Cap at EI stop'' is the share of seed-1--50 episodes in which some configuration has consumed every recorded checkpoint when EI stops; ``No cap hit'' is the contrast over the group's other episodes, pooled and without a band, on seeds 1--50 and 51--100. No verdict is attached.}
\label{tab:horizon-split}
\fitwidth{%
\begin{tabular}{llrrcrr}
\toprule
Group & $\lambda$ & seeds 1--50 [band] & seeds 51--100 [band] & Tasks $+$ & Cap at EI stop & No cap hit \\
 & & & & (seeds 1--50; 51--100) & & (seeds 1--50; 51--100) \\
\midrule
LCBench & 0.0001 & $+0.00014$ [-0.00009, +0.00037] & $+0.00003$ [-0.00016, +0.00021] & 13/35 ; 8/35 & $46.5\%$ & $+0.00028$ ; $+0.00007$ \\
LCBench & 0.0002 & $+0.00009$ [-0.00019, +0.00036] & $-0.00001$ [-0.00017, +0.00016] & 11/35 ; 9/35 & $37.4\%$ & $+0.00019$ ; $+0.00005$ \\
LCBench & 0.0005 & $-0.00047$ [-0.00135, +0.00040] & $-0.00046$ [-0.00084, -0.00007]$^{\ddagger}$ & 8/35 ; 4/35 & $24.2\%$ & $+0.00002$ ; $-0.00022$ \\
LCBench & 0.001 & $-0.00049$ [-0.00073, -0.00024]$^{\ddagger}$ & $-0.00038$ [-0.00060, -0.00017]$^{\ddagger}$ & 5/35 ; 4/35 & $10.2\%$ & $-0.00051$ ; $-0.00036$ \\
LCBench & 0.002 & $-0.00042$ [-0.00082, -0.00002]$^{\ddagger}$ & $-0.00052$ [-0.00080, -0.00025]$^{\ddagger}$ & 6/35 ; 4/35 & $0.0\%$ & $-0.00042$ ; $-0.00052$ \\
LCBench & 0.005 & $-0.00017$ [-0.00048, +0.00013] & $-0.00016$ [-0.00036, +0.00003] & 3/35 ; 3/35 & $0.0\%$ & $-0.00017$ ; $-0.00016$ \\
LCBench & 0.01 & $-0.00069$ [-0.00159, +0.00021] & $-0.00071$ [-0.00171, +0.00030] & 4/35 ; 4/35 & $0.0\%$ & $-0.00069$ ; $-0.00071$ \\
\midrule
PD1, long horizon & 0.0001 & $+0.00691$ [+0.00139, +0.01244]$^{\dagger}$ & $+0.00783$ [+0.00033, +0.01534]$^{\dagger}$ & 2/3 ; 2/3 & $9.3\%$ & $+0.00762$ ; $+0.00890$ \\
PD1, long horizon & 0.0002 & $+0.01728$ [+0.00695, +0.02761]$^{\dagger}$ & $+0.01946$ [+0.00864, +0.03028]$^{\dagger}$ & 3/3 ; 3/3 & $4.7\%$ & $+0.01813$ ; $+0.02041$ \\
PD1, long horizon & 0.0005 & $+0.00840$ [+0.00203, +0.01478]$^{\dagger}$ & $+0.00634$ [+0.00354, +0.00914]$^{\dagger}$ & 3/3 ; 3/3 & $0.0\%$ & $+0.00840$ ; $+0.00634$ \\
PD1, long horizon & 0.001 & $+0.00826$ [+0.00030, +0.01621]$^{\dagger}$ & $+0.00816$ [+0.00134, +0.01497]$^{\dagger}$ & 2/3 ; 2/3 & $0.0\%$ & $+0.00826$ ; $+0.00816$ \\
PD1, long horizon & 0.002 & $+0.00857$ [-0.00138, +0.01852] & $+0.00786$ [-0.00126, +0.01699] & 2/3 ; 2/3 & $0.0\%$ & $+0.00857$ ; $+0.00786$ \\
PD1, long horizon & 0.005 & $+0.00739$ [-0.00154, +0.01631] & $+0.01013$ [-0.00041, +0.02066] & 2/3 ; 3/3 & $0.0\%$ & $+0.00739$ ; $+0.01013$ \\
PD1, long horizon & 0.01 & $+0.01336$ [+0.00707, +0.01966]$^{\dagger}$ & $+0.00818$ [+0.00215, +0.01420]$^{\dagger}$ & 2/3 ; 2/3 & $0.0\%$ & $+0.01336$ ; $+0.00818$ \\
\midrule
PD1, short horizon & 0.0001 & $-0.00027$ [-0.00044, -0.00010]$^{\ddagger}$ & $-0.00027$ [-0.00044, -0.00011]$^{\ddagger}$ & 1/12 ; 3/12 & $91.3\%$ & $-0.00011$ ; $+0.00022$ \\
PD1, short horizon & 0.0002 & $-0.00043$ [-0.00055, -0.00031]$^{\ddagger}$ & $-0.00039$ [-0.00053, -0.00026]$^{\ddagger}$ & 1/12 ; 2/12 & $88.3\%$ & $-0.00027$ ; $-0.00035$ \\
PD1, short horizon & 0.0005 & $-0.00137$ [-0.00202, -0.00071]$^{\ddagger}$ & $-0.00107$ [-0.00150, -0.00064]$^{\ddagger}$ & 0/12 ; 0/12 & $70.8\%$ & $-0.00152$ ; $-0.00139$ \\
PD1, short horizon & 0.001 & $-0.00104$ [-0.00177, -0.00031]$^{\ddagger}$ & $-0.00124$ [-0.00183, -0.00065]$^{\ddagger}$ & 1/12 ; 1/12 & $41.3\%$ & $-0.00067$ ; $-0.00080$ \\
PD1, short horizon & 0.002 & $-0.00090$ [-0.00183, +0.00003] & $-0.00066$ [-0.00142, +0.00009] & 2/12 ; 3/12 & $13.2\%$ & $-0.00057$ ; $-0.00043$ \\
PD1, short horizon & 0.005 & $-0.00038$ [-0.00121, +0.00046] & $-0.00038$ [-0.00096, +0.00020] & 3/12 ; 2/12 & $0.0\%$ & $-0.00038$ ; $-0.00038$ \\
PD1, short horizon & 0.01 & $-0.00017$ [-0.00055, +0.00020] & $-0.00043$ [-0.00155, +0.00069] & 3/12 ; 3/12 & $0.0\%$ & $-0.00017$ ; $-0.00043$ \\
\midrule
ODBench & 0.0001 & $+0.00073$ [-0.00007, +0.00152] & $+0.00106$ [+0.00000, +0.00212]$^{\dagger}$ & 6/9 ; 8/9 & $11.3\%$ & $+0.00083$ ; $+0.00120$ \\
ODBench & 0.0002 & $+0.00118$ [-0.00036, +0.00272] & $+0.00089$ [-0.00014, +0.00193] & 6/9 ; 4/9 & $7.3\%$ & $+0.00128$ ; $+0.00096$ \\
ODBench & 0.0005 & $+0.00109$ [-0.00028, +0.00245] & $+0.00090$ [-0.00126, +0.00305] & 5/9 ; 5/9 & $2.0\%$ & $+0.00111$ ; $+0.00091$ \\
ODBench & 0.001 & $+0.00221$ [-0.00032, +0.00474] & $+0.00046$ [-0.00111, +0.00203] & 5/9 ; 4/9 & $0.9\%$ & $+0.00223$ ; $+0.00047$ \\
ODBench & 0.002 & $+0.00056$ [-0.00129, +0.00242] & $+0.00031$ [-0.00257, +0.00318] & 5/9 ; 3/9 & $0.4\%$ & $+0.00057$ ; $+0.00031$ \\
ODBench & 0.005 & $-0.00067$ [-0.00249, +0.00115] & $-0.00093$ [-0.00239, +0.00053] & 2/9 ; 1/9 & $0.0\%$ & $-0.00067$ ; $-0.00093$ \\
ODBench & 0.01 & $-0.00013$ [-0.00070, +0.00044] & $-0.00018$ [-0.00059, +0.00024] & 3/9 ; 3/9 & $0.0\%$ & $-0.00013$ ; $-0.00018$ \\
\bottomrule
\end{tabular}
}
\end{table}
}
\newcommand{\BHorizonFinding}{%
On the 3 long-horizon PD1 tasks H2 stopping raises mean utility at every cost, by $0.0069$ to $0.0173$ on seeds 1--50, with bands above zero at 5 of 7 costs; on the 12 short-horizon PD1 tasks its mean is negative at every cost, at most $-0.00017$, with bands below zero at 4 of 7 costs and only 11 of 84 task--cost means favoring H2 stopping. Seeds 51--100 repeat both patterns, with 5 long-horizon bands above zero and 7 of 7 short-horizon means negative.
}

\newcommand{\FLSensTable}{%
\begin{table}[ht]\centering\scriptsize\setlength{\tabcolsep}{3pt}\renewcommand{\arraystretch}{0.8}
\caption{EI stopping (EI-S) under GP-atJ with one of its settings changed, minus EI-S at the primary settings, $U^{\text{GP-atJ}}_{\text{EI-S}}(\text{changed})-U^{\text{GP-atJ}}_{\text{EI-S}}(\text{primary})$: the fresh-observation noise variance scaled by $\frac14$ or $4$, or both prior variances scaled by $\frac12$ or $2$, paired per episode on the same seeds 1--50, pools and costs. Positive values favor the changed setting. Scores lie in $[0,1]$, and utility is in the same units; $^\dagger$/$^\ddagger$: pointwise 95\% seed-block band above/below zero. $\Delta\tau$: mean change in increments.}
\label{tab:fixed-law-sensitivity}
\fitwidth{\begin{tabular}{@{}lrrrrrrrrr@{}}\toprule
 & & \multicolumn{4}{c}{$U^{\text{GP-atJ}}_{\text{EI-S}}(\text{changed})-U^{\text{GP-atJ}}_{\text{EI-S}}(\text{primary})$} & \multicolumn{4}{c}{$\Delta\tau$} \\ \cmidrule(lr){3-6}\cmidrule(lr){7-10}
Bench. & $\lambda$ & Noise $\times\frac14$ & Noise $\times4$ & Prior $\times\frac12$ & Prior $\times2$ & Noise $\times\frac14$ & Noise $\times4$ & Prior $\times\frac12$ & Prior $\times2$ \\ \midrule
LCBench & 0.0001 & $-0.0002$ & $-0.0046^{\ddagger}$ & $-0.0001$ & $-0.0005^{\ddagger}$ & $-39.5$ & $+85.1$ & $-11.3$ & $+14.6$ \\
 & 0.0002 & $+0.0020^{\dagger}$ & $-0.0101^{\ddagger}$ & $+0.0002$ & $-0.0008^{\ddagger}$ & $-29.9$ & $+72.5$ & $-9.8$ & $+8.3$ \\
 & 0.0005 & $+0.0055^{\dagger}$ & $-0.0245^{\ddagger}$ & $+0.0025^{\dagger}$ & $-0.0026^{\ddagger}$ & $-18.0$ & $+59.1$ & $-10.0$ & $+7.6$ \\
 & 0.001 & $+0.0081^{\dagger}$ & $-0.0388^{\ddagger}$ & $+0.0027^{\dagger}$ & $-0.0065^{\ddagger}$ & $-11.6$ & $+43.9$ & $-6.2$ & $+7.5$ \\
 & 0.002 & $+0.0051^{\dagger}$ & $-0.0423^{\ddagger}$ & $+0.0050^{\dagger}$ & $-0.0073^{\ddagger}$ & $-3.9$ & $+23.8$ & $-5.1$ & $+4.5$ \\
 & 0.005 & $+0.0031^{\dagger}$ & $-0.0210^{\ddagger}$ & $+0.0164^{\dagger}$ & $-0.0121^{\ddagger}$ & $-1.3$ & $+5.4$ & $-6.2$ & $+3.7$ \\
 & 0.01 & $+0.0007^{\dagger}$ & $-0.0169^{\ddagger}$ & $+0.0353^{\dagger}$ & $-0.0297^{\ddagger}$ & $-0.2$ & $+1.8$ & $-5.4$ & $+4.0$ \\
\midrule
PD1 & 0.0001 & $+0.0030^{\dagger}$ & $-0.0039^{\ddagger}$ & $-0.0053^{\ddagger}$ & $+0.0038^{\dagger}$ & $-71.0$ & $+181.5$ & $-34.8$ & $+14.8$ \\
 & 0.0002 & $+0.0078^{\dagger}$ & $-0.0168^{\ddagger}$ & $+0.0012$ & $+0.0081^{\dagger}$ & $-46.1$ & $+172.3$ & $-19.5$ & $+14.9$ \\
 & 0.0005 & $+0.0063^{\dagger}$ & $-0.0359^{\ddagger}$ & $+0.0009$ & $-0.0038^{\ddagger}$ & $-22.4$ & $+87.1$ & $-15.5$ & $+13.4$ \\
 & 0.001 & $+0.0106^{\dagger}$ & $-0.0470^{\ddagger}$ & $+0.0030^{\dagger}$ & $-0.0044^{\ddagger}$ & $-13.4$ & $+49.3$ & $-11.0$ & $+7.2$ \\
 & 0.002 & $+0.0088^{\dagger}$ & $-0.0382^{\ddagger}$ & $+0.0006$ & $-0.0095^{\ddagger}$ & $-5.4$ & $+18.4$ & $-4.4$ & $+6.6$ \\
 & 0.005 & $+0.0043^{\dagger}$ & $-0.0235^{\ddagger}$ & $-0.0146^{\ddagger}$ & $+0.0014$ & $-0.8$ & $+2.7$ & $-4.0$ & $+2.1$ \\
 & 0.01 & $+0.0121^{\dagger}$ & $-0.0227^{\ddagger}$ & $+0.0069^{\dagger}$ & $-0.0063^{\ddagger}$ & $+0.1$ & $+1.2$ & $-2.8$ & $+2.5$ \\
\midrule
ODBench & 0.0001 & $-0.0130^{\ddagger}$ & $+0.0134^{\dagger}$ & $-0.0006$ & $+0.0006$ & $-7.1$ & $+12.2$ & $-3.2$ & $+1.1$ \\
 & 0.0002 & $-0.0127^{\ddagger}$ & $+0.0119^{\dagger}$ & $-0.0000$ & $+0.0006$ & $-6.7$ & $+10.7$ & $-4.0$ & $+0.7$ \\
 & 0.0005 & $-0.0119^{\ddagger}$ & $+0.0108^{\dagger}$ & $-0.0010$ & $+0.0010$ & $-5.5$ & $+9.3$ & $-4.8$ & $+0.8$ \\
 & 0.001 & $-0.0073^{\ddagger}$ & $+0.0090^{\dagger}$ & $+0.0043^{\dagger}$ & $+0.0024^{\dagger}$ & $-4.0$ & $+8.8$ & $-4.9$ & $+1.6$ \\
 & 0.002 & $-0.0033^{\ddagger}$ & $+0.0007$ & $+0.0094^{\dagger}$ & $-0.0008$ & $-2.7$ & $+7.3$ & $-4.4$ & $+2.2$ \\
 & 0.005 & $+0.0012$ & $-0.0142^{\ddagger}$ & $+0.0207^{\dagger}$ & $-0.0125^{\ddagger}$ & $-1.4$ & $+4.7$ & $-4.2$ & $+2.8$ \\
 & 0.01 & $+0.0085^{\dagger}$ & $-0.0269^{\ddagger}$ & $+0.0471^{\dagger}$ & $-0.0329^{\ddagger}$ & $-1.3$ & $+3.0$ & $-4.7$ & $+3.5$ \\
\bottomrule\end{tabular}}\end{table}
}

\newcommand{\LawScaleAtAtJTable}{%
\begin{table}[ht]\centering\scriptsize\setlength{\tabcolsep}{3pt}\renewcommand{\arraystretch}{0.8}
\caption{EI stopping (EI-S) under GP-atJ minus EI-S under GP-at, $U^{\text{GP-atJ}}_{\text{EI-S}}-U^{\text{GP-at}}_{\text{EI-S}}$, with both surrogates at the same settings: the primary settings, the noise variance scaled by $\frac14$ or $4$, the prior variances scaled by $\frac12$ or $2$, or a noise variance calibrated per task from the recorded curves of the other tasks of its benchmark. Paired per episode on the pools and seeds 1--50 of Table~\ref{tab:body-effects}; scores lie in $[0,1]$, and utility is in the same units. $^\dagger$/$^\ddagger$: simultaneous 95\% seed-block band above/below zero, over all 126 cells (critical value 3.60; Appendix~\ref{app:shared-uncertainty}). Every seed was observed at the primary settings before these runs; no verdict is attached.}
\label{tab:law-scale-at-atj}
\fitwidth{\begin{tabular}{@{}lrrrrrrr@{}}\toprule
Bench. & $\lambda$ & Primary & Noise $\times\frac14$ & Noise $\times4$ & Prior $\times\frac12$ & Prior $\times2$ & Calibrated noise \\ \midrule
LCBench & 0.0001 & $+0.0006^{\dagger}$ & $+0.0005$ & $+0.0001$ & $+0.0014^{\dagger}$ & $+0.0001$ & $+0.0003$ \\
 & 0.0002 & $+0.0013^{\dagger}$ & $+0.0009^{\dagger}$ & $+0.0006^{\dagger}$ & $+0.0008$ & $+0.0008^{\dagger}$ & $+0.0002$ \\
 & 0.0005 & $+0.0021^{\dagger}$ & $+0.0016^{\dagger}$ & $+0.0014^{\dagger}$ & $+0.0016^{\dagger}$ & $+0.0022^{\dagger}$ & $+0.0002$ \\
 & 0.001 & $+0.0018^{\dagger}$ & $+0.0007$ & $+0.0002$ & $+0.0023^{\dagger}$ & $+0.0019^{\dagger}$ & $-0.0001$ \\
 & 0.002 & $+0.0025^{\dagger}$ & $+0.0011^{\dagger}$ & $+0.0045^{\dagger}$ & $+0.0014$ & $+0.0007$ & $-0.0002$ \\
 & 0.005 & $+0.0029^{\dagger}$ & $+0.0049^{\dagger}$ & $+0.0008$ & $+0.0040^{\dagger}$ & $+0.0017$ & $+0.0047^{\dagger}$ \\
 & 0.01 & $+0.0024$ & $+0.0020$ & $+0.0017$ & $+0.0174^{\dagger}$ & $+0.0049^{\dagger}$ & $+0.0026$ \\
\midrule
PD1 & 0.0001 & $+0.0022$ & $+0.0004$ & $+0.0003$ & $+0.0008$ & $+0.0006$ & $+0.0002$ \\
 & 0.0002 & $+0.0008$ & $+0.0009$ & $-0.0000$ & $+0.0039^{\dagger}$ & $+0.0004^{\dagger}$ & $+0.0005$ \\
 & 0.0005 & $+0.0012^{\dagger}$ & $+0.0012$ & $+0.0006$ & $+0.0036^{\dagger}$ & $+0.0021^{\dagger}$ & $+0.0004$ \\
 & 0.001 & $+0.0032^{\dagger}$ & $+0.0007^{\dagger}$ & $+0.0039^{\dagger}$ & $+0.0015$ & $+0.0018^{\dagger}$ & $+0.0006$ \\
 & 0.002 & $-0.0019$ & $-0.0012$ & $-0.0002$ & $+0.0044^{\dagger}$ & $-0.0028^{\ddagger}$ & $-0.0030^{\ddagger}$ \\
 & 0.005 & $+0.0081^{\dagger}$ & $+0.0071^{\dagger}$ & $+0.0056^{\dagger}$ & $+0.0082^{\dagger}$ & $+0.0012^{\dagger}$ & $+0.0053^{\dagger}$ \\
 & 0.01 & $+0.0089^{\dagger}$ & $+0.0083^{\dagger}$ & $+0.0123^{\dagger}$ & $+0.0160^{\dagger}$ & $+0.0044^{\dagger}$ & $+0.0072^{\dagger}$ \\
\midrule
ODBench & 0.0001 & $+0.0003$ & $-0.0008$ & $+0.0006$ & $+0.0010$ & $-0.0007$ & $+0.0009$ \\
 & 0.0002 & $+0.0001$ & $+0.0020$ & $+0.0008$ & $+0.0030$ & $+0.0003$ & $+0.0012$ \\
 & 0.0005 & $-0.0004$ & $+0.0007$ & $+0.0025$ & $+0.0041^{\dagger}$ & $+0.0003$ & $+0.0015^{\dagger}$ \\
 & 0.001 & $+0.0025$ & $+0.0004$ & $+0.0024^{\dagger}$ & $+0.0042^{\dagger}$ & $+0.0003$ & $+0.0038^{\dagger}$ \\
 & 0.002 & $+0.0031^{\dagger}$ & $+0.0015$ & $+0.0038^{\dagger}$ & $+0.0087^{\dagger}$ & $-0.0007$ & $+0.0027^{\dagger}$ \\
 & 0.005 & $+0.0075^{\dagger}$ & $+0.0064^{\dagger}$ & $+0.0084^{\dagger}$ & $+0.0164^{\dagger}$ & $+0.0004$ & $+0.0083^{\dagger}$ \\
 & 0.01 & $+0.0178^{\dagger}$ & $+0.0197^{\dagger}$ & $+0.0157^{\dagger}$ & $+0.0338^{\dagger}$ & $+0.0018$ & $+0.0176^{\dagger}$ \\
\bottomrule\end{tabular}}\end{table}
}

\newcommand{\LawScaleAtAtJFinding}{%
At the primary settings GP-atJ has the higher mean utility in 19 of 21 cells, with bands above zero in 13 and below in 0. At the five other settings, in the order of the columns, the bands lie above zero in 9, 10, 15, 9 and 8 cells and below in 0, 0, 0, 1 and 1.
}

\newcommand{\InModelOraclePreregTable}{%
\begin{table}[t]
\centering\scriptsize\setlength{\tabcolsep}{3pt}
\caption{Pre-registered exact in-model gap of EI stopping under GP-atJ on 840 reduced instances drawn from seeds 1--50, 40 per benchmark and cost (Appendix~\ref{app:inmodel-oracle}). $3{\times}2$: the three arms with the largest EI, two further increments each; $2{\times}3$, on the same frozen states: the two arms with the largest EI, three further increments each. Optimal: share with $V_\lambda^\star=V_{\pi_{\rm EI}}$; within $\lambda$: share with $V_\lambda^\star-V_{\pi_{\rm EI}}<\lambda$, both with Wilson 95\% intervals; recovered: share of the total gap $\sum(V_\lambda^\star-V_{\pi_{\rm EI}})$ that ideal H2 stopping recovers, with a 95\% instance-bootstrap interval. Rows by benchmark are the pre-registered estimands; rows by cost pool the three benchmarks and are descriptive.}
\label{tab:inmodel-oracle-prereg}
\fitwidth{%
\begin{tabular}{lrrrrrrr}
\toprule
 & & \multicolumn{3}{c}{$3{\times}2$: 3 arms, 2 increments} & \multicolumn{3}{c}{$2{\times}3$: 2 arms, 3 increments} \\
\cmidrule(lr){3-5}\cmidrule(lr){6-8}
Instances & $n$ & Optimal [95\%] & Within $\lambda$ [95\%] & Recovered [95\%] & Optimal [95\%] & Within $\lambda$ [95\%] & Recovered [95\%] \\
\midrule
LCBench & 280 & 74.6\% [69, 79] & 95.4\% [92, 97] & 99.8\% [99.7, 100.0] & 60.4\% [55, 66] & 77.1\% [72, 82] & 85.0\% [77, 91] \\
PD1 & 280 & 71.8\% [66, 77] & 88.9\% [85, 92] & 99.6\% [99, 100] & 66.8\% [61, 72] & 79.3\% [74, 84] & 92.5\% [87, 97] \\
ODBench & 280 & 71.1\% [66, 76] & 85.0\% [80, 89] & 98.5\% [98, 99] & 55.4\% [50, 61] & 69.3\% [64, 74] & 81.4\% [74, 87] \\
\midrule
All & 840 & 72.5\% [69, 75] & 89.8\% [88, 92] & 99.2\% [98.9, 99.5] & 60.8\% [57, 64] & 75.2\% [72, 78] & 85.5\% [81, 89] \\
\midrule
$\lambda=0.0001$ & 120 & 74.2\% [66, 81] & 83.3\% [76, 89] & 99.3\% [99, 100] & 51.7\% [43, 60] & 56.7\% [48, 65] & 74.3\% [60, 84] \\
$\lambda=0.0002$ & 120 & 62.5\% [54, 71] & 79.2\% [71, 85] & 97.5\% [96, 99] & 51.7\% [43, 60] & 59.2\% [50, 68] & 93.2\% [89, 96] \\
$\lambda=0.0005$ & 120 & 75.0\% [67, 82] & 88.3\% [81, 93] & 99.3\% [98, 100] & 55.8\% [47, 64] & 70.0\% [61, 77] & 79.5\% [67, 88] \\
$\lambda=0.001$ & 120 & 59.2\% [50, 68] & 85.0\% [78, 90] & 99.2\% [99, 100] & 52.5\% [44, 61] & 68.3\% [60, 76] & 90.5\% [83, 96] \\
$\lambda=0.002$ & 120 & 68.3\% [60, 76] & 94.2\% [88, 97] & 99.3\% [99, 100] & 57.5\% [49, 66] & 80.8\% [73, 87] & 83.8\% [72, 93] \\
$\lambda=0.005$ & 120 & 82.5\% [75, 88] & 98.3\% [94, 100] & 99.5\% [99, 100] & 75.8\% [67, 83] & 91.7\% [85, 95] & 81.8\% [65, 94] \\
$\lambda=0.01$ & 120 & 85.8\% [78, 91] & 100.0\% [97, 100] & 100.0\% [100.0, 100.0] & 80.8\% [73, 87] & 100.0\% [97, 100] & 85.1\% [70, 95] \\
\bottomrule
\end{tabular}}
\end{table}
}

\newcommand{\InModelOraclePreregFinding}{%
On 840 pre-registered $3{\times}2$ instances over all seven costs, the EI stop is exactly optimal in 72.5\% and within one increment's cost in 89.8\%, and ideal H2 stopping, which uses the exact advantage, recovers 99.2\% of the gap (85.5\% on $2{\times}3$ instances).
}

\newcommand{\InModelOracleCostSplitFinding}{%
The gap concentrates at low cost: at the two highest costs the stop is optimal in 84\% and 78\% of $3{\times}2$ and $2{\times}3$ instances, and below them in 68\% and 54\%.
}

\newcommand{\ByLawNoninferiorityTable}{%
\begin{table}[ht]
\centering\scriptsize\setlength{\tabcolsep}{3pt}
\caption{Noninferiority of EI stopping (EI-S) to H2 stopping (H2-S) at $\delta=\lambda$ under every surrogate run with H2 stopping. Each cell gives the one-sided simultaneous 95\% upper confidence limit (UCL) of $\Delta_{\text{H2-S}}=\E[U_{\text{H2-S}}-U_{\text{EI-S}}]$, and its verdict (NI: UCL $<\lambda$; open: not established). All limits are proper one-sided limits from the fixed-SE studentized signed-maximum seed-block bootstrap of Appendix~\ref{app:shared-uncertainty} (GP-atJ: the 21-column H2-S-minus-EI-S family of the 50-seed extension; GP-at and GP-a: their 50-seed blocks, one 21-column family per surrogate, 10,000 replicates, seed 2026091101). Every row uses seeds 1--50 on the same pools, and every verdict here is descriptive: the pre-registered fresh-seed verdicts for GP-atJ are in Table~\ref{tab:body-effects}, the pre-registered ones for GP-at and the descriptive ones for GP-a in Table~\ref{tab:cross-law-fresh}. Scores lie in $[0,1]$, and the UCL is in the same units.}
\label{tab:noninferiority-by-law}
\fitwidth{%
\begin{tabular}{llrrrrrrr}
\toprule
 & & \multicolumn{7}{c}{$\lambda=\delta$} \\
\cmidrule(lr){3-9}
Surrogate & Bench. & 0.0001 & 0.0002 & 0.0005 & 0.001 & 0.002 & 0.005 & 0.01 \\
\midrule
GP-atJ, seeds 1--50 & LCBench & $+0.00035$ open & $+0.00034$ open & $+0.00034$ NI & $-0.00025$ NI & $-0.00005$ NI & $+0.00011$ NI & $+0.00015$ NI \\
 & PD1 & $+0.00220$ open & $+0.00505$ open & $+0.00182$ open & $+0.00245$ open & $+0.00293$ open & $+0.00291$ NI & $+0.00374$ NI \\
 & ODBench & $+0.00147$ open & $+0.00261$ open & $+0.00236$ open & $+0.00456$ open & $+0.00229$ open & $+0.00102$ NI & $+0.00040$ NI \\
\midrule
GP-at, seeds 1--50 & LCBench & $+0.00054$ open & $+0.00024$ open & $-0.00042$ NI & $+0.00013$ NI & $+0.00011$ NI & $+0.00033$ NI & $+0.00015$ NI \\
 & PD1 & $+0.00548$ open & $+0.00574$ open & $+0.00198$ open & $+0.00486$ open & $+0.00415$ open & $+0.00434$ NI & $+0.00346$ NI \\
 & ODBench & $+0.00174$ open & $+0.00081$ open & $+0.00147$ open & $+0.00598$ open & $+0.00241$ open & $+0.00203$ NI & $+0.00066$ NI \\
\midrule
GP-a, seeds 1--50 & LCBench & $+0.00048$ open & $+0.00017$ NI & $-0.00023$ NI & $+0.00011$ NI & $+0.00011$ NI & $+0.00022$ NI & $-0.00005$ NI \\
 & PD1 & $+0.00467$ open & $+0.00568$ open & $+0.00231$ open & $+0.00315$ open & $+0.00518$ open & $+0.00462$ NI & $+0.00308$ NI \\
 & ODBench & $+0.00156$ open & $+0.00088$ open & $+0.00571$ open & $+0.00417$ open & $+0.00157$ NI & $+0.00101$ NI & $+0.00125$ NI \\
\bottomrule
\end{tabular}}
\end{table}
}

\newcommand{\EISurrogateThreeWayTable}{%
\begin{table*}[t]\centering\scriptsize\setlength{\tabcolsep}{3pt}\renewcommand{\arraystretch}{0.85}
\caption{EI stopping (EI-S) under the three surrogates, on the pools of Table~\ref{tab:body-effects}: its mean utility under GP-atJ, and EI-S under GP-at and under GP-a minus EI-S under GP-atJ, paired by seed, on the development seeds 1--50 and on the fresh seeds 51--100. Scores lie in $[0,1]$, and utility is in the same units. [band]: two-sided simultaneous 95\% band over the 21 cells of each contrast and seed set (Appendix~\ref{app:shared-uncertainty}; $^{\dagger}$/$^{\ddagger}$: above/below zero, so GP-atJ behind/ahead). Descriptive: no hypothesis was declared.}
\label{tab:ei-surrogate-three-way}
\fitwidth{\begin{tabular}{@{}lrrllrll@{}}\toprule
 & & \multicolumn{3}{c}{Seeds 1--50} & \multicolumn{3}{c}{Fresh seeds 51--100} \\ \cmidrule(lr){3-5}\cmidrule(lr){6-8}
Bench. & $\lambda$ & $U^{\text{GP-atJ}}_{\text{EI-S}}$ & GP-at $-$ GP-atJ [band] & GP-a $-$ GP-atJ [band] & $U^{\text{GP-atJ}}_{\text{EI-S}}$ & GP-at $-$ GP-atJ [band] & GP-a $-$ GP-atJ [band] \\ \midrule
LCBench & 0.0001 & $0.7627$ & $-0.0006$ [-0.0011, -0.0002]$^{\ddagger}$ & $-0.0009$ [-0.0014, -0.0004]$^{\ddagger}$ & $0.7642$ & $-0.0004$ [-0.0007, -0.0000]$^{\ddagger}$ & $-0.0005$ [-0.0008, -0.0002]$^{\ddagger}$ \\
 & 0.0002 & $0.7544$ & $-0.0013$ [-0.0017, -0.0008]$^{\ddagger}$ & $-0.0014$ [-0.0019, -0.0009]$^{\ddagger}$ & $0.7559$ & $-0.0008$ [-0.0012, -0.0005]$^{\ddagger}$ & $-0.0010$ [-0.0014, -0.0007]$^{\ddagger}$ \\
 & 0.0005 & $0.7372$ & $-0.0021$ [-0.0030, -0.0012]$^{\ddagger}$ & $-0.0019$ [-0.0029, -0.0010]$^{\ddagger}$ & $0.7389$ & $-0.0022$ [-0.0035, -0.0010]$^{\ddagger}$ & $-0.0019$ [-0.0030, -0.0007]$^{\ddagger}$ \\
 & 0.001 & $0.7197$ & $-0.0018$ [-0.0026, -0.0009]$^{\ddagger}$ & $-0.0026$ [-0.0036, -0.0016]$^{\ddagger}$ & $0.7212$ & $-0.0014$ [-0.0023, -0.0005]$^{\ddagger}$ & $-0.0018$ [-0.0029, -0.0008]$^{\ddagger}$ \\
 & 0.002 & $0.6939$ & $-0.0025$ [-0.0035, -0.0015]$^{\ddagger}$ & $-0.0036$ [-0.0047, -0.0025]$^{\ddagger}$ & $0.6972$ & $-0.0024$ [-0.0033, -0.0015]$^{\ddagger}$ & $-0.0032$ [-0.0042, -0.0023]$^{\ddagger}$ \\
 & 0.005 & $0.6215$ & $-0.0029$ [-0.0048, -0.0010]$^{\ddagger}$ & $-0.0031$ [-0.0051, -0.0010]$^{\ddagger}$ & $0.6267$ & $-0.0027$ [-0.0047, -0.0008]$^{\ddagger}$ & $-0.0030$ [-0.0050, -0.0009]$^{\ddagger}$ \\
 & 0.01 & $0.5356$ & $-0.0024$ [-0.0062, +0.0015] & $-0.0033$ [-0.0077, +0.0011] & $0.5443$ & $-0.0002$ [-0.0039, +0.0035] & $-0.0007$ [-0.0050, +0.0037] \\
\midrule
PD1 & 0.0001 & $0.7616$ & $-0.0022$ [-0.0043, -0.0000]$^{\ddagger}$ & $-0.0035$ [-0.0055, -0.0016]$^{\ddagger}$ & $0.7641$ & $-0.0028$ [-0.0051, -0.0004]$^{\ddagger}$ & $-0.0060$ [-0.0085, -0.0035]$^{\ddagger}$ \\
 & 0.0002 & $0.7451$ & $-0.0008$ [-0.0014, -0.0001]$^{\ddagger}$ & $-0.0015$ [-0.0022, -0.0008]$^{\ddagger}$ & $0.7464$ & $-0.0003$ [-0.0016, +0.0010] & $-0.0022$ [-0.0043, -0.0002]$^{\ddagger}$ \\
 & 0.0005 & $0.7262$ & $-0.0012$ [-0.0021, -0.0004]$^{\ddagger}$ & $-0.0025$ [-0.0037, -0.0012]$^{\ddagger}$ & $0.7263$ & $-0.0015$ [-0.0029, -0.0002]$^{\ddagger}$ & $-0.0024$ [-0.0048, -0.0000]$^{\ddagger}$ \\
 & 0.001 & $0.7043$ & $-0.0032$ [-0.0053, -0.0011]$^{\ddagger}$ & $-0.0044$ [-0.0063, -0.0025]$^{\ddagger}$ & $0.7030$ & $-0.0028$ [-0.0049, -0.0008]$^{\ddagger}$ & $-0.0049$ [-0.0073, -0.0024]$^{\ddagger}$ \\
 & 0.002 & $0.6776$ & $+0.0019$ [-0.0006, +0.0044] & $+0.0017$ [-0.0017, +0.0050] & $0.6769$ & $+0.0024$ [-0.0003, +0.0051] & $+0.0015$ [-0.0017, +0.0048] \\
 & 0.005 & $0.6057$ & $-0.0081$ [-0.0108, -0.0054]$^{\ddagger}$ & $-0.0084$ [-0.0112, -0.0056]$^{\ddagger}$ & $0.6029$ & $-0.0069$ [-0.0091, -0.0047]$^{\ddagger}$ & $-0.0086$ [-0.0116, -0.0057]$^{\ddagger}$ \\
 & 0.01 & $0.4990$ & $-0.0089$ [-0.0112, -0.0066]$^{\ddagger}$ & $-0.0116$ [-0.0145, -0.0087]$^{\ddagger}$ & $0.4966$ & $-0.0103$ [-0.0131, -0.0075]$^{\ddagger}$ & $-0.0129$ [-0.0166, -0.0093]$^{\ddagger}$ \\
\midrule
ODBench & 0.0001 & $0.3605$ & $-0.0003$ [-0.0020, +0.0014] & $-0.0017$ [-0.0038, +0.0004] & $0.3570$ & $+0.0003$ [-0.0015, +0.0020] & $-0.0013$ [-0.0029, +0.0003] \\
 & 0.0002 & $0.3532$ & $-0.0001$ [-0.0027, +0.0024] & $-0.0017$ [-0.0047, +0.0012] & $0.3500$ & $-0.0001$ [-0.0018, +0.0015] & $-0.0024$ [-0.0045, -0.0003]$^{\ddagger}$ \\
 & 0.0005 & $0.3340$ & $+0.0004$ [-0.0016, +0.0024] & $-0.0043$ [-0.0091, +0.0006] & $0.3315$ & $-0.0002$ [-0.0032, +0.0027] & $-0.0022$ [-0.0052, +0.0008] \\
 & 0.001 & $0.3057$ & $-0.0025$ [-0.0061, +0.0010] & $-0.0046$ [-0.0081, -0.0012]$^{\ddagger}$ & $0.3056$ & $-0.0025$ [-0.0045, -0.0005]$^{\ddagger}$ & $-0.0053$ [-0.0087, -0.0018]$^{\ddagger}$ \\
 & 0.002 & $0.2606$ & $-0.0031$ [-0.0056, -0.0006]$^{\ddagger}$ & $-0.0083$ [-0.0112, -0.0054]$^{\ddagger}$ & $0.2598$ & $-0.0024$ [-0.0053, +0.0005] & $-0.0067$ [-0.0100, -0.0034]$^{\ddagger}$ \\
 & 0.005 & $0.1427$ & $-0.0075$ [-0.0104, -0.0046]$^{\ddagger}$ & $-0.0166$ [-0.0211, -0.0122]$^{\ddagger}$ & $0.1407$ & $-0.0070$ [-0.0098, -0.0042]$^{\ddagger}$ & $-0.0140$ [-0.0180, -0.0100]$^{\ddagger}$ \\
 & 0.01 & $-0.0330$ & $-0.0178$ [-0.0225, -0.0131]$^{\ddagger}$ & $-0.0380$ [-0.0449, -0.0310]$^{\ddagger}$ & $-0.0398$ & $-0.0184$ [-0.0232, -0.0136]$^{\ddagger}$ & $-0.0379$ [-0.0449, -0.0308]$^{\ddagger}$ \\
\bottomrule\end{tabular}}\end{table*}
}
\newcommand{\EISurrogateThreeWayFinding}{%
By the simultaneous bands, EI stopping under GP-atJ attains higher utility than under GP-at in fifteen of 21 cells on the development seeds and fourteen on the fresh seeds, and is never lower; against GP-a it is higher in sixteen and seventeen cells and never lower. At the two lowest costs the three surrogates differ by at most $0.0060$.%
}
\newcommand{\EISharingTestsTable}{%
\begin{table}[ht]\centering\scriptsize\setlength{\tabcolsep}{3pt}
\caption{EI stopping's utility at each step of added transient sharing (F3). Mean: the step's EI-S utility difference averaged over the 21 cells and the 50 seed blocks; $p$: one-sample $t$-test of the 50 per-seed-block averages against zero (Wilcoxon signed-rank in brackets); Pos.: seed blocks with a positive average; Above/Below: cells whose two-sided simultaneous 95\% band, over one joint family of all 63 cells of the seed set (critical values 3.32 and 3.34), lies above/below zero.}
\label{tab:ei-sharing-tests}
\begin{tabular}{@{}llrrrrr@{}}\toprule
Seeds & Step & Mean & $p$ [Wilcoxon] & Pos. & Above & Below \\ \midrule
1--50 & GP-at $-$ GP-a & $+0.0026$ & $<10^{-16}$ [$<10^{-13}$] & 49/50 & 9 & 0 \\
 & GP-atJ $-$ GP-at & $+0.0032$ & $<10^{-15}$ [$<10^{-11}$] & 47/50 & 14 & 0 \\
 & GP-atJ $-$ GP-a & $+0.0058$ & $<10^{-21}$ [$<10^{-14}$] & 49/50 & 16 & 0 \\
\midrule
51--100 & GP-at $-$ GP-a & $+0.0026$ & $<10^{-14}$ [$<10^{-13}$] & 47/50 & 8 & 0 \\
 & GP-atJ $-$ GP-at & $+0.0030$ & $<10^{-16}$ [$<10^{-11}$] & 48/50 & 13 & 0 \\
 & GP-atJ $-$ GP-a & $+0.0056$ & $<10^{-23}$ [$<10^{-14}$] & 49/50 & 15 & 0 \\
\bottomrule\end{tabular}\end{table}
}
\newcommand{\EISharingFinding}{%
Each step from GP-a to GP-at to GP-atJ raises EI stopping's utility averaged over the 21 cells, significantly on both seed sets ($t$-test, $p<10^{-14}$); cell by cell, under one simultaneous 95\% family per seed set, the gain is significant in 8 to 14 of 21 cells, and no cell is significantly worse (Table~\ref{tab:ei-sharing-tests}).%
}
\newcommand{\EISharingFindingFresh}{%
Each step from GP-a to GP-at to GP-atJ raises EI stopping's utility averaged over the 21 cells ($p<10^{-14}$), significantly in 8 and 13 of 21 cells individually, and no cell is significantly worse (Table~\ref{tab:ei-sharing-tests}).%
}

\newcommand{\FullScaleBoundsTable}{%
\begin{table}[ht]\centering\scriptsize\setlength{\tabcolsep}{3.5pt}\renewcommand{\arraystretch}{0.9}
\caption{Full-scale bounds on the gap $V^\star_\lambda-V_{\pi_{\rm EI}}$ at the 840 pre-registered first EI stops of Table~\ref{tab:inmodel-oracle-prereg} under GP-atJ, with all 32 arms and the recorded horizons (40 states per cell), and the long-lookahead policy $\pi_{\rm LA}$ replayed on the recorded curves. Utility in units of $10^{-3}$. Upper: information-relaxation bound under the planning law, and its Gaussian part. Lower: value of $\pi_{\rm LA}$ under the planning law, with its one-sided 95\% Monte Carlo limit. Verdict: premature if that limit exceeds $\lambda$, near-optimal if the upper bound is below $\lambda$, open otherwise. Realized: $U_{\pi_{\rm LA}}-U_{\text{EI-S}}$ on the recorded curves from the same states (seeds 1--50) and from 840 fresh first EI stops drawn the same way (seeds 51--100, whose value under the planning law is listed), with one-sided 95\% task-clustered bootstrap limits; $^\dagger$/$^\ddagger$: lower limit above zero / upper limit below zero; $^*$: also after Holm over the 14 premature cells, the pre-registered family.}
\label{tab:full-scale-bounds}
\fitwidth{\begin{tabular}{@{}llrrrrl rr rrr@{}}\toprule
 & & \multicolumn{5}{c}{Bounds under the planning law} & \multicolumn{2}{c}{Seeds 1--50} & \multicolumn{3}{c}{Seeds 51--100} \\
\cmidrule(lr){3-7}\cmidrule(lr){8-9}\cmidrule(l){10-12}
Bench. & $\lambda$ & Upper & Gaussian part & Lower & [limit] & Verdict & Realized & [limits] & Planning law & Realized & [limits] \\ \midrule
LCBench & 0.0001 & 1{,}509 & 36.9 & 18.4 & $\ge$17.9 & premature & +5.72 & [-0.460, +14.9] & 9.26 & +0.232 & [-0.657, +1.50] \\
 & 0.0002 & 2{,}708 & 30.5 & 11.7 & $\ge$11.2 & premature & +0.613 & [-0.971, +2.64] & 13.4 & +8.90$^\dagger$ & [+1.11, +18.0] \\
 & 0.0005 & 1{,}282 & 34.5 & 7.75 & $\ge$7.47 & premature & +3.41 & [-0.0253, +8.63] & 2.86 & -0.750$^\ddagger$$^*$ & [-1.35, -0.298] \\
 & 0.001 & 1{,}538 & 29.8 & 6.86 & $\ge$6.59 & premature & -1.01 & [-3.10, +0.228] & 7.29 & +1.02 & [-0.854, +3.49] \\
 & 0.002 & 502 & 33.4 & 1.39 & $\ge$1.19 & open & +0.0102 & [-0.410, +0.546] & 3.52 & -0.700$^\ddagger$ & [-1.38, -0.166] \\
 & 0.005 & 2{,}730 & 38.4 & 0.872 & $\ge$0.725 & open & -0.250 & [-0.732, 0.00] & 0.153 & -0.125 & [-0.366, 0.00] \\
 & 0.01 & 2{,}314 & 48.9 & 0 & $\ge$0 & open & 0.00 & [0.00, 0.00] & 0.110 & -0.0597 & [-0.175, 0.00] \\
\midrule
PD1 & 0.0001 & 1{,}182 & 47.1 & 26.6 & $\ge$26.2 & premature & +16.1 & [-2.09, +36.9] & 17.7 & +19.0 & [-2.23, +48.3] \\
 & 0.0002 & 1{,}332 & 60.9 & 27.1 & $\ge$26.8 & premature & +25.7$^\dagger$ & [+3.77, +55.0] & 20.0 & +1.07 & [-5.38, +10.5] \\
 & 0.0005 & 1{,}311 & 49.3 & 17.7 & $\ge$17.4 & premature & +12.7 & [-1.18, +30.1] & 28.3 & +9.55 & [-4.62, +25.8] \\
 & 0.001 & 1{,}071 & 61.1 & 15.4 & $\ge$15.1 & premature & +2.56 & [-4.21, +11.7] & 13.7 & +6.09 & [-8.39, +26.6] \\
 & 0.002 & 1{,}649 & 56.2 & 4.09 & $\ge$3.90 & premature & +5.76 & [-1.11, +17.5] & 5.44 & +1.57 & [-1.35, +6.29] \\
 & 0.005 & 1{,}927 & 76.2 & 2.83 & $\ge$2.55 & open & -0.361 & [-1.87, +0.890] & 1.85 & -0.384 & [-0.905, 0.00] \\
 & 0.01 & 3{,}634 & 74.4 & 0.616 & $\ge$0.556 & open & +4.56 & [0.00, +13.0] & 2.39 & +2.26 & [0.00, +4.99] \\
\midrule
ODBench & 0.0001 & 27{,}441 & 49.0 & 45.4 & $\ge$44.9 & premature & +4.46$^\dagger$ & [+0.459, +8.99] & 46.5 & +12.9$^\dagger$$^*$ & [+4.66, +22.0] \\
 & 0.0002 & 22{,}086 & 59.9 & 50.6 & $\ge$50.0 & premature & +9.29$^\dagger$ & [+2.52, +14.5] & 45.2 & +4.58 & [-0.882, +10.3] \\
 & 0.0005 & 25{,}037 & 54.3 & 36.1 & $\ge$35.6 & premature & +0.440 & [-5.65, +8.85] & 36.3 & -1.16 & [-7.33, +3.73] \\
 & 0.001 & 24{,}661 & 57.2 & 32.6 & $\ge$32.1 & premature & +2.04 & [-7.29, +11.9] & 22.4 & -4.35 & [-11.9, +1.26] \\
 & 0.002 & 31{,}924 & 44.6 & 12.6 & $\ge$12.2 & premature & +0.107 & [-5.86, +6.33] & 13.3 & -3.65 & [-9.95, +0.600] \\
 & 0.005 & 24{,}404 & 31.9 & 2.52 & $\ge$2.29 & open & -0.360 & [-0.833, +0.0167] & 1.99 & +0.355 & [-0.714, +1.69] \\
 & 0.01 & 27{,}257 & 19.4 & 0.559 & $\ge$0.492 & open & +1.52 & [0.00, +4.79] & 0.553 & -0.367 & [-1.05, +0.164] \\
\bottomrule\end{tabular}}
\end{table}
}

\newcommand{\GateCalibrationTable}{%
\begin{table}[ht]
\centering\small
\caption{Calibration of the gate on the recorded curves: H2 stopping under GP-atJ on seeds 1--50, over every episode in which the gate opened. The certified lower bound is the lower enclosure of the selected arm's depth-two advantage where the gate first opened, which under the planning law bounds the expected gain of H2 stopping over EI stopping from that state. The realized gain is $U_{\text{H2-S}}-U_{\text{EI-S}}$ of the same episode. Means in units of $10^{-3}$, pooled over the seven costs; the band is a 95\% seed-block bootstrap band for realized minus certified.}
\label{tab:gate-calibration}
\begin{tabular}{lrrrrr}
\toprule
Benchmark & Gate opened & Certified lower & Realized & Realized $-$ certified [band] & Realized $<0$ \\
\midrule
LCBench & 2,060 & +0.11 & -1.71 & -1.82 [-2.60, -1.13] & 94\% \\
PD1 & 1,199 & +0.29 & +6.50 & +6.21 [+4.99, +7.40] & 78\% \\
ODBench & 683 & +0.46 & +3.28 & +2.81 [+0.95, +4.77] & 83\% \\
\bottomrule
\end{tabular}
\end{table}
}
\newcommand{\GateCalibrationFinding}{%
The realized gain falls below the certified lower bound on LCBench, with a band below zero, and exceeds it on PD1 and ODBench, with bands above zero.
}

\newcommand{\FindingsFigure}{%
\begin{figure}[t]
\centering
\includegraphics[width=\textwidth]{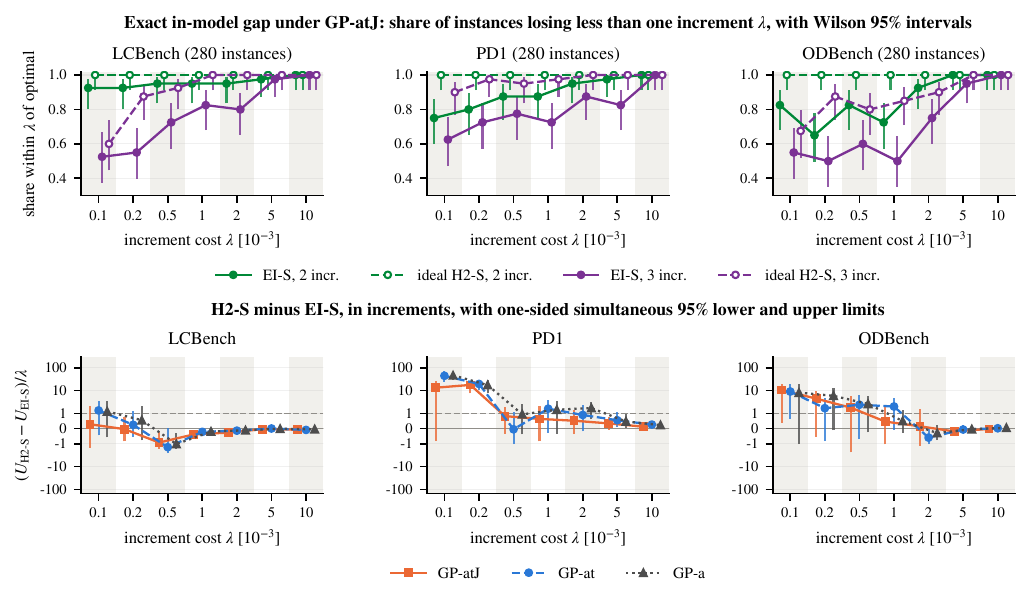}
\caption{Top: share of the 840 pre-registered reduced instances (Table~\ref{tab:inmodel-oracle-prereg}; 40 per cell) whose EI stop (EI-S; filled, solid) and ideal H2 stopping (open, dashed) lose less than $\lambda$ under GP-atJ, with three arms and two further increments per arm (green), or two arms and three increments per arm (purple); Wilson 95\% intervals. Bottom: $(U_{\text{H2-S}}-U_{\text{EI-S}})/\lambda$ for H2 stopping (H2-S) under each surrogate, (Tables~\ref{tab:body-effects} and~\ref{tab:cross-law-fresh}); markers are point estimates, whiskers span the one-sided simultaneous 95\% limits. A UCL below the dashed line at one means EI stopping is noninferior at $\delta=\lambda$; a lower limit above zero means H2-S gains, a UCL below zero that it loses. GP-atJ lower limits come from a two-sided band and are conservative (Appendix~\ref{app:shared-uncertainty}). Gate shares: Table~\ref{tab:body-effects}.}
\label{fig:findings}
\end{figure}
}

{\spanfigures\FindingsFigure}
\section{Controlled Shared Learning-Curve Models}
\label{sec:shared-models}

Our controlled family of learning-curve surrogates uses the finite-basis Gaussian model
\begin{equation*}
Y_i(t)=\tfrac12+a_i+\sum_{r=1}^{K} b_{ir}\phi_r(t)+\epsilon_i(t),\qquad\phi_r(t)=\frac{s_r+1}{s_r+t},\quad s_1<\cdots<s_K.
\end{equation*}
Here $a_i$ fixes where arm $i$'s curve settles and $b_{i1},\ldots,b_{iK}$ how it climbs on $K$
fixed timescales, with observation noise $\epsilon_i(t)$; each $\phi_r$ equals one at the first checkpoint and decays on timescale $s_r$.
All cross-configuration sharing is encoded in two matrices,
\[
\operatorname{Cov}(a_i,a_j)=v_a\,C_{a,ij},\qquad
\operatorname{Cov}(b_{ir},b_{jr})=\frac{v_t}{K}\,C_{t,ij},
\]
with independent blocks across $r$ and $\operatorname{Cov}(a_i,b_{jr})=0$; at the first checkpoint the level thus holds variance $v_a$ and the transient $v_t$. Let $R$ be the
unit-diagonal Mat\'ern-$5/2$ hyperparameter (HP) kernel~\citep{rasmussen2006gpml} and $J$ the all-ones matrix. Each name lists what is coupled. All three couple asymptotes (``a''), $C_a=(I+R)/2$. GP-a keeps transients independent, $C_t=I$. GP-at also couples
transients (``t'') through the HP kernel, $C_t=(I+R)/2$, and the primary GP-atJ adds a task-wide transient (``J''),
$C_t=I/2+R/4+J/4$. Since every $C_a,C_t$ has unit diagonal, all three share the same-arm prior covariances; they differ only in $C_t$.

\paragraph{Why Couple Transients.}
An asymptote is constant in $t$ and thus cannot enter arm $i$'s next increment
$\xi_i=\sum_r b_{ir}(\phi_r(n+1)-\phi_r(n))$ after $n$ checkpoints. For $j\ne i$ and every $t'$,
\[
\operatorname{Cov}(\xi_i,Y_j(t'))=\frac{v_t}{K}\,C_{t,ij}\sum_{r=1}^K\bigl(\phi_r(n+1)-\phi_r(n)\bigr)\phi_r(t'),
\]
with every summand negative. The increment is thus correlated with every other arm's scores under
GP-at and GP-atJ, where $C_{t,ij}>0$, and with none under asymptote-only sharing, as in GP-a, where other arms reach an arm's next score only through a scalar
loading on its asymptote that shrinks as the arm is observed (Appendix~\ref{app:asymptote-attenuation}).

\section{Experiments}
\label{sec:experiments}

\subsection{Setup and Estimands}
\label{sec:model-controller-design}

\paragraph{Benchmarks.}
We evaluate all 35 LCBench tasks~\citep{zimmer2021autopytorch}, 15 PD1 tasks~\citep{wang2024hyperbo} (Appendix~\ref{app:pd1-results}) and nine ODBench object-detection tasks~\citep{cfbo2025,ciaglia2022roboflow}, held out from model development, at seven costs $\lambda\in\{0.0001,\allowbreak 0.0002,\allowbreak 0.0005,\allowbreak 0.001,\allowbreak 0.002,\allowbreak 0.005,\allowbreak 0.01\}$, with $N=32$ candidate
configurations and 50 paired seeds per task and cost. An increment is one native checkpoint (one epoch on LCBench), a unit that differs
across benchmarks; results are reported per benchmark. The costs charge $0.01$ to $1$ percentage point of score per increment.

\paragraph{Surrogates.}
We run the family of Section~\ref{sec:shared-models} with $K=4$, $(s_1,\ldots,s_4)=(1,8,64,512)$ and
$v_a=v_t=1/32$; the timescales are geometric so that one basis spans curves of one to
several hundred checkpoints. Appendices~\ref{app:shared-temporal-b}
and~\ref{app:fixed-law-sensitivity} give the parameters and EI stopping's sensitivity to them. F1 and F2 use GP-atJ, under which EI stopping attains the highest utility of the three (F3), F3 adds GP-a and GP-at, and F4
uses the surrogates of two published methods.

\paragraph{Estimands.}
An episode's utility is $U=\tilde y_\tau-\lambda\tau$. Method $A$ minus method $B$ on one task, seed and pool is $U_A-U_B=\Delta\tilde y-\lambda\Delta\tau$, the score gain less the cost of extra increments. Comparisons
average paired differences equally over tasks and carry simultaneous 95\% bands (Appendix~\ref{app:shared-uncertainty}). Since H2 stopping (Algorithm~\ref{alg:selective-receding}) takes the actions of EI stopping until its gate first opens, $U_{\text{EI-S}}$ is read from the same episode, stopped at its first EI stop, and the two differ only where the gate opens. The expected gain $\Delta_{\text{H2-S}}=\E[U_{\text{H2-S}}-U_{\text{EI-S}}]$ averages over all episodes, including those where the gate stays closed, and over seeds and pools on the recorded curves.

EI stopping is \emph{noninferior}~\citep{wellek2010equivalence} in a cell when
the one-sided simultaneous 95\% upper confidence limit (UCL) of $\Delta_{\text{H2-S}}$ over the 21 benchmark--cost cells lies
below the tolerance $\delta=\lambda$, one increment's cost: a stop is acceptable when what it forgoes is worth less than
the one increment it declines to train. Otherwise the question is open
(Appendix~\ref{app:primary50}). The noninferiority test and F1's estimands are pre-registered.
\begin{table*}[t]\centering\small\setlength{\tabcolsep}{3pt}\renewcommand{\arraystretch}{0.9}
\caption{H2 stopping (H2-S) minus EI stopping (EI-S) under GP-atJ in the pre-registered test (Appendix~\ref{app:primary50}; development seeds: Table~\ref{tab:body-effects-dev}). Gate: share of episodes in which the gate opens. H2-S $-$ EI-S: mean difference; $^\dagger$/$^\ddagger$: its two-sided simultaneous 95\% band over the 21 cells lies above/below zero (Table~\ref{tab:body-effects-dev}); UCL: one-sided simultaneous 95\% upper confidence limit of $\E[U^{\text{GP-atJ}}_{\text{H2-S}}-U^{\text{GP-atJ}}_{\text{EI-S}}]$. Verdict at $\delta=\lambda$: NI, noninferior (UCL $<\delta$); open, not established; other surrogates in Table~\ref{tab:cross-law-fresh}.}
\label{tab:body-effects}
\fitwidth{\begin{tabular}{@{}r|rrrl|rrrl|rrrl@{}}\toprule
 & \multicolumn{4}{c|}{LCBench} & \multicolumn{4}{c|}{PD1} & \multicolumn{4}{c}{ODBench} \\
$\lambda=\delta$ & Gate & H2-S $-$ EI-S & UCL & Verdict & Gate & H2-S $-$ EI-S & UCL & Verdict & Gate & H2-S $-$ EI-S & UCL & Verdict \\ \midrule
0.0001 & 35\% & $+0.00003$ & $+0.00020$ & open & 40\% & $+0.00135$ & $+0.00273$ & open & 23\% & $+0.00106^{\dagger}$ & $+0.00205$ & open \\
0.0002 & 27\% & $-0.00001$ & $+0.00015$ & NI & 43\% & $+0.00358^{\dagger}$ & $+0.00558$ & open & 19\% & $+0.00089$ & $+0.00186$ & open \\
0.0005 & 22\% & $-0.00046^{\ddagger}$ & $-0.00010$ & NI & 31\% & $+0.00041$ & $+0.00097$ & open & 19\% & $+0.00090$ & $+0.00291$ & open \\
0.001 & 17\% & $-0.00038^{\ddagger}$ & $-0.00018$ & NI & 20\% & $+0.00064$ & $+0.00206$ & open & 29\% & $+0.00046$ & $+0.00193$ & open \\
0.002 & 13\% & $-0.00052^{\ddagger}$ & $-0.00027$ & NI & 10\% & $+0.00104$ & $+0.00274$ & open & 24\% & $+0.00031$ & $+0.00299$ & open \\
0.005 & 4\% & $-0.00016$ & $+0.00002$ & NI & 9\% & $+0.00172$ & $+0.00373$ & NI & 14\% & $-0.00093$ & $+0.00043$ & NI \\
0.01 & 4\% & $-0.00071$ & $+0.00023$ & NI & 7\% & $+0.00129$ & $+0.00268$ & NI & 5\% & $-0.00018$ & $+0.00021$ & NI \\
\bottomrule\end{tabular}}\end{table*}

\newcommand{\BControllerFindingShort}{%
The gate opens at the first EI stop in 4--29\% of episodes at the four highest costs and in 19--43\% at the three lower ones. It changes mean utility by less than one increment's cost at $\lambda\ge0.001$, and below that by up to 18 times that cost on PD1 and 11 times on ODBench.
}

\newcommand{\BNoninferiorityByLawFinding}{%
On seeds 1--50, EI stopping is noninferior to H2 stopping in 9, 9 and 11 of 21 cells under GP-atJ, GP-at and GP-a, including every cell at $\lambda\ge0.005$; these counts are descriptive. Under all three, the mean UCL at the two highest costs is 0.3 to 1.1 times its mean at the two lowest, and the UCLs stay below $0.0006$ on LCBench and below $0.006$ on PD1 and ODBench.
}
\newcommand{\BLimitRatioSentence}{%
the ratio of the mean limit at the two highest costs to its mean at the two lowest is $0.4$, $0.9$, $0.3$ under GP-atJ, $0.6$, $0.7$, $1.1$ under GP-at and $0.3$, $0.7$, $0.9$ under GP-a on LCBench, PD1 and ODBench
}
\newcommand{\BFreshBandSigns}{%
Certified continuation gains significantly in two cells (PD1 at $\lambda=0.0002$; ODBench at $\lambda=0.0001$) and loses in three (LCBench at $\lambda=0.0005$, $0.001$ and $0.002$).
}

\newcommand{\BCrossLawFreshFinding}{%
EI stopping is noninferior in 9 of 21 cells under GP-at and in 10 under GP-a (Table~\ref{tab:cross-law-fresh}).%
}
\newcommand{\CrossLawFreshParagraph}{%
Table~\ref{tab:cross-law-fresh} reports the fresh-seed limits and verdicts. Under GP-at, whose test was pre-registered, EI stopping is noninferior in 9 of 21 cells (5 on LCBench, 1 on PD1, 3 on ODBench), and 19 of the 21 verdicts repeat those of seeds 1--50 (Table~\ref{tab:noninferiority-by-law}). The registered ratio criterion holds: the ratio of the mean limit at the two highest costs to its mean at the two lowest is $0.03$, $0.7$, $0.6$ on LCBench, PD1 and ODBench, against the declared threshold of at most $2$. At the two highest costs noninferiority is shown in five of the six GP-at cells, PD1 at $\lambda=0.005$ staying open, and at the lowest cost in no cell. The sign hypothesis H3 ties the limits to Corollary~\ref{cor:receding-policy-improvement}, under which the expected in-model contrast is nonnegative; it held on PD1 and not on LCBench, where the mean is above zero at $\lambda=0.005$. Under GP-a, with the same statistic and its own bootstrap seed (2026102901), the verdicts are descriptive: EI stopping is noninferior in 10 of 21 cells (5 on LCBench, 2 on PD1, 3 on ODBench), 20 of the 21 verdicts repeat those of seeds 1--50, the ratio is $0.3$, $0.7$, $0.6$ on LCBench, PD1 and ODBench, noninferiority is shown in every cell at the two highest costs, and at the lowest cost in no cell.%
}
\newcommand{\CrossLawFreshTable}{%
\begin{table}[ht]
\centering\scriptsize\setlength{\tabcolsep}{3pt}
\caption{Noninferiority of EI stopping (EI-S) to H2 stopping (H2-S) at $\delta=\lambda$ under GP-at (pre-registered) and GP-a (descriptive) on the fresh seeds 51--100, on the pools of Table~\ref{tab:body-effects}. Each cell gives the one-sided simultaneous 95\% upper confidence limit (UCL) of $\Delta_{\text{H2-S}}=\E[U_{\text{H2-S}}-U_{\text{EI-S}}]$ from the bootstrap of Appendix~\ref{app:shared-uncertainty}, one 21-column family per surrogate (10,000 replicates; seeds 2026100501 and 2026102901), and its verdict (NI: UCL $<\lambda$; open: not established). Scores lie in $[0,1]$, and the UCL is in the same units.}
\label{tab:cross-law-fresh}
\fitwidth{%
\begin{tabular}{llrrrrrrr}
\toprule
 & & \multicolumn{7}{c}{$\lambda=\delta$} \\
\cmidrule(lr){3-9}
Surrogate & Bench. & 0.0001 & 0.0002 & 0.0005 & 0.001 & 0.002 & 0.005 & 0.01 \\
\midrule
GP-at, seeds 51--100 & LCBench & $+0.00036$ open & $+0.00025$ open & $+0.00002$ NI & $-0.00000$ NI & $-0.00013$ NI & $+0.00032$ NI & $-0.00030$ NI \\
 & PD1 & $+0.00701$ open & $+0.00618$ open & $+0.00060$ open & $+0.00396$ open & $+0.00452$ open & $+0.00556$ open & $+0.00422$ NI \\
 & ODBench & $+0.00198$ open & $+0.00097$ open & $+0.00314$ open & $+0.00462$ open & $-0.00007$ NI & $+0.00115$ NI & $+0.00075$ NI \\
\midrule
GP-a, seeds 51--100 & LCBench & $+0.00033$ open & $+0.00043$ open & $-0.00016$ NI & $+0.00007$ NI & $-0.00004$ NI & $+0.00017$ NI & $+0.00007$ NI \\
 & PD1 & $+0.00674$ open & $+0.00572$ open & $+0.00125$ open & $+0.00358$ open & $+0.00605$ open & $+0.00477$ NI & $+0.00357$ NI \\
 & ODBench & $+0.00191$ open & $+0.00261$ open & $+0.00298$ open & $+0.00140$ open & $+0.00049$ NI & $+0.00087$ NI & $+0.00197$ NI \\
\bottomrule
\end{tabular}}
\end{table}
}

\subsection{Findings under the Controlled Surrogates}
\label{sec:controlled-results}

\textbf{F1: EI stops are premature on average at the lower costs.}\label{sec:inmodel-gap}
At a fixed EI stop, increasing the cost cannot increase the optimal continuation premium. We compute the gap to the Bellman optimum exactly on small instances: EI stopping episodes under GP-atJ, frozen at their first EI stop and reduced to three arms with two further increments each ($3{\times}2$ instances) or two arms with three ($2{\times}3$; Appendix~\ref{app:inmodel-oracle}). Dropping arms and increments only removes options, so the instances can show that a stop is premature but not that it is optimal in the full problem.
\InModelOraclePreregFinding\ \InModelOracleCostSplitFinding\ (Figure~\ref{fig:findings}; Table~\ref{tab:inmodel-oracle-prereg}).
On the full pool, lower and upper bounds on the gap at the same 840 states (Table~\ref{tab:full-scale-bounds}) show the EI stop premature on average at every cost up to $0.001$, and at $0.002$ on PD1 and ODBench. At the two highest costs the lower bound is below one increment's cost, but the upper bound is too loose to establish any cell near-optimal.

\textbf{F2: On recorded curves, certified departures gain most at low costs and little at high costs, with losses recorded on LCBench.}
The gate's effect is nonnegative under the surrogate (Corollary~\ref{cor:receding-policy-improvement}); on recorded curves it is measured. \BControllerFindingShort\ The pre-registered test finds EI stopping
noninferior to H2 stopping in 10 of 21 cells, seven of the nine at $\lambda\ge0.002$. \BFreshBandSigns\ This measures the value of the certified correction, not the gap to the optimal policy (Table~\ref{tab:body-effects}; Appendix~\ref{app:primary50}).

\begin{table*}[t]\centering\scriptsize\setlength{\tabcolsep}{3pt}\renewcommand{\arraystretch}{0.8}
\caption{Score--resource tradeoffs for EI stopping (EI-S) on published surrogates: on FT-PFN, CFBO-NT (same weights, native stop, cap $B=300$) and ifBO run to $B$; on DyHPO's deep-kernel GP, DyHPO run to $B$. $\Delta\tilde y$, $\Delta\tau$: score and increment differences, EI-S minus the method; $\bar\tau_{\text{EI-S}}$: mean increments of EI-S; scores lie in $[0,1]$, and utility is in the same units; $^\dagger$/$^\ddagger$: simultaneous 95\% band above/below zero; $^{\mathrm{x}}$: $\alpha_{\rm CFBO}>1$. Bands and break-even costs: Table~\ref{tab:learned-ei}.}
\label{tab:body-f4}
\fitwidth{\begin{tabular}{@{}lr r rrr rrr r rrr@{}}\toprule
 & & \multicolumn{7}{c}{FT-PFN} & \multicolumn{4}{c}{Deep-kernel GP} \\ \cmidrule(lr){3-9}\cmidrule(lr){10-13}
 & & & \multicolumn{3}{c}{CFBO-NT, native stop, $B=300$} & \multicolumn{3}{c}{ifBO, $B=300$} & & \multicolumn{3}{c}{DyHPO, $B=300$} \\ \cmidrule(lr){4-6}\cmidrule(lr){7-9}\cmidrule(lr){11-13}
Bench. & $\lambda$ & $\bar\tau_{\text{EI-S}}$ & $U^{\text{FT-PFN}}_{\text{EI-S}}-U_{\text{CFBO-NT}}$ & $\Delta\tilde y$ & $\Delta\tau$ & $U^{\text{FT-PFN}}_{\text{EI-S}}-U_{\text{ifBO}}$ & $\Delta\tilde y$ & $\Delta\tau$ & $\bar\tau_{\text{EI-S}}$ & $U^{\text{deep-kernel GP}}_{\text{EI-S}}-U_{\text{DyHPO}}$ & $\Delta\tilde y$ & $\Delta\tau$ \\ \midrule
LCBench & 0.0001 & $117$ & $+0.0209^{\dagger}$ & $+0.0244$ & $+35$ & $+0.0103^{\dagger}$ & $-0.0080$ & $-183$ & $231$ & $+0.0036^{\dagger}$ & $-0.0033$ & $-69$ \\
 & 0.0002 & $92$ & $+0.0215^{\dagger}$ & $+0.0276$ & $+31$ & $+0.0312^{\dagger}$ & $-0.0104$ & $-208$ & $226$ & $+0.0112^{\dagger}$ & $-0.0036$ & $-74$ \\
 & 0.0005 & $66$ & $+0.0220^{\dagger}$ & $+0.0327$ & $+21$ & $+0.1030^{\dagger}$ & $-0.0141$ & $-234$ & $217$ & $+0.0375^{\dagger}$ & $-0.0038$ & $-83$ \\
 & 0.001 & $47$ & $+0.0318^{\dagger}$ & $+0.0410$ & $+9$ & $+0.2348^{\dagger}$ & $-0.0181$ & $-253$ & $207$ & $+0.0885^{\dagger}$ & $-0.0043$ & $-93$ \\
 & 0.002 & $29$ & $+0.0582^{\dagger}$ & $+0.0473$ & $-5$ & $+0.5177^{\dagger}$ & $-0.0243$ & $-271$ & $195$ & $+0.2049^{\dagger}$ & $-0.0047$ & $-105$ \\
 & 0.005$^{\mathrm{x}}$ & $13$ & $+0.1172^{\dagger}$ & $+0.0492$ & $-14$ & $+1.3946^{\dagger}$ & $-0.0391$ & $-287$ & $172$ & $+0.6334^{\dagger}$ & $-0.0056$ & $-128$ \\
 & 0.01$^{\mathrm{x}}$ & $8$ & $+0.1627^{\dagger}$ & $+0.0440$ & $-12$ & $+2.8614^{\dagger}$ & $-0.0549$ & $-292$ & $147$ & $+1.5196^{\dagger}$ & $-0.0066$ & $-153$ \\
\midrule
PD1 & 0.0001 & $180$ & $+0.0586^{\dagger}$ & $+0.0631$ & $+45$ & $+0.0076^{\dagger}$ & $-0.0044$ & $-120$ & $261$ & $+0.0008$ & $-0.0031$ & $-39$ \\
 & 0.0002 & $151$ & $+0.0581^{\dagger}$ & $+0.0671$ & $+45$ & $+0.0231^{\dagger}$ & $-0.0067$ & $-149$ & $259$ & $+0.0051^{\dagger}$ & $-0.0032$ & $-41$ \\
 & 0.0005 & $105$ & $+0.0570^{\dagger}$ & $+0.0681$ & $+22$ & $+0.0837^{\dagger}$ & $-0.0138$ & $-195$ & $254$ & $+0.0193^{\dagger}$ & $-0.0035$ & $-46$ \\
 & 0.001 & $66$ & $+0.0717^{\dagger}$ & $+0.0630$ & $-9$ & $+0.2076^{\dagger}$ & $-0.0260$ & $-234$ & $250$ & $+0.0460^{\dagger}$ & $-0.0038$ & $-50$ \\
 & 0.002 & $32$ & $+0.1226^{\dagger}$ & $+0.0597$ & $-31$ & $+0.4958^{\dagger}$ & $-0.0401$ & $-268$ & $242$ & $+0.1117^{\dagger}$ & $-0.0046$ & $-58$ \\
 & 0.005$^{\mathrm{x}}$ & $13$ & $+0.2175^{\dagger}$ & $+0.0517$ & $-33$ & $+1.3722^{\dagger}$ & $-0.0643$ & $-287$ & $207$ & $+0.4547^{\dagger}$ & $-0.0097$ & $-93$ \\
 & 0.01$^{\mathrm{x}}$ & $9$ & $+0.2528^{\dagger}$ & $+0.0415$ & $-21$ & $+2.8258^{\dagger}$ & $-0.0888$ & $-291$ & $175$ & $+1.2367^{\dagger}$ & $-0.0142$ & $-125$ \\
\midrule
ODBench & 0.0001 & $194$ & $+0.0437^{\dagger}$ & $+0.0477$ & $+40$ & $-0.0025$ & $-0.0131$ & $-106$ & $168$ & $+0.0109^{\dagger}$ & $-0.0023$ & $-132$ \\
 & 0.0002 & $127$ & $+0.0525^{\dagger}$ & $+0.0548$ & $+11$ & $+0.0163^{\dagger}$ & $-0.0183$ & $-173$ & $162$ & $+0.0250^{\dagger}$ & $-0.0025$ & $-138$ \\
 & 0.0005 & $51$ & $+0.0732^{\dagger}$ & $+0.0595$ & $-27$ & $+0.0956^{\dagger}$ & $-0.0289$ & $-249$ & $154$ & $+0.0704^{\dagger}$ & $-0.0027$ & $-146$ \\
 & 0.001 & $36$ & $+0.0711^{\dagger}$ & $+0.0536$ & $-18$ & $+0.2241^{\dagger}$ & $-0.0403$ & $-264$ & $148$ & $+0.1491^{\dagger}$ & $-0.0030$ & $-152$ \\
 & 0.002 & $27$ & $+0.0619^{\dagger}$ & $+0.0362$ & $-13$ & $+0.4910^{\dagger}$ & $-0.0544$ & $-273$ & $139$ & $+0.3187^{\dagger}$ & $-0.0037$ & $-161$ \\
 & 0.005$^{\mathrm{x}}$ & $19$ & $+0.0816^{\dagger}$ & $+0.0343$ & $-9$ & $+1.3271^{\dagger}$ & $-0.0788$ & $-281$ & $124$ & $+0.8757^{\dagger}$ & $-0.0057$ & $-176$ \\
 & 0.01$^{\mathrm{x}}$ & $15$ & $+0.1119^{\dagger}$ & $+0.0552$ & $-6$ & $+2.7456^{\dagger}$ & $-0.1093$ & $-285$ & $112$ & $+1.8701^{\dagger}$ & $-0.0081$ & $-188$ \\
\bottomrule\end{tabular}}\end{table*}

\textbf{F3: Across the three surrogates the transfer pattern holds, and more information sharing raises EI stopping's utility.}
The three surrogates differ only in how they share transients (Section~\ref{sec:shared-models}). \BCrossLawFreshFinding\ None of the three shows noninferiority at $\lambda=0.0001$, and under all three certified departures lose significantly on LCBench (Figure~\ref{fig:findings}). \EISharingFindingFresh\ This is why GP-atJ is the primary surrogate (Appendix~\ref{app:ei-surrogate-three-way}). Surrogates outside this Gaussian family are not run with H2 stopping, so the transfer may still depend on the surrogate.

\subsection{Score--Resource Tradeoffs on Published Surrogates}
\label{sec:external-comparators}

\textbf{F4: EI stopping trades score for fewer training increments, with the exchange valued by utility.} We run EI stopping on the
surrogates of ifBO and DyHPO, which are FT-PFN and a deep-kernel GP~\citep{wilson2016deep} (Table~\ref{tab:body-f4}; Appendix~\ref{app:published}). Run to a budget of 300 increments, ifBO and DyHPO return higher mean scores in every cell, using $39$ to $292$ more increments on average. EI stopping attains the higher utility when the priced increments it saves exceed the score it gives up: the break-even cost $\lambda^\star$, mean score loss over mean increments saved, ranges from $0.00002$ to $0.00038$ (Table~\ref{tab:learned-ei}), and the utility-difference band lies above zero in 20 of 21 cells against each budgeted method, the remaining cells at $\lambda=0.0001$. Against CFBO-NT (CFBO without its transfer-trained extrapolator), EI stopping has higher scores and utility-difference bands above zero in all 21 cells.

\section{Conclusion}
\label{sec:conclusion}

The two-step gate certifies EI stops made premature by delayed fidelity gains or by observation-dependent switching.
Under the surrogate, EI stops lose value at low costs and little at high costs, while full-scale near-optimality remains
open. On recorded curves, EI stopping is noninferior to its certified correction in most high-cost settings; at lower
costs the correction gains on PD1 and loses on LCBench under all three surrogates, and more information sharing raises
EI stopping's utility. On published surrogates, EI stopping trades lower scores for fewer increments, which utility prices
at each cost. The certificate finds value that one-step stopping misses; experiments on recorded curves test whether it is
realized.

\paragraph{Limitations.}
Our results concern one utility: linear in a cost per increment, risk-neutral, and defined on the best observed score.
An increment is a benchmark-defined checkpoint that also fixes the lookahead depth and the surrogate's timescales, so
findings by cost are relative to each benchmark's unit, and differences between benchmarks are confounded with that unit
(Appendix~\ref{app:formulation-omissions}). The certificate's computational cost is not charged in the objective. The
experiments replay fixed pools on three HPO benchmarks; at the lowest costs, some configuration's recorded curve has
already run out at up to nine in ten EI stops (Appendix~\ref{app:horizon-split}). The surrogates are fixed in advance and
fitted to no curves; we test neither a fitted planning law nor cost-aware regret-based stopping.

\subsection*{AI Use Statement}
In this work, we used generative AI tools to implement the methods and the experiment code, to clean and reformat
the benchmark datasets, to interpret results, and to audit the scope and wording of theoretical and empirical claims.
We have not used generative AI tools to generate synthetic data sets, to help develop theoretical models or conceptual
frameworks, to provide critical ingredients for proofs, to assist in the
writing of proofs, to propose or refine hypotheses, to design the research methodology or
experiments, or to assist with translation, and qualitative and thematic data analysis is not applicable to this work.
Additionally, we used generative AI tools to create and edit software code, to orchestrate experiment runs, to create
tables and figures, to draft and edit expository parts of the paper for readability, to check wording and format
references, and to search for information.
We have reviewed all AI-assisted work.
AI-generated code was verified by the authors and is covered by the automated test suite; every reported number, table
and figure is regenerated from stored artifacts and was checked against them; and every reference was checked against
its original source.
We take responsibility for the final content of this work, including text, claims, or artifacts produced with the aid
of generative AI.

\bibliographystyle{plainnat}
\bibliography{p1}

\clearpage
\appendix
\renewcommand{\topfraction}{0.95}\renewcommand{\bottomfraction}{0.95}
\renewcommand{\textfraction}{0.03}\renewcommand{\floatpagefraction}{0.85}
\setcounter{topnumber}{4}\setcounter{bottomnumber}{3}\setcounter{totalnumber}{6}
\section*{Appendix}

The appendix follows the order of the paper. Appendix~\ref{app:supp-theory}, after the notation of Table~\ref{tab:notation}, completes the theory of
Sections~\ref{sec:formulation} and~\ref{sec:theory} and Appendix~\ref{app:proofs} proves it;
Appendix~\ref{app:impl} specifies the surrogates and the certificate; Appendix~\ref{app:provenance} gives the
experimental protocol and the statistics; Appendices~\ref{app:controlled-results} and~\ref{app:published} report
F1--F3 and F4 in full; Appendix~\ref{app:scope-future} states what the formulation leaves out. Every number quoted
in Appendices~\ref{app:controlled-results} and~\ref{app:published} is read from a table.

\section{Notation and Additional Theory}
\label{app:supp-theory}

\begin{table}[ht]
\centering\footnotesize
\setlength{\tabcolsep}{4pt}
\renewcommand{\arraystretch}{1.05}
\caption{Notation. Section~\ref{sec:formulation} defines the model and Bellman
quantities, Section~\ref{sec:theory} the headroom $d(S)$ and the two-step quantities and policies, and
Appendix~\ref{app:inmodel-oracle} ideal H2 stopping $\pi_2^{\rm ideal}$.}
\label{tab:notation}
\fitwidth{%
\begin{tabular}{@{}ll@{\hspace{12pt}}ll@{\hspace{12pt}}ll@{}}
\toprule
\multicolumn{2}{@{}l}{\textit{Model}} & \multicolumn{2}{l}{\textit{Bellman}} & \multicolumn{2}{l}{\textit{Two-step certificate and policies}} \\
\midrule
$S_b$ & history state after $b$ advances & $V_\lambda^\star$, $Q_\lambda^\star$ & optimal value and action value & $g(S)$ & largest available EI, augmented by $0$ \\
$\tilde y(S)$ & incumbent score, $\tilde y_b=\tilde y(S_b)$ & $\Delta_\lambda^\star(S,i)$ & centered advantage of arm $i$ & $A^{\rm pool}_{2,i}(S)$ & pool-aware two-step advantage \\
$h_\lambda(S)$ & stop-now payoff $\tilde y(S)-\lambda b$ & $O_i(S)$ & downstream option value & $A_{2,i}(S)$ & same-arm two-step advantage \\
$\cI(S)$ & available arms & $\mathcal S_\lambda$, $\mathcal S^{\mathrm{myo}}_\lambda$ & Bellman and one-step stop sets & $\Gamma_i(S)$ & switching premium $A^{\rm pool}_{2,i}-A_{2,i}$ \\
$d(S)$ & headroom $y_{\max}-\tilde y(S)$ & $\pi_{\rm EI}$ & EI stopping (EI-S) & $\underline A^{\rm pool}_{2,i}(S)$ & certified lower enclosure \\
$\EI_i(S)$ & one-step expected improvement & $\pi_{\rm H2}$ & H2 stopping (H2-S), the certified policy & $U$ & episode utility $\tilde y_\tau-\lambda\tau$ \\
$\tau_\pi$ & first stop of policy $\pi$; $\tau_{\rm EI}$, $\tau_{\rm H2}$ for $\pi_{\rm EI}$, $\pi_{\rm H2}$ & $V^{(L)}$ & value with at most $L$ further increments & $\pi_2^{\rm ideal}$ & ideal H2 stopping (appendix) \\
$\E_{\theta,\pi}[\,\cdot\mid S]$ & expectation under the planning law and policy $\pi$ from $S$ & $\E_\theta[\,\cdot\mid S,i]$ & one-step expectation over $Y_{b+1}$ & $\kappa(S)$ & candidate arm at an EI stop \\
\midrule
\multicolumn{2}{@{}l}{\textit{Setup}} & \multicolumn{2}{l}{\textit{Surrogate family (Section~\ref{sec:shared-models})}} & \multicolumn{2}{l}{\textit{Estimands (Section~\ref{sec:experiments})}} \\
\midrule
$\lambda$ & cost per increment & $a_i$ & asymptote of arm $i$ & $U_A-U_B$, $\Delta\tilde y$, $\Delta\tau$ & paired utility, score and increment differences, method $A$ minus $B$ \\
$\theta$, $p_\theta$ & surrogate parameters; planning law & $b_{ir}$, $\phi_r$, $s_r$ & transient coefficient, basis and timescale, $r\le K$ & $\Delta_{\text{H2-S}}$, UCL & expected gain of H2-S over EI-S; its one-sided simultaneous 95\% upper confidence limit \\
$N$, $t_{i,b}$ & pool size; local fidelity of arm $i$ & $v_a$, $v_t$ & level and transient prior variances & $\delta$ & noninferiority tolerance, $\delta=\lambda$ \\
$\cC_b$, $F$ & common context; successor map & $C_a$, $C_t$ & cross-configuration correlations & $\lambda^\star$ & break-even cost per increment \\
$H(S)$ & finite cap on further increments & $R$, $J$ & Mat\'ern HP kernel; all-ones matrix & & \\
\bottomrule
\end{tabular}}
\end{table}

\subsection{Standing Assumptions}
\label{app:assumptions}

\begin{assumption}[Fixed planning law]\label{assm:fixed-sim}
The history state space is standard Borel; $F,h_\lambda$ and the fixed transition
kernel $p_\theta$ are measurable; the finite-valued availability correspondence
has measurable graph. Transitions increment $b$ and update the incumbent as
defined above, including all predetermined predictive updates.
\end{assumption}
\begin{assumption}[Bounded scores]\label{assm:bounded-perf}
All scores and incumbents remain in fixed $[y_{\min},y_{\max}]$ bounds chosen
before deployment. LCBench validation accuracy and PD1's $1-\texttt{valid/error\_rate}$
use $[0,1]$, without per-task rescaling.
\end{assumption}
\begin{assumption}[Coercive cost]\label{assm:coercive}
The cost of $b$ increments is $\lambda b$ with $\lambda>0$, where $b$ is the global count of increments.
\end{assumption}
Randomized policies are represented by an exogenous seed that, conditional on the current state, is
independent of all future potential observation streams.
Assumption~\ref{assm:fixed-sim} fixes the rule mapping history to a predictive distribution, including all predetermined updates. A prescribed retraining procedure can define such a kernel; a rollout must then include that procedure at every fantasized history. Our Gaussian implementation fixes hyperparameters so that those updates admit the enclosure of Section~\ref{sec:analytical-enclosure} (Appendix~\ref{app:no-certificate-learned}).
Assumption~\ref{assm:bounded-perf} rules out unbounded scores and rescaling by observed outcomes; the incumbent's gain is therefore bounded. Assumption~\ref{assm:coercive} rules out free increments; every increment thus has a positive cost, and EI stopping trains finitely often in expectation (Lemma~\ref{lem:ei-beats-stop}).

\subsection{Well-Posedness and Forward Closure}
\label{app:forward-closure}

\begin{proposition}[Finiteness and value-iteration convergence]\label{prop:existence}
Under Assumptions~\ref{assm:fixed-sim}--\ref{assm:coercive}, the value in~\eqref{eq:opt-problem} is finite and is the monotone limit of the finite-depth values $V^{(0)}=h_\lambda$ and
$V^{(L+1)}(S_b)=\max\{h_\lambda(S_b),\max_{i\in\cI(S_b)}\E_\theta[V^{(L)}(F(S_b,i,Y_{b+1}))\mid S_b,i]\}$.
Every policy whose expected utility is at least that of stopping immediately satisfies
\[
\E_{\theta,\pi}[\tau_\pi-b\mid S_b]\le \frac{d(S_b)}{\lambda}
\le\frac{y_{\max}-y_{\min}}{\lambda}.
\]
For normalized scores the final bound is $1/\lambda$. It controls the mean of the remaining duration, whose support may be unbounded.
\end{proposition}

Formally, EI stopping (Section~\ref{sec:theory}) is
\[
\pi_{\rm EI}(S_b)=
\begin{cases}
\mathrm{stop},&\cI(S_b)=\varnothing\ \text{or}\ \max_{i\in\cI(S_b)}\EI_i(S_b)\le\lambda,\\
\min\argmax_{i\in\cI(S_b)}\EI_i(S_b),&\text{otherwise.}
\end{cases}
\]
Proposition~\ref{prop:existence} applies to $\pi_{\rm EI}$ with no further hypothesis, and
Lemma~\ref{lem:ei-beats-stop} is the freeze--thaw BO counterpart of the complete-evaluation guarantee
of \citet{xie2025coststop}.

\begin{lemma}[Where EI stopping continues, so does the Bellman rule]\label{lem:free-direction}
Under Assumptions~\ref{assm:fixed-sim}--\ref{assm:coercive}, for every state, $\max_i\Delta_\lambda^\star(S,i)\ge\max_i(\EI_i(S)-\lambda)$. Whenever EI stopping continues, immediate stopping is strictly suboptimal, although the arm it advances need not be Bellman-maximizing.
\end{lemma}
\begin{proof}
Equation~\eqref{eq:delta-recursion} has $O_i(S)\ge0$ for every available action.
\end{proof}

\begin{assumption}[Joint forward-closed stop set]\label{assm:monotone}
For every $S_b\in\mathcal S^{\mathrm{myo}}_\lambda$ and every $i\in\cI(S_b)$,
$F(S_b,i,Y_{b+1})\in\mathcal S^{\mathrm{myo}}_\lambda$ almost surely.
\end{assumption}
The quantifier over every action is needed: closure under a single selected action would leave other profitable
escapes open. Under Assumption~\ref{assm:monotone}, Theorem~\ref{thm:myopic-optimal} is the classical monotone
one-stage-look-ahead argument specialized to the EI stopping action set; it classifies stop and continue but not
the Bellman-optimal continuation arm.

\subsection{Same-Arm Reduction and Switching Premium}
\label{app:same-arm}

At an EI stop $S_b\in\mathcal S^{\mathrm{myo}}_\lambda$ with root action $i$ and
$S_{b+1}=F(S_b,i,Y_{b+1})$, the same-arm value allows only a second advance of the root arm:
\begin{equation}
A_{2,i}(S_b)=(\EI_i(S_b)-\lambda)
+\E_\theta\!\left[(\EI^{\rm cap}_i(S_{b+1})-\lambda)^+\mid S_b,i\right]
\le A^{\rm pool}_{2,i}(S_b).
\label{eq:same-arm-depth-two-advantage}
\end{equation}

\paragraph{Switching Premium.}
Let $M_{-i}(S_{b+1})$ be the zero-augmented maximum EI of feasible children other
than $i$, and define $\Gamma_i(S_b)=A^{\rm pool}_{2,i}(S_b)-A_{2,i}(S_b)\ge0$ under the same planning law and root action.
The difference of the two positive-part child values is positive exactly when
$M_{-i}(S_{b+1})>\max\{\lambda,\EI^{\rm cap}_i(S_{b+1})\}$. Since it is bounded and
nonnegative, $\Gamma_i(S_b)>0$ iff this event has positive root-conditional
probability. Switching is \emph{essential to a profitable H2 departure} when
$A_{2,i}\le0<A^{\rm pool}_{2,i}$; positive $\Gamma_i$ alone need not offset
the root deficit. The shared-context example of Appendix~\ref{app:counterexample} has
$A_{2,D}=-1/5$ and $A^{\rm pool}_{2,D}=1/10$.

\begin{assumption}[No cross-arm resurrection]\label{assm:no-resurrection}
For every $j\ne i$,
\[
\Prob_\theta\!\left(j\notin\cI(S_{b+1})\ \lor\ \EI_j(S_{b+1})\le\lambda\mid S_b,i\right)=1.
\]
\end{assumption}

\begin{corollary}[Same-arm reduction]\label{cor:same-arm-reduction}
Under Assumptions~\ref{assm:fixed-sim}--\ref{assm:coercive}, at an EI stop, Assumption~\ref{assm:no-resurrection} implies
$A^{\rm pool}_{2,i}(S)=A_{2,i}(S)$
for every available root arm $i$. Without that condition,
$A_{2,i}(S)\le A^{\rm pool}_{2,i}(S)$ remains valid.
\end{corollary}

\begin{proposition}[Per-arm conditioning removes cross-arm resurrection]\label{prop:gp-pool}
Under arm-local availability, at a state in $\mathcal S^{\mathrm{myo}}_\lambda$, advancing arm $i$ under a frozen per-arm planning kernel leaves the complete predictive law of every $j\ne i$ unchanged, including any deterministic per-arm state. Hence $\EI_j(S_{b+1})\le\EI_j(S_b)\le\lambda$ for every $j\ne i$ available at the successor, and Assumption~\ref{assm:no-resurrection} holds.
\end{proposition}

\subsection{Delayed Improvement and One-Step Error}
\label{app:counterexample}
Two cases show how an EI stop can be premature. The first has one arm, incumbent $q=0$ and cost
$\lambda=1/8$; the arm's curve is $(0,0)$ or $(0,1)$ with equal probability. Since the first-step EI is zero, EI stopping ends the search. Observation $0$ at
the first increment leaves both curves possible, the next fidelity then has EI $1/2$, which gives
\[
O_1(S_0)=3/8,
\qquad
\Delta_\lambda^\star(S_0,1)=-1/8+3/8=1/4.
\]
Thus nonnegative option value can strictly reverse the stopping decision without re-ranking or switching. Forward closure fails because $S_0\in\mathcal S^{\mathrm{myo}}_\lambda$ but $S_1\notin\mathcal S^{\mathrm{myo}}_\lambda$. Forward closure is sufficient but not necessary: reducing the expected second observation to any $\mu_2$ with $\lambda<\mu_2\le2\lambda$ still violates closure but gives $\mu_2-2\lambda\le0$, making initial stopping Bellman-optimal.

The second case has incumbent $q=1/2$, cost $\lambda=1/5$, a diagnostic arm $D$ and three target arms
$T_1,T_2,T_3$; the shared remaining cap is two and a latent state $\ell$ is uniform on $\{1,2,3\}$.
The one-step diagnostic returns one of the distinct values $0,1/8,1/4$, all below the incumbent, and thereby identifies $\ell$; target $T_j$ returns one when $j=\ell$ and zero otherwise; hence $\EI_D=0$ and $\EI_{T_j}=1/6<\lambda$, and EI stopping ends the search. Every standalone
one-arm value is zero, yet observing the diagnostic and then the identified target has centered value
$1/10$. Equation~\eqref{eq:depth-two-advantage} detects this exactly:
$A^{\rm pool}_{2,D}=-1/5+(1/2-1/5)=1/10$, whereas the exhausted same-arm rollout
does not. Under shared-context dependence, profitable continuation can therefore coexist with zero standalone values, which justify neither stopping nor zero-value action pruning.

\subsection{Asymptote Sharing at the Next Increment}
\label{app:asymptote-attenuation}
We prove the claim of Section~\ref{sec:shared-models} that asymptote-only sharing reaches an arm's next increment only weakly: an asymptote does not enter the next increment; other arms therefore reach it through one scalar loading.

\paragraph{Level--Transient Surrogates.}
Write $Y_i(t)=m+a_i+u_i(t)+\epsilon_i(t)$ with $a\sim\mathcal N(0,\Sigma^a_0)$, a centered
Gaussian transient process $u=(u_1,\ldots,u_N)$ independent of $a$, and independent fresh noise
$\epsilon_i(t)\sim\mathcal N(0,\nu_i(t))$, $\nu_i(t)>0$, independent of $(a,u)$. This surrogate has
\emph{asymptote-only sharing} if $u_1,\ldots,u_N$ are mutually independent. GP-a has
it; GP-at and GP-atJ do not. Let arm $i$ have a finite set of observed times $\mathcal T_i$ with latest time $n\ge1$,
observations $\cC^{(i)}$, and let $\cC^{(-i)}$ be the observations of the other arms. With
$k_u(t,\mathcal T_i)=\operatorname{Cov}(u_i(t),u_i(\mathcal T_i))$, $\mathbf K_{u,i}=\operatorname{Cov}(u_i(\mathcal T_i))$ and
$D_i=\operatorname{diag}\nu_i(\mathcal T_i)$, the loading is $w_i(t)=1-k_u(t,\mathcal T_i)[\mathbf K_{u,i}+D_i]^{-1}\mathbf1$,
so that $\operatorname{Cov}(Y_i(t_*),Y_j(t'_*)\mid\cC)=w_i(t_*)\Sigma^a_{ij}w_j(t'_*)$ for $i\ne j$ under
asymptote-only sharing, with $\Sigma^a$ the asymptote posterior covariance, and we call
$\xi_i=u_i(n+1)-u_i(n)$ arm $i$'s next increment.

\begin{proposition}[Increment decomposition]
\label{prop:increment-decomposition}
For every level--transient surrogate:
\begin{enumerate}
\item[(i)] $Y_i(n+1)=m+a_i+u_i(n)+\xi_i+\epsilon_i(n+1)$, and $\xi_i$ is independent of the
asymptotes $a$.
\item[(ii)] Under asymptote-only sharing, $\xi_i$ is conditionally independent of
$\cC^{(-i)}$ given $(a_i,\cC^{(i)})$. With constants $\mu_i$ and $\omega_i^2$ that depend on $\cC^{(i)}$ alone,
\[
\begin{aligned}
\E[\xi_i\mid\cC]&=\mu_i+\bigl(w_i(n+1)-w_i(n)\bigr)\,\E[a_i\mid\cC],\\
\operatorname{Var}(\xi_i\mid\cC)&=\omega_i^2+\bigl(w_i(n+1)-w_i(n)\bigr)^2\operatorname{Var}(a_i\mid\cC).
\end{aligned}
\]
The same holds for $Y_i(n+1)$ with the loading $w_i(n+1)$ in place of $w_i(n+1)-w_i(n)$.
Hence other arms' observations reach arm $i$'s one-step predictive law only through the two
posterior moments of $a_i$, scaled by $w_i(n+1)$ and $w_i(n+1)^2$.
\item[(iii)] Under asymptote-only sharing, $\operatorname{Cov}(\xi_i,Y_j(t'))=0$ for every
$j\ne i$ and $t'\ge1$. Under the finite-basis surrogates of Section~\ref{sec:shared-models},
\[
\operatorname{Cov}(\xi_i,Y_j(t'))=\frac{v_t}{K}\,C_{t,ij}\sum_{r=1}^K
\bigl(\phi_r(n+1)-\phi_r(n)\bigr)\phi_r(t'),
\]
which is nonzero if and only if $C_{t,ij}\ne0$. Under GP-atJ,
$C_{t,ij}\ge1/4$ for every pair.
\end{enumerate}
\end{proposition}

\begin{proof}
(i) Substitute $u_i(n+1)=u_i(n)+\xi_i$. $\xi_i$ is a function of $u$, which is independent
of $a$.

(ii) Under asymptote-only sharing, $(u_i,\epsilon_i)$ is independent of
$(a,u_{-i},\epsilon_{-i})$; hence conditioning on $a_i$ leaves $(u_i,\epsilon_i)$ independent of
$(a_{-i},u_{-i},\epsilon_{-i})$. Given $a_i$, $\cC^{(i)}$ is a function of $(u_i,\epsilon_i)$ and
$\cC^{(-i)}$ a function of $(a_{-i},u_{-i},\epsilon_{-i})$; therefore $\xi_i$ is independent of
$\cC^{(-i)}$ given $(a_i,\cC^{(i)})$. Given $a_i$, the residual
$y_i(\mathcal T_i)-(m+a_i)\mathbf1=u_i(\mathcal T_i)+\epsilon_i(\mathcal T_i)$ is Gaussian with covariance
$\mathbf K_{u,i}+D_i$, which gives
$\E[u_i(t)\mid a_i,\cC^{(i)}]=k_u(t,\mathcal T_i)[\mathbf K_{u,i}+D_i]^{-1}(y_i(\mathcal T_i)-m\mathbf1)-
k_u(t,\mathcal T_i)[\mathbf K_{u,i}+D_i]^{-1}\mathbf1\,a_i$, and the conditional variance does not
depend on $a_i$. Differencing at $t=n+1$ and $t=n$ gives
$\E[\xi_i\mid a_i,\cC^{(i)}]=\mu_i+(w_i(n+1)-w_i(n))a_i$ with constant variance $\omega_i^2$. By
conditional independence, $\E[\xi_i\mid a_i,\cC]=\E[\xi_i\mid a_i,\cC^{(i)}]$. The towers of
expectation and variance then give the display. For $Y_i(n+1)$, the coefficient of $a_i$ in
$\E[m+a_i+u_i(n+1)\mid a_i,\cC^{(i)}]$ is $w_i(n+1)$, and $\epsilon_i(n+1)$ adds a constant variance.

(iii) Under asymptote-only sharing, $\xi_i$ is a function of $u_i$, and
$Y_j(t')$ is a function of $(a_j,u_j,\epsilon_j(t'))$, which is independent of $u_i$.
Under a finite-basis surrogate, $\xi_i=\sum_r b_{ir}(\phi_r(n+1)-\phi_r(n))$, and
$\operatorname{Cov}(b_{ir},b_{jr'})=\mathbb 1_{r=r'}(v_t/K)C_{t,ij}$ with $(b_{ir})\perp(a,\epsilon)$.
Each $\phi_r$ is positive and strictly decreasing; every summand is therefore negative and the sum is
nonzero. The unit-diagonal Mat\'ern matrix $R$ has nonnegative entries, which gives $C_{t,ij}=R_{ij}/4+1/4\ge1/4$ under GP-atJ.
\end{proof}

\begin{corollary}[GP-a, one observation]
\label{cor:gpa-increment-loading}
Under GP-a, $u_i(t)=\sum_{r=1}^K b_{ir}\phi_r(t)$ with independent $b_{ir}\sim\mathcal N(0,v_t/K)$, so that
$k_u(t,t')=(v_t/K)\sum_r\phi_r(t)\phi_r(t')$. With a single observation at time $n$ ($\mathcal T_i=\{n\}$) and
noise variance $\nu=\nu_i(n)$,
\[
w_i(n)=\frac{\nu}{k_u(n,n)+\nu},\qquad
0\le w_i(n+1)-w_i(n)=\frac{k_u(n,n)-k_u(n+1,n)}{k_u(n,n)+\nu}\le\frac{1}{s_1+n+1}.
\]
\end{corollary}

\begin{proof}
With one observation, $w_i(t)=1-k_u(t,n)/(k_u(n,n)+\nu)$, which gives both identities. Each $\phi_r$ is positive
and decreasing, so $k_u(n+1,n)\le k_u(n,n)$ and the difference is nonnegative. Dropping $\nu$ from the denominator,
it is at most $\sum_r\phi_r(n)^2\rho_r/\sum_r\phi_r(n)^2$ with $\rho_r=1-\phi_r(n+1)/\phi_r(n)=1/(s_r+n+1)$, a
weighted mean of the $\rho_r$, each at most $1/(s_1+n+1)$.
\end{proof}

At the parameters of Section~\ref{sec:experiments}, $v_t=1/32$, $(s_1,\ldots,s_4)=(1,8,64,512)$ and the noise of
Appendix~\ref{app:shared-temporal-b}, and $n=1$, the corollary gives $w_i(1)=0.008$, $w_i(2)-w_i(1)=0.112\le1/3$ and
$w_i(2)=0.120$; at $n=10$ the increment loading is $0.014$. The arm's own observation absorbs the asymptote in the
level, and other arms bear on its next increment only through a loading that shrinks as the arm is observed.
Part~(iii) is the design reason for GP-at and GP-atJ: under them another arm's scores
bear on arm $i$'s next increment directly through $C_t$.
The statements concern the Gaussian law before the one-step truncation to $[0,1]$; the truncated
predictive is a function of the Gaussian mean and variance, and part~(ii) therefore carries over to the
planning law.

\subsection{Search and Switching}
\label{app:related-switching}
A freeze--thaw BO policy chooses at each step which arm to advance or whether to stop. Unlike optimal switching
between regimes~\citep{carmonaludkovski2008}, an arm that is not advanced does not evolve, which is the structure of
Pandora-type search. The index of \citet{weitzman1979} assumes independent boxes, each opened in a single stage; a
freeze--thaw arm is inspected over many stages and, under a shared surrogate, is correlated with the others.
Independent and correlated Pandora models clarify how inspection can direct
later search~\citep{weitzman1979,chawla2023correlated}; posterior BO indices have
different updating semantics~\citep{xie2024pbgi}. Multistage and nested Pandora models extend Weitzman's index to boxes inspected in several stages whose value is
claimed only after the last~\citep{kleinberg2016descending,gupta2019markovian,bowers2024nested}, and Markov search
processes with costly choices generalize them further~\citep{bowers2026cms}. A freeze--thaw arm differs in that every
reading counts toward the incumbent, so the search may leave an arm at any stage and keep its best reading, as on
the lines of Pandora's box with order constraints, whose readings are independent~\citep{boodaghians2023order}. Knowledge-gradient
policies value the same option term through the expected gain of one more measurement~\citep{frazier2008kg},
and their trace-aware multi-fidelity form values continuing a training curve at its
cost~\citep{wu2019takg}; both are acquisition rules without a certified stop.

\section{Proofs}
\label{app:proofs}

\subsection{Centered Continuation Identity}
\label{app:delta-recursion}

\begin{lemma}[Recursion~\eqref{eq:delta-recursion}]\label{lem:delta-recursion}
Under Assumptions~\ref{assm:fixed-sim}--\ref{assm:coercive}, Equation~\eqref{eq:delta-recursion} holds with $O_i(S_b)\ge0$, and $O_i(S_b)=0$ iff $\max_{j\in\cI(S_{b+1})}\Delta_\lambda^\star(S_{b+1},j)\le0$ almost surely.
\end{lemma}
\begin{proof}
The successor payoff satisfies
\[
 h_\lambda(S_{b+1})-h_\lambda(S_b)=(Y_{b+1}-\tilde y_b)^+-\lambda,
\]
whose conditional expectation is $\EI_i(S_b)-\lambda$. Substituting
\[
 V_\lambda^\star(S_{b+1})=h_\lambda(S_{b+1})
  +\max\{0,\max_{j\in\cI(S_{b+1})}\Delta_\lambda^\star(S_{b+1},j)\}
\]
into the definition of $\Delta_\lambda^\star(S_b,i)$ gives~\eqref{eq:delta-recursion}. The second summand is the expectation of a nonnegative random variable and therefore vanishes exactly when that variable is zero almost surely.
\end{proof}

\subsection{Value Iteration and EI Stopping}

\paragraph{Proof of Proposition~\ref{prop:existence}.}
Fix a state $S_b$, and let $V^{(L)}(S_b)$ be the value when at most $L$ further increments are allowed. Allowing one more increment cannot lower value, hence $V^{(L)}\le V^{(L+1)}$. Every terminal payoff reachable from $S_b$ is at most $y_{\max}-\lambda b$, hence the sequence has a finite pointwise limit $V^{(\infty)}\le y_{\max}-\lambda b$. Every depth-$L$ policy is admissible, which gives $V^{(L)}\le V_\lambda^\star$ and $V^{(\infty)}\le V_\lambda^\star$.

Conversely, take any admissible policy $\pi$, write $D=\tau_\pi-b$, and replace its stopping time by $\tau_\pi^{(L)}=\tau_\pi\wedge(b+L)$. This is feasible for the depth-$L$ problem, and
\[
 |h_\lambda(S_{\tau_\pi^{(L)}})|
 \le \max\{|y_{\min}|,|y_{\max}|\}+\lambda(b+D),
\]
whose right-hand side is integrable. Dominated convergence therefore gives
\[
 \E_{\theta,\pi} h_\lambda(S_{\tau_\pi})
 =\lim_{L\to\infty}\E_{\theta,\pi} h_\lambda(S_{\tau_\pi^{(L)}})
 \le V^{(\infty)}(S_b).
\]
Taking the supremum over admissible $\pi$ yields $V_\lambda^\star\le V^{(\infty)}$, and thus $V^{(L)}\uparrow V_\lambda^\star$. At a fixed successor state $S_{b+1}$, all iterates lie between $y_{\min}-\lambda(b+1)$ and $y_{\max}-\lambda(b+1)$. Bounded convergence, followed by the maximum over the finite arm set, permits passage to the limit in the finite-depth recursion and gives~\eqref{eq:bellman-linear}.

If a policy is weakly better than stopping immediately, then
\[
 0\le \E_{\theta,\pi}[\tilde y_{\tau_\pi}-\tilde y_b-\lambda D]
 \le d(S_b)-\lambda\E_{\theta,\pi} D,
\]
which proves $\E_{\theta,\pi} D\le d(S_b)/\lambda$. This controls the mean only; with bounded scores, the stopping time can still have unbounded support.

\begin{proof}[Proof of Lemma~\ref{lem:ei-beats-stop}]
For every state and available action, $\max\{\tilde y_k,Y_{k+1}\}-\tilde y_k=(Y_{k+1}-\tilde y_k)^+$, hence $\E_\theta[h_\lambda(S_{k+1})-h_\lambda(S_k)\mid S_k,i]=\EI_i(S_k)-\lambda$. Under
$\pi_{\rm EI}$ the advanced arm attains $g(S_k)>\lambda$ on $\{\tau_{\rm EI}>k\}$; thus
$h_\lambda(S_{k\wedge\tau_{\rm EI}})$ is a submartingale; all expectations below are $\E_{\theta,\pi_{\rm EI}}[\,\cdot\mid S_b]$. Optional stopping at $\tau_n=\tau_{\rm EI}\wedge(b+n)$
gives $\E[h_\lambda(S_{\tau_n})]\ge h_\lambda(S_b)$, hence
$\lambda\,\E[\tau_n-b]\le\E[\tilde y_{\tau_n}-\tilde y_b]\le d(S_b)$ for every $n$, and monotone
convergence yields $\E[\tau_{\rm EI}-b]\le d(S_b)/\lambda<\infty$. Increments are bounded by $(y_{\max}-y_{\min})+\lambda$, hence
$|h_\lambda(S_{\tau_n})|\le|h_\lambda(S_b)|+[(y_{\max}-y_{\min})+\lambda](\tau_{\rm EI}-b)$ is an
integrable dominating variable, and dominated convergence transfers the inequality to $\tau_{\rm EI}$.
The displayed identity is the same telescoping sum without the bounding step, and its first
term is $g(S_b)-\lambda$; the inequality is strict when $g(S_b)>\lambda$.
\end{proof}

\subsection{Exact Stopping and the Depth-Two Rollout Bound}

\paragraph{Proof of Theorem~\ref{thm:myopic-optimal}.}
Lemma~\ref{lem:free-direction} gives $\mathcal{S}_\lambda\subseteq\mathcal{S}^{\mathrm{myo}}_\lambda$. For the reverse inclusion, fix $S_b\in\mathcal{S}^{\mathrm{myo}}_\lambda$. Forward-closedness makes this set absorbing under every action. For any adapted continuation policy, with $i_k$ the arm it advances at step $k$ and $\mathcal F_k$ the history up to $S_k$, at every state reached before it stops
\[
 \E_\theta[h_\lambda(S_{k+1})-h_\lambda(S_k)\mid\mathcal F_k]
 =\EI_{i_k}(S_k)-\lambda\le0.
\]
Thus $h_\lambda(S_k)$ is a supermartingale. For an admissible stopping time $\tau$, apply optional stopping first to $\tau_n=\tau\wedge(b+n)$. The increments are bounded by $(y_{\max}-y_{\min})+\lambda$, hence
\[
 |h_\lambda(S_{\tau_n})|
 \le |h_\lambda(S_b)|+[(y_{\max}-y_{\min})+\lambda](\tau-b),
\]
an integrable dominating variable. Dominated convergence then yields
\[
 \E_{\theta,\pi}[h_\lambda(S_\tau)\mid S_b]\le h_\lambda(S_b).
\]
Immediate stopping attains the right-hand side, which gives $S_b\in\mathcal{S}_\lambda$. The two stopping regions coincide. Outside $\mathcal S^{\mathrm{myo}}_\lambda$, Lemma~\ref{lem:free-direction} makes immediate stopping strictly suboptimal; inside it, immediate stopping is optimal, with ties resolved in its favor. This proves the stop/continue classification; EI-greedy allocation away from the set needs a separate argument.
For $S_b\in\mathcal S^{\mathrm{myo}}_\lambda$, forward closure puts every successor in the same set almost surely; the region equality just proved makes each such successor a Bellman stopping state. The centered identity therefore has $O_i(S_b)=0$ for every available $i$.

\begin{proof}[Proof of Proposition~\ref{prop:depth-two-rollout}]
First, $\pi_{\rm EI}$ is admissible and no worse than stopping under the stated assumptions: from any state $S_b$, Lemma~\ref{lem:ei-beats-stop} gives $\E_{\theta,\pi_{\rm EI}}[\tau_{\rm EI}-b\mid S_b]\le d(S_b)/\lambda$ and $V_{\pi_{\rm EI}}(S_b)\ge h_\lambda(S_b)$.

Now fix $S_b\in\mathcal S^{\mathrm{myo}}_\lambda$ and force arm $i$. At a child
$S_{b+1}=F(S_b,i,Y_{b+1})$, if $\pi_{\rm EI}$ stops, then
$(g(S_{b+1})-\lambda)^+=0$. If it advances an EI-maximizing arm $j$,
Lemma~\ref{lem:ei-beats-stop} at the next state gives, since its first summand is $\EI_j(S_{b+1})-\lambda$ and
every later summand is positive because $\pi_{\rm EI}$ advances only where $g>\lambda$,
\[
V_{\pi_{\rm EI}}(S_{b+1})
\ge h_\lambda(S_{b+1})+\EI_j(S_{b+1})-\lambda
= h_\lambda(S_{b+1})+(g(S_{b+1})-\lambda)^+.
\]
The same inequality holds in the stopping case. Taking its root expectation gives
\[
Q_{\pi_{\rm EI}}(S_b,i)
\ge\E_\theta\!\left[h_\lambda(S_{b+1})+
(g(S_{b+1})-\lambda)^+\mid S_b,i\right]
=h_\lambda(S_b)+A^{\rm pool}_{2,i}(S_b).
\]
At the EI stop $V_{\pi_{\rm EI}}(S_b)=h_\lambda(S_b)$; subtracting this
common value gives the second inequality in~\eqref{eq:selective-lower-chain}.
\end{proof}

\subsection{H2 Stopping Improves on EI Stopping}
\label{app:receding-proof}
\begin{proof}[Proof of Corollary~\ref{cor:receding-policy-improvement}]
Let $T$ be the one-step policy-evaluation operator of $\pi_{\rm H2}$: $(TV)(S)=h_\lambda(S)$ where $\pi_{\rm H2}$ stops, and
$(TV)(S)=\E_\theta[V(F(S,i,Y))\mid S,i]$ where it advances $i$. All states below lie in the $\pi_{\rm H2}$-reachable subtree of the
initial state. Where $\pi_{\rm H2}$ takes the action of $\pi_{\rm EI}$, a common advance or a common stop,
$(TV_{\pi_{\rm EI}})(S)=V_{\pi_{\rm EI}}(S)$. Otherwise $S$ is an EI stop at which the gate opens, where $V_{\pi_{\rm EI}}(S)=h_\lambda(S)$, and Proposition~\ref{prop:depth-two-rollout} with the gate of
Algorithm~\ref{alg:selective-receding} (line~\ref{line:gate}) gives
\[
(TV_{\pi_{\rm EI}})(S)=Q_{\pi_{\rm EI}}(S,\kappa(S))\ge h_\lambda(S)+A^{\rm pool}_{2,\kappa(S)}(S)
\ge h_\lambda(S)+\underline A^{\rm pool}_{2,\kappa(S)}(S)>V_{\pi_{\rm EI}}(S).
\]
Thus $TV_{\pi_{\rm EI}}\ge V_{\pi_{\rm EI}}$, and since $T$ is monotone, $T^nV_{\pi_{\rm EI}}$ is nondecreasing in $n$. By induction on
the cap, $(T^{n}V)(S)=V_{\pi_{\rm H2}}(S)$ for every bounded $V$ and every $n>H(S)$: when $H(S)=0$ no arm is available and
$TV=h_\lambda=V_{\pi_{\rm H2}}$ at $S$, and every successor $S'$ has $H(S')\le H(S)-1$. Hence
$V_{\pi_{\rm H2}}=T^{n}V_{\pi_{\rm EI}}\ge V_{\pi_{\rm EI}}$ on the subtree. At a state where the gate opens,
$V_{\pi_{\rm H2}}(S)=(TV_{\pi_{\rm H2}})(S)\ge(TV_{\pi_{\rm EI}})(S)>V_{\pi_{\rm EI}}(S)$ by the display.
Since $V_{\pi_{\rm H2}}\ge V_{\pi_{\rm EI}}\ge h_\lambda$ on the subtree, Proposition~\ref{prop:existence} also bounds
the expected number of further increments of $\pi_{\rm H2}$ by $d(S)/\lambda$.
\end{proof}
The argument is policy improvement~\citep{tesauro1996online,bertsekas1997rollout}: with exact values of the base
policy, departing only where an action beats the base policy's value needs no further condition on the base policy.
Lemma~\ref{lem:ei-beats-stop} enters only through Proposition~\ref{prop:depth-two-rollout}, which bounds
$Q_{\pi_{\rm EI}}$ from below without simulating $\pi_{\rm EI}$ and so makes each departure certifiable.

\subsection{Slope-Screened Endpoint Enclosures}
\label{app:slope-endpoint-proof}
The notation is that of Section~\ref{sec:analytical-enclosure}, and the statement below contains Proposition~\ref{prop:endpoint-main}; no monotonicity in $\sigma$ is assumed.
\begin{proposition}[Slope-screened endpoint enclosure]
\label{prop:slope-endpoint}
For $X\sim\mathcal N(\mu,\sigma^2)$ restricted to $[0,1]$, fixed $\sigma>0$,
and $c\in[0,1]$, let $\psi(\mu,\sigma,c)=\E[(X-c)^+]$. Then
\[
0\le\partial_\mu \psi\le\Prob(X>c),\qquad
\partial_c \psi=-\Prob(X>c).
\]
Consequently $f_j(z)=\psi(\mu_j(z),\sigma_j,\max\{q,z\})$ is nonincreasing
on $[0,1]$ if $\gamma_j\le0$, with a maximum at $0$. If $0\le \gamma_j\le1$, it is
nondecreasing below $q$ and nonincreasing above $q$, with a maximum at $q$.
Plateaus and $q\in\{0,1\}$ are included. Suppose directed-rounding intervals
enclose the coefficients and the entire slope interval lies in one of these
cases. On a cell $[z_-,z_+]$ not crossing $q$, rigorous endpoint bounds $\underline f_j,\overline f_j$
give $[\underline f_j(z_-),\overline f_j(z_+)]$ in the increasing case and $[\underline f_j(z_+),\overline f_j(z_-)]$ in the
decreasing case, enclosing $f_j(z)$ for every $z$ in that cell and every true
enclosed coefficient tuple, including correlated coefficients.
\end{proposition}
\begin{proof}
Write $p_\mu$ for the normalized restricted-normal density,
$\chi(x)=(x-c)^+$, $\psi=\E[\chi(X)]$, and $P=\Prob(X>c)$. Differentiating the
normalizer as well as the Gaussian numerator gives
\[
\partial_\mu p_\mu(x)=\frac{x-\E[X]}{\sigma^2}p_\mu(x),\qquad
\partial_\mu \psi=\frac{\operatorname{Cov}(\chi(X),X)}{\sigma^2}\ge0.
\]
Indeed, for an independent copy $X'$, twice this covariance equals
$\E[(\chi(X)-\chi(X'))(X-X')]\ge0$. Since
$p_\mu'(x)=-(x-\mu)p_\mu(x)/\sigma^2$, integration by parts yields
\begin{align*}
\frac{\operatorname{Cov}(\chi(X),X)}{\sigma^2}
&=\E[\chi'(X)]-p_\mu(1)(\chi(1)-\psi)-p_\mu(0)(\psi-\chi(0))\\
&=P-p_\mu(1)(1-c-\psi)-p_\mu(0)\psi\le P.
\end{align*}
Both boundary corrections are nonnegative because $0\le \psi\le1-c$.
Bounded support and positive fixed scale justify differentiation under the
integral; the kink has probability zero, giving $\partial_c \psi=-P$.
Since $\partial_\mu \psi\in[0,P]$, the chain rule gives $f_j'=\gamma_j\,\partial_\mu\psi$ below
$q$ and $f_j'=\gamma_j\,\partial_\mu\psi-P$ above $q$. Their signs prove both cases; continuity
joins them at $q$ and gives the support-endpoint cases by one-sided limits.

For each fixed true coefficient tuple, these monotonicity statements hold
throughout each cell. Directed endpoint evaluations enclose that tuple's two
endpoint values, hence its entire cell range. Enclosing correlated coefficients by intervals only enlarges the admissible set; the argument therefore applies to each true tuple held fixed across observations.
The same reasoning bounds every tuple's peak, proving the pruning test,
including equality. Whole-cell interval evaluation covers all unclassified
slopes. Taking the maximum with stop and all remaining feasible child excesses
preserves pointwise enclosure. The true root-mass vector belongs to the set
$\mathcal P$ of mass vectors within the rigorous cell bounds (Appendix~\ref{app:enclosure-procedure}); optimizing over $\mathcal P$ thus gives the displayed integral
bounds even when masses and excesses are dependent. Adding the enclosed root
EI minus cost proves the selected-root advantage enclosure.
\end{proof}

\subsection{Same-Arm Reduction and Per-Arm Conditioning}

\begin{proof}[Proof of Corollary~\ref{cor:same-arm-reduction}]
At an EI stop, no cross-arm resurrection makes every available child arm
$j\ne i$ satisfy $\EI_j(S_{b+1})\le\lambda$ almost surely. Hence
\[
(g(S_{b+1})-\lambda)^+=(\EI^{\rm cap}_i(S_{b+1})-\lambda)^+
\quad\text{almost surely}.
\]
Substitution in Equations~\eqref{eq:depth-two-advantage}
and~\eqref{eq:same-arm-depth-two-advantage} gives equality. Without the condition,
the pointwise pool maximum still dominates the same-arm term.
\end{proof}

\paragraph{Proof of Proposition~\ref{prop:gp-pool}.}
For an unadvanced arm $j$, its conditioning observations, local fidelity and any deterministic per-arm state are unchanged. Fixed kernel hyperparameters, truncation limits, and variance floor leave its complete predictive law unchanged. Arm-local availability ensures that every $j\ne i$ available at the successor was already available before advancing $i$. Because $(y-c)^+$ is pointwise non-increasing in $c$ and the incumbent cannot decrease, $\EI_j$ cannot increase. Since $\EI_j(S_b)\le\lambda$ at a state in $\mathcal S^{\mathrm{myo}}_\lambda$, $\EI_j(S_{b+1})\le\lambda$, which is Assumption~\ref{assm:no-resurrection}. That assumption concerns only $j\ne i$ and leaves the refreshed predictive law of the advanced arm unconstrained.
The finite-history state, deterministic per-arm state update, fixed kernel, and analytic truncated-Gaussian law define measurable state and predictive updates, satisfying Assumption~\ref{assm:fixed-sim}.

\section{Surrogates and Certificate Implementation}
\label{app:impl}

\subsection{GP-at and GP-atJ}
\label{app:shared-temporal-b}
Section~\ref{sec:shared-models} gives the model and Section~\ref{sec:experiments} its scales $(1,8,64,512)$ and
variances $v_a=v_t=1/32$; the asymptote block thus has covariance $C_a/32$ and each of the four transient coefficient
blocks $C_t/128$. The scales are geometric, hence one basis covers curves of one to several hundred checkpoints without
estimating a rate from scores, and integer scales keep the enclosure inputs of Section~\ref{sec:analytical-enclosure}
exact. The observation variance is $2^{-16}+2^{-10}/(1+t)^2+2^{-30}$ and the predictive SD floor $2^{-16}$, below the
scale implied by the fresh noise. The Mat\'ern matrix $R$ uses lengthscales equal to the median positive pairwise gaps
of the configuration arrays. All values were fixed before any result, are shared by both surrogates and all benchmarks, and
are fitted to no curves; revealed scores enter only through Gaussian conditioning. Fixed Gaussian conditioning gives
fantasy means affine in the root observation and fantasy-independent scales, and sequential one-step truncation to
$[0,1]$ gives the planning law of Section~\ref{sec:analytical-enclosure}. Directed-rounding coefficient
construction and a verified inverse-residual contraction bound supply the root and child enclosures of the gate.

\subsection{Certificate Procedure, Cost and Claim Scope}
\label{app:enclosure-procedure}
We compile the full feasible child roster after the root, respecting both
arm-local and remaining increment caps, and certify positive scales. We
classify a raw MPFR~\citep{fousse2007mpfr} slope interval only when its upper endpoint is at most
zero, or its lower endpoint is at least zero and upper endpoint at most one;
otherwise we retain whole-cell interval EI evaluation. For a
classified child, we evaluate its global peak at $0$ or $q$ and prune its
excess only if the rigorous peak upper endpoint is at most $\lambda$, imported
as the exact binary input; equality suffices, since then $(f_j(z)-\lambda)^+=0$ everywhere.
We partition $[0,1]$ at $q$ as well as the outer grid points, memoize endpoint evaluations, and form
on each cell $C$ an interval $[\underline\rho_C,\overline\rho_C]$ for
\[
\rho_i(z)=\max\bigl(\{0\}\cup
\{f_j(z)-\lambda:j\in\cI(F(S,i,z))\}\bigr).
\]
The stop option remains zero; an empty feasible child roster gives exactly zero excess.
With $Z$ the root observation and rigorous root-mass bounds $\underline p_C,\overline p_C$, we compute
\[
\min_{p\in\mathcal P}\sum_C p_C\underline\rho_C
\le \E_\theta[\rho_i(Z)\mid S,i]\le
\max_{p\in\mathcal P}\sum_C p_C \overline\rho_C,\qquad
\mathcal P=\{p:\underline p_C\le p_C\le\overline p_C,\ \sum_Cp_C=1\},
\]
a self-validating quadrature~\citep{corlissrall1987,johansson2018integration} that
encloses \eqref{eq:depth-two-advantage} with the child maximum inside the expectation.
Root-normalizer checks are retained even when all children are screened out. A tolerance-unmet but valid
enclosure remains rigorous; numerical exceptions, including a failed interval Cholesky factorization, fail closed:
the gate stays closed and the EI outcome is kept. The compiler encloses the frozen affine Gaussian updates and rejects
cases unless variance-floor inactivity is proved. The selected-root lower bound certifies $A^{\rm pool}_{2,i}>0$, and an
open gate concerns the selected root alone. Every study uses the gate of Algorithm~\ref{alg:selective-receding}
(line~\ref{line:gate}), and the bound is certified anew at every EI stop, without assuming forward closure.

The point planner estimates $A^{\rm pool}_{2,i}(S)$ for each feasible root by floating-point quadrature, allowing every feasible child arm. It chooses $\kappa(S)$ with the largest estimate and certifies that root alone. Selection directs the computation toward a promising departure; the rigorous lower bound determines whether to take it.

\paragraph{Cost.}
Under GP-at and GP-atJ the posterior carries $K+1$ coefficients per configuration, and its dimension $N(K+1)$ thus does not grow with the number of observations. An EI decision conditions on one observation by a rank-one
update and evaluates the $N$ arms' EI, in $O(N^2)$ operations. Only an EI stop reaches the gate.
There the point planner evaluates, for each of the $N$ roots, the EI of every feasible child at $m_{\rm q}$ quadrature nodes,
in $O(N^3+m_{\rm q}N^2)$ floating-point operations. The certificate inverts the $N(K+1)$-dimensional posterior
precision with verified residual bounds, encloses the $N$ root EIs on at most $m_{\rm root}$ cells each and the
selected root's excess on $m_{\rm out}$ outer cells with at most $m_{\rm child}$ cells per child EI, in
$O(N^3+(m_{\rm root}+m_{\rm out}m_{\rm child})N)$ interval operations at fixed precision. The gate therefore costs a
polynomial in $N$, independent of the remaining budget and the horizon, and an episode of $\tau$ increments
that reaches the gate $G$ times costs $O(\tau N^2+G(N^3+m_{\rm q}N^2))$ floating-point operations and $G$ certificates. The studies use $N=32$, $K=4$,
$m_{\rm q}\le32+64+128$ nodes over three quadrature orders, $m_{\rm root}=64$, $m_{\rm out}=32$, $m_{\rm child}=4$ and 256-bit
precision.

\paragraph{Scope of the Three Claims.}
The \emph{ideal policy comparison} concerns the fixed planning law. The
\emph{numerical certificate} verifies the EI boundary and the selected root's
advantage from the exact binary state inputs, after floating-point EI and the point planner choose the EI stopping
action and the candidate. Transferring Corollary~\ref{cor:receding-policy-improvement} to the floating-point policy
additionally requires all EI stopping actions to agree with exact-arithmetic EI throughout its reachable subtree, beyond
the replay path. The \emph{empirical comparison} asks whether the certified departure is profitable on benchmark curves, which
must be tested even when the expectation under the planning law is positive.

\section{Experimental Protocol}
\label{app:provenance}

\subsection{Benchmarks}
\label{app:pd1-results}
\emph{LCBench.} All 35 tasks of the \texttt{data\_2k\_lw} release~\citep{zimmer2021autopytorch}, with 2,000
configurations each and one checkpoint per epoch over 50 epochs; the score is validation accuracy.

\emph{PD1.} We use 15 tasks of \citet{wang2024hyperbo} and the 242 of 400 configurations with complete,
non-diverged phase-1 records on every task. Eight further tasks carry at most six usable checkpoints each, against 12
to 1,413 for the 15 retained, and are excluded because replay caps would dominate the stopping decision there;
the exclusion uses only native horizon length and is applied when the benchmark cache is built. The shared
intersection gives every comparison the same finite pool but conditions evaluation on configurations that survive on
every task. We replay stored scores without interpolation and drop every curve's final observation, which follows a
shorter terminal interval. Of the 15 tasks, 13 use 1,000-optimizer-step spacing between retained checkpoints, LM1B 29,590
and UniRef50 10,000; one checkpoint, and its $\lambda$ charge, therefore denotes different amounts of training on different tasks. The three
tasks with more than 100 checkpoints, the two ImageNet ResNet workloads and the WMT translation Transformer, have 250 to
1,413; the other 12 tasks have 12 to 52. The four hyperparameter coordinates are transformed and min--max scaled once over
the 242-point candidate set.

\emph{ODBench.} The object-detection learning-curve dataset of \citet{cfbo2025}, built on datasets from Roboflow
100~\citep{ciaglia2022roboflow}; the paper leaves it unnamed, and we use the name of its public release. Each of the
release's 30 tasks pairs a dataset with one of three detectors, Faster R-CNN with a ResNet-50 or HRNet backbone or
SSDLite with MobileNetV2, and holds 500 configurations of batch size, learning rate, momentum and weight decay trained
for 20 epochs. We use nine tasks, datasets 8--10 with each detector, and their native bounded validation score.
Hyperparameter coordinates are normalized from the known candidate values. ODBench is held out: nothing is fitted or
selected on it.

\subsection{Pools, Seeds and Episodes}
Each seed draws one pool per task: $N=32$ configurations, uniformly without replacement, from the task's candidate
set. Every method and comparator of a task--seed pair runs on that pool. One method under one surrogate on seeds 1--50 is
$20{,}650$ episodes, $12{,}250$ on LCBench, $5{,}250$ on PD1 and $3{,}150$ on ODBench (tasks $\times$ 50 seeds
$\times$ 7 costs). EI stopping and H2 stopping run on seeds 1--50 under GP-a, GP-at and GP-atJ. The H2 stopping episode reproduces the actions of EI stopping until its gate first opens; the EI outcome of each episode is thus its exact pre-gate prefix.
H2 stopping under GP-atJ also runs on fresh seeds 51--100, drawn from a stated generator before the pre-registered test
(Appendix~\ref{app:primary50}), and under GP-at and GP-a on the same seeds (Appendix~\ref{sec:shared-results}). The comparators of Appendix~\ref{app:published} run on the pools of seeds 1--50.

\subsection{Seed-Block Bands and Noninferiority Limits}
\label{app:shared-uncertainty}
The bands use the bootstrapped maximum studentized deviation~\citep{romano2005snooping}. For each benchmark, let $X_{\ell j}$ be the equally task-averaged paired utility contrast of seed $\ell$ in column $j$,
one contrast at one cost. Because the same $50$ seeds identify pools across all tasks, costs and methods, we resample whole
seed blocks with tasks held fixed: each replicate applies one multinomial count vector to the entire block matrix,
which preserves the dependence across tasks and the pairing across costs. The bands therefore quantify seed--pool
variability within the fixed $35/15/9$ task rosters. With $\mathrm{se}_j=\operatorname{sd}(X_{1:50,j})/\sqrt{50}$
and $10{,}000$ replicates, the two-sided simultaneous band is $\bar X_j\pm c\,\mathrm{se}_j$, with $c$ the
95th percentile (higher order statistic) of $\max_j|\bar X_j^*-\bar X_j|/\mathrm{se}_j$ over the columns with
$\mathrm{se}_j>0$; the one-sided upper confidence limit (UCL) is $\bar X_j+c^+\mathrm{se}_j$, with $c^+$ the
95th percentile of $\max_j(\bar X_j^*-\bar X_j)/\mathrm{se}_j$. Coverage is approximate~\citep{romano2005stepwise}.
EI stopping is noninferior in a cell when the UCL of $\Delta_{\text{H2-S}}$ lies below $\delta=\lambda$; a band that
contains zero establishes neither.

The families are: for each surrogate on seeds 1--50 and for GP-atJ on seeds 51--100, the 21 benchmark--cost cells
(for GP-atJ on seeds 1--50, critical values 3.14 two-sided and 3.04 one-sided; its two-sided bands come from a rerun
of every episode on another platform, in which one cell mean, LCBench at $\lambda=0.0002$, differs from the stored one by
$2\times10^{-7}$); for the horizon split, 28 columns per seed set; for the published surrogates, the 21 cells of each
comparison; for the surrogate-scale study of Table~\ref{tab:law-scale-at-atj}, its 126 columns. The surrogate-sensitivity bands are
pointwise. The bands concern replayed
utility; agreement between ideal EI stopping actions and the executed floating-point policy is a separate question.

\subsection{Reproducibility, Computing Infrastructure and Assets}
\label{app:reproducibility}
Code is available at \url{https://github.com/longvus/EIS}; full episode records are available from the authors on request.

\paragraph{Computing Infrastructure.}
Every episode is a table-lookup replay of released benchmark curves. The experiments ran under Python 3.12 on an
8-core Intel Core i7-11800H at 2.30\,GHz with 16\,GB of memory and Linux 6.8 (x86-64) and on an 8-core Apple M2 with
8\,GB of memory and macOS 14 (arm64). Gaussian conditioning and certified enclosure run on CPU. The DyHPO comparator
ran on a single NVIDIA RTX 3080 Laptop GPU with 8\,GB of memory. The studies on the published surrogates, with their
ifBO and CFBO-NT comparators, ran on NVIDIA A30, A100, V100, T4 and L40S GPUs and AMD EPYC 7702 CPUs of a university
cluster.

\paragraph{Assets and Licenses.}
The LCBench curves~\citep{zimmer2021autopytorch} and the PD1 workload collection~\citep{wang2024hyperbo} are
distributed under Creative Commons Attribution 4.0; the separate LCBench code repository, which we do not use, is
Apache 2.0. The ifBO/FT-PFN comparator uses the \texttt{ifbo} package (MIT License) and the DyHPO comparator the
implementation of \citet{wistuba2022dyhpo} (Apache License 2.0). The ODBench curves and the CFBO comparator
come from the public release of \citet{cfbo2025}, which states no license: its repository carries no license file, and
its data is distributed through a file-sharing link. Their paper attributes its tasks to the collection of
\citet{ciaglia2022roboflow}, and the curve files carry no provenance metadata. We redistribute none of these releases;
each benchmark is obtained from its original source.

\section{Results under the Controlled Surrogates}
\label{app:controlled-results}

\subsection{F1: Exact In-Model Gap on Reduced Instances}
\label{app:inmodel-oracle}
Each instance freezes an EI stopping episode under GP-atJ on a pool of seeds 1--50 at its first EI stop and keeps the available arms
with the largest EI: three arms with two further increments each ($3{\times}2$ instances), or, on the same frozen state, two arms
with three further increments each ($2{\times}3$ instances), so that a gain needing three increments can exist. Under the planning law each
revealed score updates the Gaussian posterior by the Gaussian rule, and each one-step predictive is that posterior's
Gaussian restricted to $[0,1]$ (Section~\ref{sec:analytical-enclosure}). Here each such predictive is replaced by
five equal-probability points at its bin means, which preserves every predictive mean and defines a fixed planning
law. Each point is a conditional mean, so by Jensen's inequality the replacement can only lower an EI, and the
frozen state is an EI stop under the discretized law as well; backward induction over the full outcome tree then gives $V_\lambda^\star$, $V_{\pi_{\rm EI}}$ and $V_{\pi_2^{\rm ideal}}$ exactly, where $\pi_2^{\rm ideal}$, \emph{ideal H2 stopping}, is
$\pi_{\rm H2}$ with the exact advantage $A^{\rm pool}_{2,\kappa(S)}$ in place of its certified bound; the ordering
$V_\lambda^\star\ge V_{\pi_2^{\rm ideal}}\ge V_{\pi_{\rm EI}}$ is checked at every node.

\paragraph{What the Reduction Bounds.} Dropping arms and increments only removes options. Under the discretized planning law,
each instance's gap $V_\lambda^\star-V_{\pi_{\rm EI}}$ is therefore a lower bound on the gap of the full pool with its
full horizons, and the optimal and within-$\lambda$ shares are upper bounds on the full pool's. The instances can thus
show that an EI stop is premature, but they cannot show that the stop is optimal in the full problem. These
bounds are statements about the discretized planning law.

\paragraph{Pre-Registration.} The draw, 40 task--seed pairs per benchmark and cost, 840 instances, and seven
expectations were fixed before any instance was solved, after a pilot draw of 180 instances (protocol in the code release). Table~\ref{tab:inmodel-oracle-prereg} reports every estimand. Five expectations were met, on
the $3{\times}2$ instances: an optimal share of at least $70\%$ ($72.5\%$), with lower Wilson~\citep{wilson1927probable} limit at
least $65\%$ ($69\%$); a lower Wilson limit of the within-$\lambda$ share of at least $85\%$ ($88\%$); recovery by ideal H2 stopping of at least $95\%$ ($99.2\%$); and an optimal share of at least $60\%$ on every benchmark ($71.1$
to $74.6\%$). Two were missed: the within-$\lambda$ share on the $3{\times}2$ instances, $89.8\%$ against $90\%$, and recovery on the
$2{\times}3$ instances, $85.5\%$ $[81,89]$ against $90\%$.

\paragraph{Results.} On the $2{\times}3$ instances the EI stop is optimal in $60.8\%$ of instances and within $\lambda$ in
$75.2\%$, against $72.5\%$ and $89.8\%$ on the $3{\times}2$ instances, and two-step lookahead recovers $85.5\%$ of the gap
against $99.2\%$: delayed gains beyond two increments exist under the planning law and are not all recovered by two-step
lookahead. The gap depends on cost (rows by $\lambda$ in Table~\ref{tab:inmodel-oracle-prereg}). At
$\lambda\ge0.005$ the EI stop is optimal in $82.5$ to $85.8\%$ of $3{\times}2$ and $75.8$ to $80.8\%$ of
$2{\times}3$ instances and within $\lambda$ in at least $91.7\%$ on either; at the five lower costs it is optimal in
$59.2$ to $75.0\%$ and $51.7$ to $57.5\%$. Exact metalevel computations show the same dependence on
cost~\citep{callaway2018learning}. The rows by cost were not pre-registered.
\InModelOraclePreregTable

\paragraph{Full Scale.} The reduced instances bound the full pool's gap from one side only. A second pre-registered
study bounds it from both sides at the same 840 states, with all 32 arms and every recorded increment
(Table~\ref{tab:full-scale-bounds}). A long-lookahead policy $\pi_{\rm LA}$, which advances the arm for which the expected
improvement of its $n$-th further score over the incumbent most exceeds $n\lambda$, over a grid of $n$ up to the arm's
remaining increments, and runs for at most 300 increments, gives a Monte Carlo lower bound under the planning law
(its index uses the Gaussian marginal of the $n$-th score, a proxy that affects the policy's quality but not the
validity of the bound); an information relaxation that reveals the component the arms share, with a martingale
penalty~\citep{brownsmithsun2010}, gives an upper bound. The rebuilt states match the reduced study's, and on the reduced
instances the full-scale code reproduces the exact optimum. At every cost up to $0.001$, and at $0.002$ on PD1 and
ODBench, $\pi_{\rm LA}$ beats the EI stop by more than one increment's cost: once the whole pool and horizon are in play,
the EI stop is premature on average under the planning law, and the pre-registered statement that it is near-optimal at
full scale is contradicted there. At the two highest costs $\pi_{\rm LA}$ gains less than one increment's cost, as the
reduced instances suggest, but no cell can be certified near-optimal, because the relaxation, and the term that carries
it over to the planning law's restriction to $[0,1]$, leave the upper bound loose. At low cost, then, lookahead over more
arms and increments finds value under the planning law that two steps on three arms miss.

\paragraph{On Recorded Curves.} Two pre-registered replays run $\pi_{\rm LA}$ on the recorded curves, from the same
states and from 840 fresh first EI stops on seeds 51--100. Less of the value under the planning law arrives than it
promises: on ODBench the realized gain is a small fraction of it, and on LCBench and PD1 it is closer in size but too
variable across tasks to establish in most cells. The replay pays only at the lowest costs, where an increment is cheapest, and in too few premature cells
for either pre-registered statement, which reads mixed on seeds 1--50 and inconclusive on fresh seeds; the pre-registered
pooled test on fresh seeds finds a small positive gain. On LCBench the shortfall agrees with the calibration of the gate,
whose realized gains fall short of even the certified lower bound (Appendix~\ref{app:primary50}).
\FullScaleBoundsTable

\subsection{F2: Certified Continuation under GP-atJ}
\label{app:primary50}
\begin{table}[ht]\centering\scriptsize\setlength{\tabcolsep}{3pt}\renewcommand{\arraystretch}{0.8}
\caption{H2 stopping (H2-S) minus EI stopping (EI-S) under GP-atJ: on seeds 1--50, the pools of every other table (descriptive, with no tolerance attached), and the two-sided simultaneous 95\% bands of the pre-registered test of Table~\ref{tab:body-effects} ($^\dagger$/$^\ddagger$: above/below zero). Gate and UCL as in Table~\ref{tab:body-effects}.}
\label{tab:body-effects-dev}
\begin{tabular}{@{}lrrrrr@{}}\toprule
 & & \multicolumn{3}{c}{Seeds 1--50} & Pre-registered test \\ \cmidrule(lr){3-5}\cmidrule(lr){6-6}
Bench. & $\lambda$ & Gate & H2-S $-$ EI-S [band] & UCL & H2-S $-$ EI-S [band] \\ \midrule
LCBench & 0.0001 & 34\% & $+0.00014$ [-0.00008, +0.00036] & $+0.00035$ & $+0.00003$ [-0.00015, +0.00021] \\
 & 0.0002 & 25\% & $+0.00009$ [-0.00018, +0.00035] & $+0.00034$ & $-0.00001$ [-0.00017, +0.00015] \\
 & 0.0005 & 21\% & $-0.00047$ [-0.00131, +0.00036] & $+0.00034$ & $-0.00046$ [-0.00083, -0.00009]$^{\ddagger}$ \\
 & 0.001 & 18\% & $-0.00049$ [-0.00072, -0.00025]$^{\ddagger}$ & $-0.00025$ & $-0.00038$ [-0.00059, -0.00017]$^{\ddagger}$ \\
 & 0.002 & 12\% & $-0.00042$ [-0.00080, -0.00003]$^{\ddagger}$ & $-0.00005$ & $-0.00052$ [-0.00079, -0.00026]$^{\ddagger}$ \\
 & 0.005 & 3\% & $-0.00017$ [-0.00047, +0.00012] & $+0.00011$ & $-0.00016$ [-0.00035, +0.00002] \\
 & 0.01 & 4\% & $-0.00069$ [-0.00155, +0.00018] & $+0.00015$ & $-0.00071$ [-0.00168, +0.00026] \\
\midrule
PD1 & 0.0001 & 39\% & $+0.00117$ [+0.00010, +0.00223]$^{\dagger}$ & $+0.00220$ & $+0.00135$ [-0.00008, +0.00278] \\
 & 0.0002 & 43\% & $+0.00311$ [+0.00111, +0.00511]$^{\dagger}$ & $+0.00505$ & $+0.00358$ [+0.00151, +0.00565]$^{\dagger}$ \\
 & 0.0005 & 31\% & $+0.00059$ [-0.00069, +0.00186] & $+0.00182$ & $+0.00041$ [-0.00017, +0.00099] \\
 & 0.001 & 18\% & $+0.00082$ [-0.00087, +0.00250] & $+0.00245$ & $+0.00064$ [-0.00083, +0.00212] \\
 & 0.002 & 12\% & $+0.00100$ [-0.00100, +0.00300] & $+0.00293$ & $+0.00104$ [-0.00072, +0.00281] \\
 & 0.005 & 10\% & $+0.00118$ [-0.00061, +0.00297] & $+0.00291$ & $+0.00172$ [-0.00036, +0.00380] \\
 & 0.01 & 6\% & $+0.00253$ [+0.00129, +0.00377]$^{\dagger}$ & $+0.00374$ & $+0.00129$ [-0.00015, +0.00273] \\
\midrule
ODBench & 0.0001 & 24\% & $+0.00073$ [-0.00004, +0.00149] & $+0.00147$ & $+0.00106$ [+0.00004, +0.00209]$^{\dagger}$ \\
 & 0.0002 & 19\% & $+0.00118$ [-0.00030, +0.00266] & $+0.00261$ & $+0.00089$ [-0.00011, +0.00189] \\
 & 0.0005 & 24\% & $+0.00109$ [-0.00022, +0.00240] & $+0.00236$ & $+0.00090$ [-0.00119, +0.00298] \\
 & 0.001 & 33\% & $+0.00221$ [-0.00022, +0.00464] & $+0.00456$ & $+0.00046$ [-0.00105, +0.00198] \\
 & 0.002 & 28\% & $+0.00056$ [-0.00122, +0.00234] & $+0.00229$ & $+0.00031$ [-0.00247, +0.00309] \\
 & 0.005 & 19\% & $-0.00067$ [-0.00241, +0.00107] & $+0.00102$ & $-0.00093$ [-0.00234, +0.00048] \\
 & 0.01 & 5\% & $-0.00013$ [-0.00067, +0.00042] & $+0.00040$ & $-0.00018$ [-0.00058, +0.00022] \\
\bottomrule\end{tabular}\end{table}

The contrast is H2 stopping minus EI stopping under GP-atJ, with tasks weighted equally within each benchmark
(Table~\ref{tab:body-effects} for the fresh seeds, Table~\ref{tab:body-effects-dev} for seeds 1--50). On seeds 1--50 the two-sided bands carry no equivalence or noninferiority margin; the
one-sided limits measure how much certified continuation can recover. \BCeilingFinding\ No tolerance is
attached to these UCLs, which use the seeds behind every other result; the verdicts come from the
pre-registered test.

\paragraph{Pre-Registered Noninferiority on Fresh Seeds.} Before the first fresh-seed episode, the pre-registration
fixed the tolerance $\delta=\lambda$, one increment's cost; the family, the $21$ benchmark--cost cells; the decision rule
UCL $<\delta$ with the one-sided bootstrap of Appendix~\ref{app:shared-uncertainty} at a new bootstrap seed; 50 fresh episode seeds disjoint from seeds 1--50; and an expectation read from the seeds 1--50 UCLs: noninferiority on
LCBench at every cost but $0.0001$ and $0.0002$, and on PD1 and ODBench at $0.005$ and $0.01$
alone. \BNoninferiorityFinding\ The verdicts match the expectation on PD1 and ODBench and exceed it by one cell on
LCBench, where the fresh-seed limit at $\lambda=0.0002$ fell below one increment's cost. In two open cells, PD1 at
$\lambda=0.0002$ and ODBench at $0.0001$, the two-sided band lies above zero, and certified continuation thus gains there; in the other nine the limit exceeds one increment's cost while the band contains zero. The UCL of a cell is the
smallest tolerance at which it would read noninferior, and Table~\ref{tab:body-effects} therefore admits any other tolerance; only
$\delta=\lambda$ was pre-registered.

\paragraph{Calibration of the Gate.} An open gate is a proof under the planning law: the certified lower bound where the
gate first opens bounds the expected gain of continuing under the planning law. The recorded curves show how much of it arrives
(Table~\ref{tab:gate-calibration}). \GateCalibrationFinding\ The means hide how the gains are spread: on every benchmark
most departures lose on the recorded curve, and on PD1 and ODBench a minority of large gains carries the mean above the
bound. On LCBench the planning law overstates what continuing is worth, which is where H2 stopping's band falls below
zero at the middle costs (Table~\ref{tab:body-effects}). The certificate is sound under the planning law; what a departure
earns depends on how well the planning law predicts the next increments.
\GateCalibrationTable

\subsection{F3: Noninferiority across Surrogates}
\label{sec:shared-results}
Table~\ref{tab:noninferiority-by-law} applies the statistic and tolerance of the pre-registered test to every surrogate run with H2 stopping, on seeds 1--50 and the same pools. These are the seeds on which the surrogates were developed, so the verdicts are descriptive, and F3 reads the table for the shape of the limit across costs rather than for the verdicts: \BLimitRatioSentence. \BNoninferiorityByLawFinding

\paragraph{Cross-Surrogate Test on Fresh Seeds.}
\label{par:cross-law-fresh}
To make the cross-surrogate reading confirmatory, a second pre-registered study reruns EI stopping and H2 stopping under GP-at on the fresh seeds 51--100 of the GP-atJ test, with the same pools, statistic, tolerance and decision rule, and with the shape of the limit across costs declared as a hypothesis before the first fresh episode: under GP-at the ratio above stays at or below $2$ on every benchmark. The pre-registration, seed derivation and a parity gate that reproduces every seeds 1--50 block of Table~\ref{tab:noninferiority-by-law} to within $10^{-9}$ before any fresh episode runs are in the supplement. GP-a runs on the same fresh seeds and pools with the same statistic, and its verdicts are descriptive.
\CrossLawFreshParagraph
\CrossLawFreshTable
\ByLawNoninferiorityTable

\subsection{F3: Information Sharing and the Primary Surrogate}
\label{app:ei-surrogate-three-way}
GP-atJ is the primary surrogate because EI stopping attains the highest utility under it. Table~\ref{tab:ei-surrogate-three-way} sets EI stopping's utility under the three surrogates side by side on the same pools and seeds, paired by seed block, on the development seeds 1--50 and on the fresh seeds 51--100; the fresh seeds were not used in choosing. \EISurrogateThreeWayFinding\ \EISharingFinding
\EISurrogateThreeWayTable
\EISharingTestsTable

\subsection{Horizon and Saturation}
\label{app:horizon-split}
The retained PD1 tasks differ in native horizon by two orders of magnitude (Appendix~\ref{app:pd1-results}).
Table~\ref{tab:horizon-split} splits the GP-atJ contrast of Appendix~\ref{app:primary50} into the three PD1 tasks with
more than 100 checkpoints and the other 12, on both seed sets. The split is post hoc and its bands carry no verdict.
\BHorizonFinding\ The 12 short-horizon tasks, with 12 to 52 checkpoints, resemble LCBench's 50 and likewise
favor EI stopping wherever a band excludes zero.

Replay caps bind at the cheap costs, most often on the short PD1 tasks and on LCBench, and nowhere at the two highest
costs (Table~\ref{tab:horizon-split}). A binding cap truncates both rules at the same state, and leaving out the episodes
with a cap hit does not change the pattern (last column); on LCBench the mean rises slightly at the cheapest costs, as
the negative mean at $0.0005$ comes largely from the capped episodes.
\BHorizonTable

\subsection{Sensitivity to the Surrogate's Settings}
\label{app:fixed-law-sensitivity}
We rerun EI stopping under GP-atJ with one of its settings changed at a time: the fresh-observation noise variance (both its
constant and its decaying term) scaled by $\frac14$ or $4$, or both prior variances $v_a$ and $v_t$ scaled by $\frac12$
or $2$. Pools, seeds, costs, caps and the stopping rule are unchanged, and each variant is paired per episode with the
primary settings (Table~\ref{tab:fixed-law-sensitivity}).

No change helps everywhere: each moves mean utility by up to $0.047$, in a direction, judged by band, that
depends on the benchmark. Less fresh noise shortens episodes and helps on LCBench and PD1 but hurts on ODBench; more
noise lengthens them, as a lower cost would, and hurts on LCBench and PD1 while helping on ODBench at most costs. When the
arm with the largest EI lies below the incumbent, its EI grows with the predictive standard deviation, so cost and the
surrogate's scale are confounded: $\lambda$ is the formulation's single economic parameter, and the scale is a second
one, fixed in advance. Halving the prior variances helps on LCBench and ODBench and is mixed on PD1. Choosing a setting on
ODBench would compromise its held-out role.

The same rescalings, and a noise variance calibrated per task from the recorded curves of the other tasks of its
benchmark, were then applied to GP-at and GP-atJ together, paired per episode on seeds 1--50, to ask whether
GP-atJ's advantage under EI stopping depends on the surrogate's scale (Table~\ref{tab:law-scale-at-atj}).
\LawScaleAtAtJFinding\ The advantage thus survives every rescaling tried, with at most one band below zero.
\FLSensTable
\LawScaleAtAtJTable

\section{Score--Resource Tradeoffs on Published Surrogates}
\label{app:published}

\subsection{Comparators}
\label{app:comparator-impl}
Every comparator runs on the pool and seed of the matched GP-atJ episode, reads the same fixed-table scores
(Assumption~\ref{assm:bounded-perf}) and is capped at $B=300$ increments, following \citet{cfbo2025}. A per-task
rescaling by the extremes of the full curve tensor would reveal the task optimum; the utilities are therefore on our scale and
do not transfer to the numbers published with CFBO. FT-PFN accepts training steps only as fractions of the curve
length, and DyHPO's deep-kernel GP divides every budget by the benchmark's maximum; both are thus given each task's
native curve length, which our Gaussian surrogates never use. ifBO and DyHPO have no stopping
rule~\citep{ifbo2024,wistuba2022dyhpo} and use all $B$ increments; their actions do not depend on $\lambda$, and one trajectory per task and seed thus gives the utility at all seven costs. For ifBO and CFBO-NT the first
increment goes to a uniformly random configuration.

\paragraph{ifBO.} ifBO~\citep{ifbo2024} scores each configuration by MFPI-random, the probability that its score after
$h$ more steps exceeds $f_{\rm best}+10^{-u}(1-f_{\rm best})$, with $h$ uniform between one step and the curve length
(capped by the remaining budget) and $u\sim\mathcal U(1,4)$ redrawn at every step, and advances the selected
configuration by one step. As in the code of \citet{ifbo2024}, FT-PFN receives each configuration's curve under its own
identifier, numbered from one by first evaluation, and an unevaluated configuration is queried with
identifier zero. A partially trained configuration keeps its identifier. Up to 100 unevaluated configurations are drawn
per step, covering the whole 32-configuration pool. The budget is $B=300$, below the 1,000 steps
of \citet{ifbo2024}, since a 32-configuration pool cannot absorb them on any ODBench task or on two PD1
tasks.

\paragraph{CFBO-NT.} The no-transfer baseline of \citet{cfbo2025}, the only comparator with its own stopping rule,
keeps CFBO's acquisition, the expected utility improvement maximized over future increments, and its regret-based
stopping rule with the published adaptive threshold ($0.2$ when the improvement probability is $1/2$, Beta-CDF shape
$e^{-1}$). It uses FT-PFN in place of the full method's transfer-trained extrapolator, and the public code's context encoding, which gives no configuration its own identifier. Its utility $\tilde y_b-\alpha_{\rm CFBO}b/B$ with
$\alpha_{\rm CFBO}=300\lambda$ equals ours up to the cap. \citet{cfbo2025} define $0\le\alpha_{\rm CFBO}\le1$ and
evaluate $\alpha_{\rm CFBO}\le2^{-2}$; our costs give $\alpha_{\rm CFBO}$ from $0.03$ to $3$; $\lambda\ge0.001$ lies above the
evaluated grid and $\lambda\ge0.005$ outside the defined range.

\paragraph{DyHPO.} DyHPO~\citep{wistuba2022dyhpo} runs as released, through its own experiment loop. Its acquisition
is EI at the next fidelity over the best score observed at that fidelity (or, if none, at lower fidelities), with no
cost term, and it advances the selected configuration by one increment. Its deep-kernel GP is trained within each
episode (10 restarts of $1{,}000$ epochs, then $50$ per increment), and its feature scaler is refit on each pool; it therefore sees pool-relative hyperparameter features where the other methods use fixed configuration arrays.

\subsection{EI Stopping on FT-PFN}
\label{app:ftpfn-ei}
\emph{Policy.} The rule uses ifBO's FT-PFN checkpoint and curve-identifier tokenization, and the pools, episode seeds
and policy seeds of the ifBO comparator. Because FT-PFN cannot be evaluated on an empty context, the first increment is
ifBO's random initialization. At every later step the rule scores each configuration with another increment
available by FT-PFN's EI of that increment over the best observed score, computed exactly from the bar distribution,
advances the largest with ties to the lowest index, and stops at the first step with $\max_i\EI_i\le\lambda$ or at
$B$. Its actions before the stop do not depend on $\lambda$, and one trajectory per episode thus gives all seven costs.

\emph{Results} (Tables~\ref{tab:body-f4} and~\ref{tab:learned-ei}). Relative to ifBO run to $B$, EI stopping reduces the score by $0.4$ to $10.9$ percentage points and saves $106$ to $292$ increments. The priced savings exceed the score reduction with a utility-difference band above zero in every cell except the lowest ODBench cost. The break-even costs value these observed exchanges, holding each cell's outcomes fixed. Against the adaptive CFBO-NT reference on the same weights, EI stopping ends with higher scores and utility-difference bands above zero in every cell, both inside and outside the defined $\alpha_{\rm CFBO}$ range; its increment difference ranges from $-33$ to $+45$.
\begin{table*}[t]\centering\scriptsize\setlength{\tabcolsep}{3pt}\renewcommand{\arraystretch}{0.9}
\caption{Utility valuation of EI stopping's (EI-S) score--resource tradeoffs in Table~\ref{tab:body-f4}, on seeds 1--50. Entries are equal-task paired utility differences with the simultaneous 95\% band over the 21 cells of each column ($^\dagger$/$^\ddagger$: band above/below zero); scores lie in $[0,1]$, and utility is in the same units. On FT-PFN: EI-S minus CFBO-NT (same weights, $\alpha_{\rm CFBO}=300\lambda$) and minus ifBO run to $B=300$. On DyHPO's deep-kernel GP: EI-S minus DyHPO run to $B$, and minus EI-S on FT-PFN. $\lambda^\star=\Delta\tilde y/\Delta\tau$ is the cell-specific break-even cost against a method run to $B$, holding that cell's outcomes fixed. For $\lambda>\lambda^\star$, the priced increment savings exceed the score loss in that cell; its endpoint, score and duration change across costs. $^{\mathrm{x}}$: outside the range $\alpha_{\rm CFBO}\le1$ of \citet{cfbo2025}.}
\label{tab:learned-ei}
\fitwidth{\begin{tabular}{@{}rrrrrrr@{}}\toprule
 & \multicolumn{3}{c}{FT-PFN} & \multicolumn{3}{c}{Deep-kernel GP} \\ \cmidrule(lr){2-4}\cmidrule(lr){5-7}
$\lambda$ & $-$ CFBO-NT [band] & $-$ ifBO, $B$ [band] & $\lambda^\star$ & $-$ DyHPO, $B$ [band] & $\lambda^\star$ & $-$ EI-S on FT-PFN [band] \\ \midrule
\multicolumn{7}{@{}l}{\textit{LCBench}} \\
0.0001 & $+0.0209$ [+0.0176, +0.0241]$^{\dagger}$ & $+0.0103$ [+0.0086, +0.0119]$^{\dagger}$ & 0.00004 & $+0.0036$ [+0.0023, +0.0049]$^{\dagger}$ & 0.00005 & $-0.0092$ [-0.0109, -0.0076]$^{\ddagger}$ \\
0.0002 & $+0.0215$ [+0.0179, +0.0251]$^{\dagger}$ & $+0.0312$ [+0.0290, +0.0334]$^{\dagger}$ & 0.00005 & $+0.0112$ [+0.0097, +0.0127]$^{\dagger}$ & 0.00005 & $-0.0226$ [-0.0244, -0.0207]$^{\ddagger}$ \\
0.0005 & $+0.0220$ [+0.0174, +0.0266]$^{\dagger}$ & $+0.1030$ [+0.1000, +0.1060]$^{\dagger}$ & 0.00006 & $+0.0375$ [+0.0351, +0.0399]$^{\dagger}$ & 0.00005 & $-0.0680$ [-0.0714, -0.0647]$^{\ddagger}$ \\
0.001 & $+0.0318$ [+0.0236, +0.0399]$^{\dagger}$ & $+0.2348$ [+0.2310, +0.2386]$^{\dagger}$ & 0.00007 & $+0.0885$ [+0.0839, +0.0931]$^{\dagger}$ & 0.00005 & $-0.1489$ [-0.1549, -0.1428]$^{\ddagger}$ \\
0.002 & $+0.0582$ [+0.0457, +0.0706]$^{\dagger}$ & $+0.5177$ [+0.5135, +0.5220]$^{\dagger}$ & 0.00009 & $+0.2049$ [+0.1958, +0.2140]$^{\dagger}$ & 0.00005 & $-0.3154$ [-0.3263, -0.3045]$^{\ddagger}$ \\
0.005$^{\mathrm{x}}$ & $+0.1172$ [+0.0905, +0.1439]$^{\dagger}$ & $+1.3946$ [+1.3888, +1.4005]$^{\dagger}$ & 0.00014 & $+0.6334$ [+0.6081, +0.6587]$^{\dagger}$ & 0.00004 & $-0.7638$ [-0.7910, -0.7365]$^{\ddagger}$ \\
0.01$^{\mathrm{x}}$ & $+0.1627$ [+0.1250, +0.2004]$^{\dagger}$ & $+2.8614$ [+2.8523, +2.8706]$^{\dagger}$ & 0.00019 & $+1.5196$ [+1.4771, +1.5622]$^{\dagger}$ & 0.00004 & $-1.3444$ [-1.3907, -1.2980]$^{\ddagger}$ \\
\midrule
\multicolumn{7}{@{}l}{\textit{PD1}} \\
0.0001 & $+0.0586$ [+0.0451, +0.0721]$^{\dagger}$ & $+0.0076$ [+0.0021, +0.0130]$^{\dagger}$ & 0.00004 & $+0.0008$ [-0.0033, +0.0048]$$ & 0.00008 & $-0.0027$ [-0.0081, +0.0027]$$ \\
0.0002 & $+0.0581$ [+0.0427, +0.0734]$^{\dagger}$ & $+0.0231$ [+0.0175, +0.0288]$^{\dagger}$ & 0.00005 & $+0.0051$ [+0.0011, +0.0091]$^{\dagger}$ & 0.00008 & $-0.0139$ [-0.0194, -0.0085]$^{\ddagger}$ \\
0.0005 & $+0.0570$ [+0.0426, +0.0715]$^{\dagger}$ & $+0.0837$ [+0.0773, +0.0901]$^{\dagger}$ & 0.00007 & $+0.0193$ [+0.0150, +0.0236]$^{\dagger}$ & 0.00008 & $-0.0603$ [-0.0661, -0.0544]$^{\ddagger}$ \\
0.001 & $+0.0717$ [+0.0521, +0.0912]$^{\dagger}$ & $+0.2076$ [+0.2004, +0.2149]$^{\dagger}$ & 0.00011 & $+0.0460$ [+0.0405, +0.0515]$^{\dagger}$ & 0.00008 & $-0.1575$ [-0.1645, -0.1506]$^{\ddagger}$ \\
0.002 & $+0.1226$ [+0.0985, +0.1468]$^{\dagger}$ & $+0.4958$ [+0.4880, +0.5037]$^{\dagger}$ & 0.00015 & $+0.1117$ [+0.1019, +0.1214]$^{\dagger}$ & 0.00008 & $-0.3801$ [-0.3920, -0.3682]$^{\ddagger}$ \\
0.005$^{\mathrm{x}}$ & $+0.2175$ [+0.1753, +0.2597]$^{\dagger}$ & $+1.3722$ [+1.3666, +1.3779]$^{\dagger}$ & 0.00022 & $+0.4547$ [+0.4312, +0.4782]$^{\dagger}$ & 0.00010 & $-0.9135$ [-0.9369, -0.8900]$^{\ddagger}$ \\
0.01$^{\mathrm{x}}$ & $+0.2528$ [+0.2069, +0.2988]$^{\dagger}$ & $+2.8258$ [+2.8192, +2.8324]$^{\dagger}$ & 0.00030 & $+1.2367$ [+1.1887, +1.2847]$^{\dagger}$ & 0.00011 & $-1.5850$ [-1.6339, -1.5362]$^{\ddagger}$ \\
\midrule
\multicolumn{7}{@{}l}{\textit{ODBench}} \\
0.0001 & $+0.0437$ [+0.0353, +0.0520]$^{\dagger}$ & $-0.0025$ [-0.0052, +0.0002]$$ & 0.00012 & $+0.0109$ [+0.0097, +0.0122]$^{\dagger}$ & 0.00002 & $+0.0131$ [+0.0102, +0.0159]$^{\dagger}$ \\
0.0002 & $+0.0525$ [+0.0421, +0.0629]$^{\dagger}$ & $+0.0163$ [+0.0102, +0.0225]$^{\dagger}$ & 0.00011 & $+0.0250$ [+0.0236, +0.0265]$^{\dagger}$ & 0.00002 & $+0.0084$ [+0.0023, +0.0145]$^{\dagger}$ \\
0.0005 & $+0.0732$ [+0.0607, +0.0856]$^{\dagger}$ & $+0.0956$ [+0.0886, +0.1026]$^{\dagger}$ & 0.00012 & $+0.0704$ [+0.0678, +0.0731]$^{\dagger}$ & 0.00002 & $-0.0255$ [-0.0329, -0.0182]$^{\ddagger}$ \\
0.001 & $+0.0711$ [+0.0553, +0.0869]$^{\dagger}$ & $+0.2241$ [+0.2149, +0.2332]$^{\dagger}$ & 0.00015 & $+0.1491$ [+0.1444, +0.1539]$^{\dagger}$ & 0.00002 & $-0.0753$ [-0.0854, -0.0652]$^{\ddagger}$ \\
0.002 & $+0.0619$ [+0.0427, +0.0810]$^{\dagger}$ & $+0.4910$ [+0.4808, +0.5011]$^{\dagger}$ & 0.00020 & $+0.3187$ [+0.3084, +0.3291]$^{\dagger}$ & 0.00002 & $-0.1726$ [-0.1884, -0.1568]$^{\ddagger}$ \\
0.005$^{\mathrm{x}}$ & $+0.0816$ [+0.0519, +0.1113]$^{\dagger}$ & $+1.3271$ [+1.3131, +1.3410]$^{\dagger}$ & 0.00028 & $+0.8757$ [+0.8513, +0.9001]$^{\dagger}$ & 0.00003 & $-0.4518$ [-0.4812, -0.4223]$^{\ddagger}$ \\
0.01$^{\mathrm{x}}$ & $+0.1119$ [+0.0734, +0.1504]$^{\dagger}$ & $+2.7456$ [+2.7280, +2.7631]$^{\dagger}$ & 0.00038 & $+1.8701$ [+1.8211, +1.9190]$^{\dagger}$ & 0.00004 & $-0.8759$ [-0.9289, -0.8229]$^{\ddagger}$ \\
\bottomrule\end{tabular}}\end{table*}

\subsection{EI Stopping on DyHPO's Surrogate}
\label{app:dyhpo-ei}
\emph{Policy.} The rule of Appendix~\ref{app:ftpfn-ei} runs on DyHPO's deep-kernel GP, the released code with its
feature extractor, scaler, training schedule, seeds and hyperparameters unchanged and retrained after every
observation, on the comparator's pools and seeds; the first increment is DyHPO's initial design. EI is DyHPO's
closed-form Gaussian EI at each configuration's next increment, taken over the best score observed so far, where DyHPO takes
the best at that fidelity; the trajectory can therefore depart from DyHPO's before any stop.

\emph{Results} (Tables~\ref{tab:body-f4} and~\ref{tab:learned-ei}). Relative to DyHPO run to $B$, EI stopping reduces the score by $0.2$ to $1.4$ percentage points and saves $39$ to $188$ increments. The utility-difference band is above zero in every cell except the lowest PD1 cost, measuring how the chosen $\lambda$ values the saved increments against the score reduction. EI stopping uses more increments on this surrogate than on FT-PFN in all but one cell; its utility-difference band relative to FT-PFN lies below zero in 18 of 21 cells.

\subsection{Certificate Requirements for the Learned Surrogates}
\label{app:no-certificate-learned}
The present certificate uses the affine Gaussian updates of Section~\ref{sec:shared-models}.
The enclosure of Proposition~\ref{prop:slope-endpoint} needs each child predictive to be a Gaussian restricted
to $[0,1]$ whose mean is affine in the root observation and whose scale does not depend on it; FT-PFN's child predictive is the
network re-evaluated on each fantasized context, with no known form, so endpoint evaluations cannot enclose
the integrand. DyHPO retrains after every observation, and its feature extractor also reads the observed curve; its refitted child predictors likewise fall outside this enclosure.

The policy-improvement argument requires a fixed operational transition rule. Frozen FT-PFN weights define such a rule, as does a prescribed DyHPO retraining procedure with its random state included. Applying Proposition~\ref{prop:depth-two-rollout} requires evaluating those same updates at every fantasized child. Omitting the refit evaluates a different transition rule. Neither freezing weights nor specifying a retraining rule supplies the rigorous integrand bounds needed by our numerical gate.

\section{Scope of the Formulation and Future Work}
\label{app:scope-future}\label{app:formulation-omissions}
\paragraph{The Gap between the Planning Law and the Curves.}
The formulation has no term for the distance between $p_\theta$ and the process behind the curves and no uncertainty
over $\theta$. Our certified implementation fixes $\theta$ in advance; incorporating learning into its transition rule would require enclosures for the resulting predictive updates. The guarantees transfer to real curves only as far as the planning law does, and the effect
of $\lambda$ depends on $\theta$ (Appendix~\ref{app:fixed-law-sensitivity}).

\paragraph{A Payoff on the Best Noisy Reading.}
The incumbent is the best observed score, and the planning law adds fresh noise, so part of the value in \eqref{eq:opt-problem} is the chance of a favorable reading. A deployed configuration is worth its held-out score, which the formulation conflates with the noisy score that selects it; when the
surrogate's noise exceeds the benchmark's, the objective credits continuation with value the curves cannot deliver.

\paragraph{The Fidelity Unit and Other Simplifications.}
An action advances one benchmark-defined increment at cost $\lambda$, and the planning law is written in the same unit (its noise decay and basis scales are in checkpoints); the native horizon, the checkpoint spacing and the training behind a checkpoint appear in neither. The unit is one epoch on LCBench and ODBench and one retained checkpoint on PD1, so the same $\lambda$, lookahead depth and $\theta$ mean different things across benchmarks, whose differences confound the tasks with how finely each records training. Paired comparisons within a task are unaffected; invariance to regridding is untested.
Further, the cost is uniform, though real cost per epoch varies with the architecture; the pool is fixed; replay makes resumption free, records one realization of training noise per curve and caps
curves that would continue; and allocation is sequential, where freeze--thaw BO systems often run parallel workers.

\paragraph{Extensions.}\label{app:future-work}
A gate that opens only when the two-step lower bound is positive for every $\theta$ in a box declared in advance would prove the stop premature under every $\theta$ in the box; guaranteeing that continuing then improves on EI stopping under every $\theta$ would also need a robust reference rule. A payoff on the posterior mean of the recommended
configuration's latent score would remove the value of favorable noisy readings, at the cost of rederiving
\eqref{eq:delta-recursion} and the same-arm reduction. A learned surrogate trained on horizon-free inputs and then frozen would fall under Section~\ref{sec:theory}. Writing the planning law and $\lambda$ on cumulative
compute, as in cost-varying settings~\citep{xie2024pbgi,xie2025coststop}, would give $\lambda$ the same meaning on every
grid; EI stopping would still look one grid step ahead, so whether its decisions survive regridding remains to be tested.
Regret-bound and probabilistic-regret stopping~\citep{makarova2022termination,wilson2024prb} with tolerance $\lambda$ would be cost-aware rivals to EI stopping.
\end{document}